\documentclass[11pt,a4paper]{article}
\usepackage{amssymb}
\usepackage{times,latexsym}
\usepackage{url}
\usepackage[T2A,T1]{fontenc}
\usepackage[russian,english]{babel}
\usepackage{amsmath}
\usepackage{graphicx}
\usepackage{subcaption} 
\usepackage{booktabs}
\usepackage{multirow}
\usepackage{CJKutf8}
\usepackage{subcaption} 
\usepackage[marginpar]{todo}
\usepackage{xspace}
\usepackage{authblk}

\usepackage[acceptedWithA]{tacl2021v1}

\usepackage{xspace,mfirstuc,tabulary}

\newif\iftaclinstructions
\taclinstructionsfalse 
\iftaclinstructions
\renewcommand{\confidential}{}
\renewcommand{\anonsubtext}{(No author info supplied here, for consistency with
TACL-submission anonymization requirements)}
\newcommand{\instr}
\fi

\iftaclpubformat 

\else

\fi

\def\secref#1{\S\ref{sec:#1}}
\def\seclabel#1{\label{sec:#1}}

\usepackage{xcolor}

\definecolor{mygreen}{HTML}{4fb04f}
\newcommand{\green}[1]{\textcolor{mygreen}{#1}}

\definecolor{myblue}{HTML}{4c92c3}
\newcommand{\blue}[1]{\textcolor{myblue}{#1}}

\definecolor{myorange}{HTML}{ff9232}
\newcommand{\orange}[1]{\textcolor{myorange}{#1}}

\title{On the Entity-Level Alignment in Crosslingual Consistency}

\author[1,2,*]{\bf Yihong Liu}
\author[1,2,*]{\bf Mingyang Wang}
\author[3,$\dag$]{\bf François Yvon}
\author[1,2,$\dag$]{\bf Hinrich Sch\"utze}

\affil[1]{Center for Information and Language Processing, LMU Munich} \affil[2]{Munich Center for Machine Learning (MCML)} 
\affil[3]{Sorbonne Université, CNRS, ISIR, France
 \protect\\ \texttt{\{yihong, mingyang\}@cis.lmu.de}} 

\date{}

\newcounter{notecounter}
\newcommand{\enotesoff}{\long\gdef\enote##1##2{}}
\newcommand{\enoteson}{\long\gdef\enote##1##2{{
\stepcounter{notecounter}
{\large\bf
\hspace{0cm}\arabic{notecounter} $<<<$ ##1: ##2
$>>>$\hspace{1cm}}}}}

\enoteson
\enotesoff

\begin{document}
\maketitle

\def\thefootnote{*}\footnotetext{Equal contribution.}\def\thefootnote{\arabic{footnote}}
\def\thefootnote{$\dag$}\footnotetext{Equal advising.}\def\thefootnote{\arabic{footnote}}

\begin{abstract}
Multilingual large language models (LLMs) are expected to
recall factual knowledge consistently across languages.
However, the factors that give rise to such
\emph{crosslingual consistency} -- and its frequent failure
-- remain poorly understood.  
In this work, we hypothesize that
these inconsistencies may arise from failures in
\emph{entity alignment}, the process of mapping subject and
object entities into a shared conceptual space across
languages.
To test this, we assess alignment through entity-level (subject and object) translation tasks, and find that consistency is strongly correlated with alignment across all studied models, with misalignment of subjects or objects frequently resulting in inconsistencies.
Building on this insight, we propose
\textbf{\textsc{SubSub}} and \textbf{\textsc{SubInj}}, two
effective methods that integrate English translations of
subjects into prompts across languages, leading to
substantial gains in both factual recall accuracy and consistency.  
Finally, our mechanistic analysis
reveals that these interventions reinforce the 
entity representation alignment in the conceptual space
through
model's internal
pivot-language processing,
offering effective and practical strategies for improving multilingual factual prediction.

\end{abstract}

\begin{figure}[t]
  \centering
    \setlength{\belowcaptionskip}{-0.5cm}
  \includegraphics[width=0.46\textwidth]{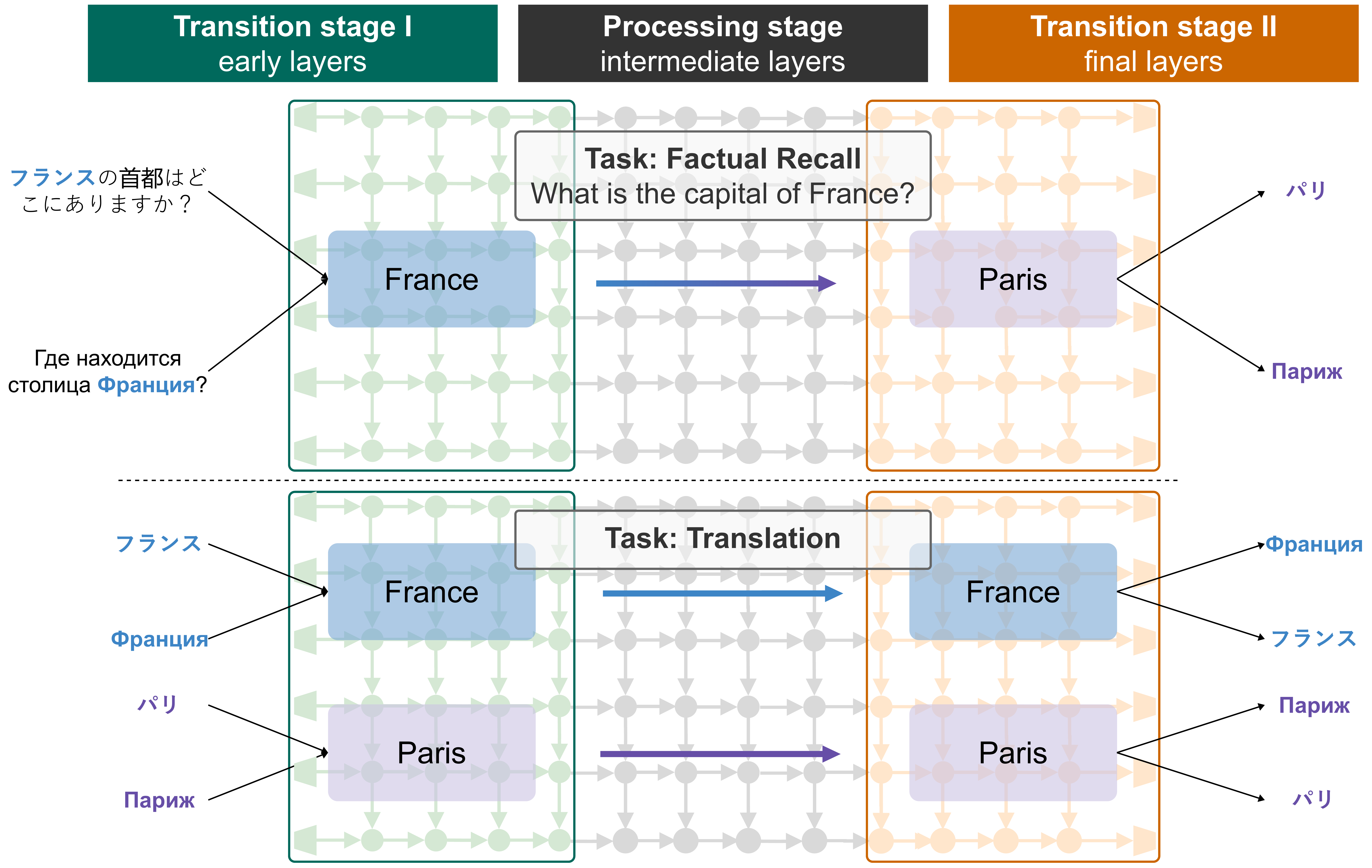}
  \caption{Analogy between consistent factual recall and entity translation. Both tasks may require mapping language-specific inputs into a shared conceptual space and projecting language-agnostic representations back into surface forms. This motivates our central hypothesis: entity alignment is important and facilitates consistent factual recall across languages.}
  \label{fig:conceptual_model}
\end{figure}

\section{Introduction}

LLMs have demonstrated remarkable capabilities in multilingual tasks, including translation, question answering, and factual recall \citep{petroni-etal-2019-language,jiang-etal-2020-x,costa2022no,englander-etal-2024-m2qa}. 
Among these, \emph{crosslingual consistency} -- the ability to recall the same fact correctly across different languages -- has emerged as a desirable property for evaluating multilingual factual grounding \citep{heinzerling-inui-2021-language,fierro-sogaard-2022-factual,qi-etal-2023-cross}. 
Yet, LLMs frequently exhibit \emph{inconsistencies}, particularly when the involved languages differ in script or linguistic structure \citep{xing2024evaluatingknowledgebasedcrosslingualinconsistency,wang2025lostmultilingualitydissectingcrosslingual,liu2025tracingmultilingualfactualknowledge}. 
Despite growing interest in this phenomenon, the underlying mechanisms that support or limit crosslingual consistency remain poorly understood.

A natural intuition suggests that consistent factual recall across languages requires a shared representation
of the \emph{entities} involved, beyond surface-level token overlap.
Humans, for instance, recognize that both
``\begin{CJK}{UTF8}{min}フランス\end{CJK}'' (Japanese) and
  ``\begin{otherlanguage*}{russian}Франція\end{otherlanguage*}''
    (Ukrainian) refer to the same entity -- \emph{France} --
    and then retrieve the relevant knowledge and express it correctly in either language. 
This requires a form of mental \emph{entity alignment} across languages. 
We hypothesize that a similar mechanism often underpins crosslingual consistency in LLMs: consistent factual recall is facilitated when entities (subject and object) are aligned across languages in the model's internal shared language-agnostic conceptual space (see illustration of the conceptual model in Figure~\ref{fig:conceptual_model}).

Such alignment facilitates the two crucial \emph{transition stages} that occur in multilingual factual recall according to the conceptual model: (1) mapping language-specific inputs into a shared conceptual space, and (2) projecting the shared latent representation back into the correct surface realization in the target languages.
Inconsistent factual recall across two languages may arise if entity alignment fails at either of these transition stages.\footnote{Inconsistency can also stem from failure in the processing stage -- i.e., the model lacks the factual knowledge so it cannot retrieve the correct object in the conceptual space, which is beyond the scope of discussion in this paper.}
This hypothesis aligns with recent work suggesting that multilingual LLMs, especially those trained predominantly on English, often operate in an \emph{English-centric latent space} \citep{wendler-etal-2024-llamas,schut-etal-2025-do,dumas2025separatingtonguethoughtactivation}. 
During multilingual factual recall, the model implicitly translates input from a non-English language into English-like conceptual representations, processes information in this \emph{pivot-language} conceptual space, and then generates output in the desired target language \citep{wang2025lostmultilingualitydissectingcrosslingual,lu2025pathstakenunderstandingmending}. 
Such a pipeline inherently relies on robust crosslingual alignment, particularly for subject and object entities that serve as anchors for factual retrieval.

To operationalize this hypothesis, we introduce an \emph{entity-level translation task} designed to probe how well models align subject and object entities between language pairs. 
This task instantiates the same transition stages posited by the conceptual model (cf. Figure~\ref{fig:conceptual_model}), offering a concrete lens through which to assess the model's crosslingual entity-level alignment capabilities. 
We then evaluate the factual recall and entity-level alignment of a spectrum of LLMs from several model families across multiple sizes, using  KLAR \citep{wang2025lostmultilingualitydissectingcrosslingual}, a multilingual factual knowledge dataset covering 17 languages.

Our key contributions are as follows:

\textbf{(i)} We provide the first systematic analysis of the relationship between \textbf{crosslingual consistency} and \textbf{entity-level alignment}, using a dedicated entity translation probing task to measure entity alignment quality.
\textbf{(ii)} In \secref{relationship}, we show that consistency is \textbf{strongly correlated} with entity alignment, and that consistent factual recall rarely occurs without correctly aligning either the subject or the object entities between two languages.
\textbf{(iii)} In \secref{remedies}, we introduce two prompting-based remedies, \textbf{\textsc{SubSub}} and \textbf{\textsc{SubInj}}, which incorporate English translation of subjects into prompts. These methods consistently and substantially enhance recall and consistency, particularly in English-centric models.
\textbf{(iv)} Finally, in \secref{mechanistic}, we analyze \textsc{SubSub} and \textsc{SubInj} through the lens of mechanistic interpretability, showing that these interventions enhance entity representation alignment through pivot-language processing and thereby facilitate consistent factual recall.

\section{Related Work}

\subsection{Crosslingual Alignment}

Prior work on crosslingual alignment has largely focused on \emph{representational alignment}, where semantically similar units across languages are mapped to nearby vectors in embedding space \citep{artetxe-schwenk-2019-massively,reimers-gurevych-2020-making}. 
This alignment is typically evaluated via retrieval tasks such as sentence alignment or word-level correspondence, which operationalize the notion of weak alignment \citep{roy-etal-2020-lareqa,hammerl-etal-2024-understanding,liu-etal-2025-transliterations}. 
However, these embedding-based evaluations remain indirect and often correlate loosely with performance on downstream tasks, particularly in generation settings \citep{kargaran2025mexamultilingualevaluationenglishcentric}, where alignment must manifest at the level of surface forms. 
To address this gap, we evaluate alignment explicitly through input-output behavior, using translation accuracy as a proxy. 
This surface-level alignment provides a more functional view of whether subject and object entities are correctly realized across languages, and we show it to be strongly 
correlated
with crosslingual consistency.




\subsection{Factual Recall and Consistency}
Pretrained language models have been shown to function as knowledge bases, as first proposed by \citet{petroni-etal-2019-language}. Following this idea, a series of studies have explored the factual knowledge stored in these models using knowledge probing techniques, focusing either on English \citep{roberts2020much, peng2022copen} or on multilingual factual knowledge \citep{jiang-etal-2020-x, kassner-etal-2021-multilingual, yin-etal-2022-geomlama,fierro-etal-2025-multilingual}.
Building on multilingual probing, recent work examines crosslingual consistency, the extent to which models return consistent answers to equivalent queries in different languages. \citet{qi-etal-2023-cross} show that LLMs often produce divergent answers across languages, highlighting inconsistencies in multilingual factual recall. \citet{wang2025lostmultilingualitydissectingcrosslingual} and \citet{lu2025pathstakenunderstandingmending} further investigate how internal representations lead to these inconsistencies, while other work traces how crosslingual consistency emerges and evolves during pretraining \citep{liu2025tracingmultilingualfactualknowledge}.

In this work, we extend this line of research by identifying entity alignment -- the model's ability to represent subject and object entities consistently across languages -- as a key factor underlying crosslingual consistency.
We show a strong correlation between entity alignment and crosslingual consistency, and consistent factual recall hardly occurs when the model fails to align entities.
Building on this insight, we propose two prompting-based interventions, which enhance entity alignment and thereby facilitate crosslingual factual consistency. 
Finally, we perform a mechanistic interpretability analysis to explain the success of these interventions, offering both theoretical insights and practical strategies for improving multilingual factual prediction in LLMs.

\section{Methodology\seclabel{methodo}}

\subsection{Languages, Models, and Dataset}\seclabel{data}

\textbf{Languages.} We consider 17 languages that span 8 language families and use 8 different scripts: Arabic (\textbf{ara\_Arab}), Catalan (\textbf{cat\_Latn}), Chinese (\textbf{zho\_Hans}), Dutch (\textbf{nld\_Latn}), English (\textbf{eng\_Latn}), French (\textbf{fra\_Latn}), Greek (\textbf{ell\_Grek}), Hebrew (\textbf{heb\_Hebr}), Hungarian (\textbf{hun\_Latn}), Japanese (\textbf{jpn\_Jpan}), Korean (\textbf{kor\_Kore}), Persian (\textbf{fas\_Arab}), Russian (\textbf{rus\_Cyrl}), Spanish (\textbf{spa\_Latn}), Turkish (\textbf{tur\_Latn}), Ukrainian (\textbf{ukr\_Cyrl}), and Vietnamese (\textbf{vie\_Latn}).

\textbf{Models.} We evaluate a diverse suite of 12 decoder-only language models spanning 4 major model families: \textbf{LLaMA} \citep{grattafiori2024llama3herdmodels}, \textbf{Qwen} \citep{yang2025qwen3technicalreport}, \textbf{Gemma} \citep{gemmateam2025gemma3technicalreport}, and \textbf{OLMo} \citep{olmo20252olmo2furious}. 
The first three are trained on highly multilingual corpora, while \textbf{OLMo} is primarily trained on English data. 
From the LLaMA family, we consider \texttt{Llama-3.2-1B}, \texttt{Llama-3.2-3B}, and \texttt{Llama-3.1-8B}. 
From the Qwen family, we consider \texttt{Qwen3-1.7B-Base}, \texttt{Qwen3-4B-Base}, and \texttt{Qwen3-8B-Base}. 
From the Gemma family, we consider \texttt{gemma-3-1b-pt}, \texttt{gemma-3-4b-pt}, and \texttt{gemma-3-12b-pt}.
Finally, we consider \texttt{OLMo-2-0425-1B}, \texttt{OLMo-2-1124-7B}, and \texttt{OLMo-2-1124-13B} from the OLMo series. 
This selection allows us to systematically study model behavior across size and family.

\textbf{Multilingual Factual Dataset.}
We conduct our investigation using KLAR \citep{wang2025lostmultilingualitydissectingcrosslingual}, a multilingual factual knowledge probing dataset.
Our evaluation covers \textbf{2,619} language-agnostic facts spanning \textbf{20} distinct relation types (cf.\ Table~\ref{tab:relation_fact_counts} in \secref{klar}). 
Each fact is represented as a triple $(s_i, r_i, o_i)$, where $s_i$ and $o_i$ are the subject and object entities, and $r_i$ is the relation. 
We denote the full language-agnostic set of facts as $\mathcal{F} = \{(s_i, r_i, o_i)\}_{i=1}^{N}$.
For each fact $(s_i, r_i, o_i) \in \mathcal{F}$ and each language $l$, KLAR provides corresponding language-specific realizations $(s_i^l, r_i, o_i^l)$, where $s_i^l$ and $o_i^l$ are the subject and object translated into language $l$. 
Thus, for any two languages $l$ and $l'$, the instances
$(s_i^l, r_i, o_i^l)$ and $(s_i^{l'}, r_i, o_i^{l'})$
represent aligned expressions of the same underlying fact in
the two languages.
KLAR also provides a set of language-specific prompt templates for each relation, which we use to construct factual recall queries in different languages (cf.\ \secref{fact_recall_evaluation}).


\subsection{Entity-Level Alignment Evaluation}\seclabel{entity_translation_evaluation}

\textbf{Task Formulation.}
To evaluate crosslingual entity-level alignment, we formulate a multilingual entity translation task over the language-agnostic fact set $\mathcal{F} = \{(s_i, r_i, o_i)\}_{i=1}^N$. 
For each language pair $(l_1, l_2)$ and each fact $(s_i, r_i, o_i) \in \mathcal{F}$, we construct two translation sub-tasks: (1) translating the subject entity $s_i^{l_1} \rightarrow s_i^{l_2}$, and (2) translating the object entity $o_i^{l_1} \rightarrow o_i^{l_2}$. 
The model is prompted in the source language $l_1$ and is expected to generate the corresponding entity in the target language $l_2$.
Each prompt includes a 3-shot demonstration consisting of aligned entity pairs 
drawn from other facts in the dataset.
An example of a subject translation prompt from Japanese to English is displayed below:
\[
\boxed{
\begin{array}{l}
\text{\begin{CJK}{UTF8}{min}日本語: フランス\end{CJK} - English: France} \\
\text{\begin{CJK}{UTF8}{min}日本語: セルビア\end{CJK} - English: Serbia} \\
\text{\begin{CJK}{UTF8}{min}日本語: イタリア\end{CJK} - English: Italy} \\
\text{\begin{CJK}{UTF8}{min}日本語: イギリス\end{CJK} - English: }
\end{array}
}
\]
The model is expected to complete the last translation with the correct entity name in the target language (``\emph{United Kingdom}'' in this case).
\textbf{Metrics.}
For each direction $(l_1 \rightarrow l_2)$ and each fact $(s_i, r_i, o_i) \in \mathcal{F}$, we evaluate whether the model’s generated output contains the correct target entity. Specifically, we define:

\textbf{Subject translation accuracy}:
  \[
    \mathrm{ACC}^{\mathrm{sub}}_{l_1 \rightarrow l_2} = \frac{1}{|\mathcal{F}|} \sum_{i=1}^{|\mathcal{F}|} \mathbf{1} \left[ s_i^{l_2} \subseteq \mathcal{M}(s_i^{l_1}) \right]
  \]
  
\textbf{Object translation accuracy}:
  \[
    \mathrm{ACC}^{\mathrm{obj}}_{l_1 \rightarrow l_2} = \frac{1}{|\mathcal{F}|} \sum_{i=1}^{|\mathcal{F}|} \mathbf{1} \left[ o_i^{l_2} \subseteq \mathcal{M}(o_i^{l_1}) \right]
  \]

\textbf{Joint translation accuracy}:
  \[
  \begin{aligned}
    \mathrm{ACC}^{\mathrm{both}}_{l_1 \rightarrow l_2} = \frac{1}{|\mathcal{F}|} \sum_{i=1}^{|\mathcal{F}|} 
    \mathbf{1} \Big[ & s_i^{l_2} \subseteq \mathcal{M}(s_i^{l_1}) \\
                    & \land\ o_i^{l_2} \subseteq \mathcal{M}(o_i^{l_1}) \Big]
  \end{aligned}
  \]
where $\mathcal{M}(x)$ denotes the model's \emph{full generated output} when prompted with $x$, and $\mathbf{1}[\cdot]$ is the indicator function. 
Our evaluation differs from prior work that only checks the first generated token \citep{geva-etal-2023-dissecting, qi-etal-2023-cross, Hernandez2024Linearity}, ensuring correctness at the string level, especially for \emph{multi-token entities}.
To assess alignment between $l_1$ and $l_2$, we average the translation accuracies in both directions. Specifically:
\[
\begin{aligned}
\mathrm{Align}^{\mathrm{sub}}(l_1, l_2) &= \frac{\mathrm{ACC}^{\mathrm{sub}}_{l_1 \rightarrow l_2} + \mathrm{ACC}^{\mathrm{sub}}_{l_2 \rightarrow l_1}}{2} \\
\mathrm{Align}^{\mathrm{obj}}(l_1, l_2) &= \frac{\mathrm{ACC}^{\mathrm{obj}}_{l_1 \rightarrow l_2} + \mathrm{ACC}^{\mathrm{obj}}_{l_2 \rightarrow l_1}}{2} \\
\mathrm{Align}^{\mathrm{both}}(l_1, l_2) &= \frac{\mathrm{ACC}^{\mathrm{both}}_{l_1 \rightarrow l_2} + \mathrm{ACC}^{\mathrm{both}}_{l_2 \rightarrow l_1}}{2}
\end{aligned}
\]
The alignment scores are computed across all $\binom{17}{2} = 136$ unique language pairs, yielding a comprehensive view of entity-level alignment.


\subsection{Factual Recall Evaluation}\seclabel{fact_recall_evaluation}

\textbf{Task Formulation.}
Given the multilingual fact set $\mathcal{F} = \{(s_i, r_i, o_i)\}_{i=1}^N$, KLAR provides a language-specific prompt template for each relation $r_i$ in every language $l$. 
Each fact can be instantiated in language $l$ as a \emph{query-answer pair} $(q_i^l, o_i^l)$, where $q_i^l$ is the prompt formed using $s_i^l$ and $r_i$, and $o_i^l$ is the language-specific realization of the object $o_i$. 
For example, the fact (\emph{France, capital, Paris}) may be rendered in English as ``\emph{Where is France's capital located? The answer is:}'' with the expected answer ``\emph{Paris}''.
This setup ensures that for each fact, we can construct 17 language-specific queries, one per language.
The queries $q_i^{l_1}$ and $q_i^{l_2}$ for languages $l_1$ and $l_2$ share the same underlying fact and differ only in surface realization.
To facilitate factual recall, we apply a 3-shot prompting strategy analogous to that used in the entity translation task (cf.\ \secref{entity_translation_evaluation}). 
Each prompt is preceded by three example question-answer pairs in the same language and relation type, selected from other facts in the dataset.
This few-shot design provides a contextual signal for relation-specific generation while preserving the language setting. 

\textbf{Metrics.}
We use \textbf{accuracy} and \textbf{crosslingual consistency} to evaluate model performance. 
The per-language factual recall accuracy is defined as:
\[
\mathrm{ACC}(l) = \frac{1}{|\mathcal{F}|} \sum_{i=1}^{|\mathcal{F}|} \mathbf{1}\left[ o_i^l \subseteq \mathcal{M}(q_i^l) \right]
\]
where $\mathcal{M}(q_i^l)$ denotes the model's complete generation for prompt $q_i^l$, and correctness is judged by string inclusion as described in \secref{entity_translation_evaluation}.
To assess crosslingual consistency of a language pair $(l_1, l_2)$, we compute the Jaccard index:\footnote{We adopt the definition of consistency used in prior work \citep{jiang-etal-2020-x,wang2025lostmultilingualitydissectingcrosslingual}, which is stricter than mere agreement: it requires that the model's predictions be both identical \emph{and} correct across the two languages.}
\[
\begin{aligned}
\mathrm{CO}(l_1, l_2) 
&= \textstyle\frac{
\sum_{i=1}^{|\mathcal{F}|} \mathbf{1} [ o_i^{l_1} \subseteq \mathcal{M}(q_i^{l_1}) \land o_i^{l_2} \subseteq \mathcal{M}(q_i^{l_2})]
}
{
\sum_{i=1}^{|\mathcal{F}|} \mathbf{1} [ o_i^{l_1} \subseteq \mathcal{M}(q_i^{l_1}) \lor o_i^{l_2} \subseteq \mathcal{M}(q_i^{l_2})]
}
\end{aligned}
\]
We compute $\mathrm{CO}(l_1, l_2)$ for all $\binom{17}{2} = 136$ language pairs to obtain a comprehensive map of crosslingual factual consistency.

\section{Relationship Between Crosslingual Consistency and Entity Alignment}\seclabel{relationship}

\begin{figure*}[h]
    \centering
    \setlength{\belowcaptionskip}{-0.4cm}
    \includegraphics[width=0.15\textwidth]{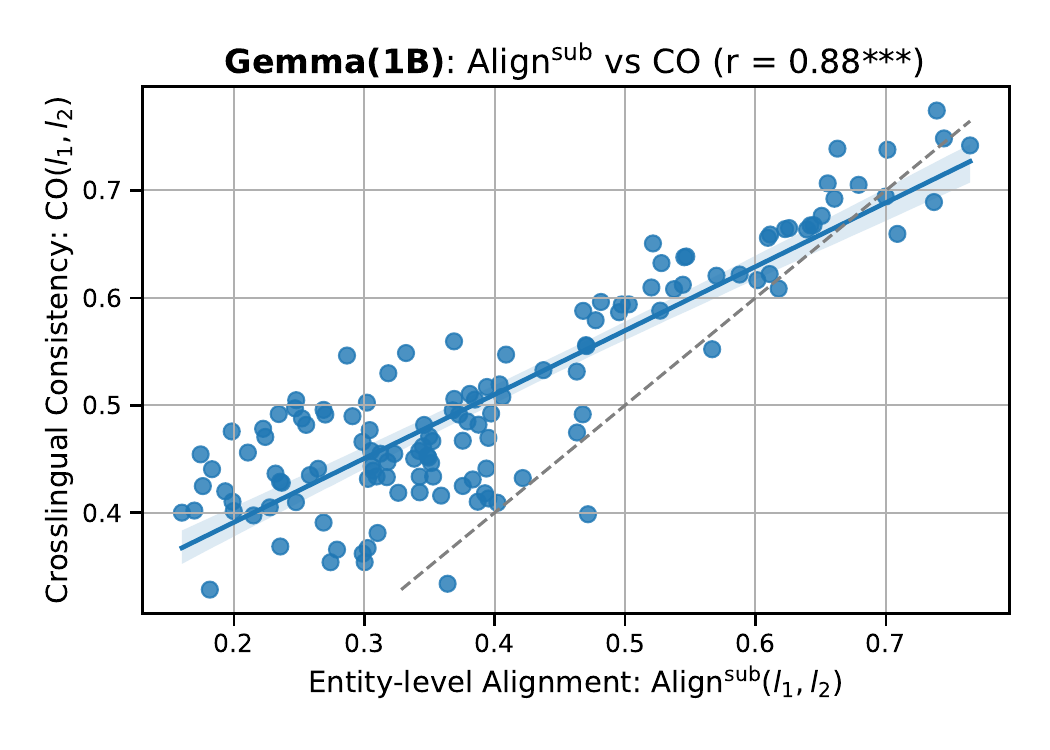}
    \includegraphics[width=0.15\textwidth]{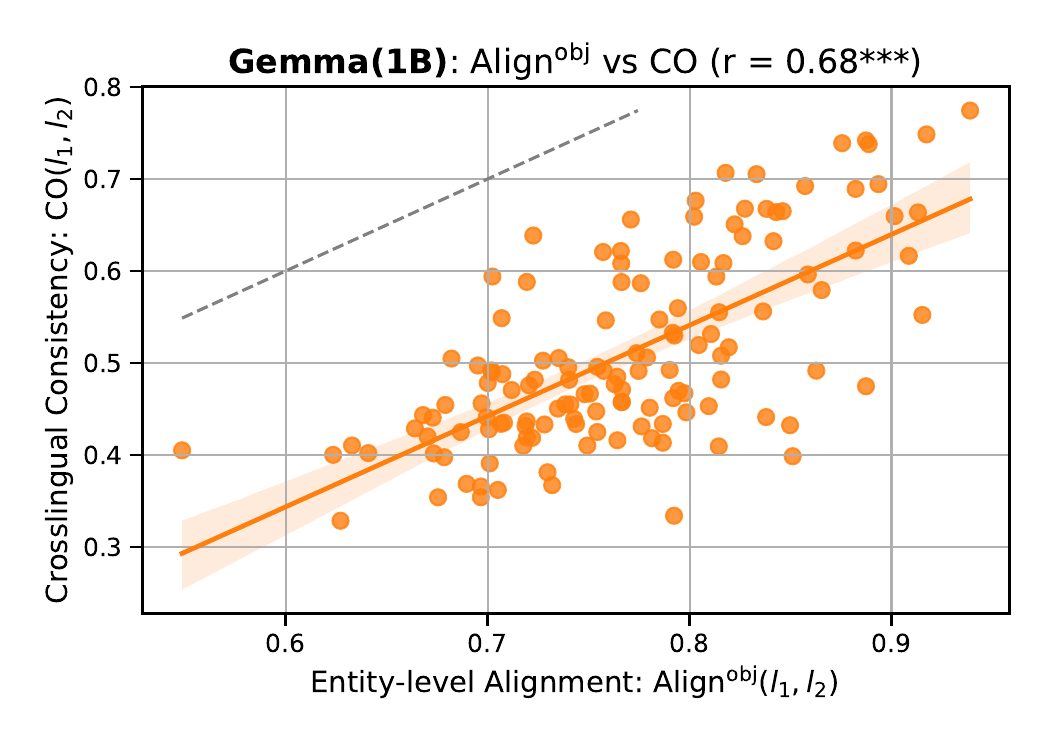}
    \includegraphics[width=0.15\textwidth]{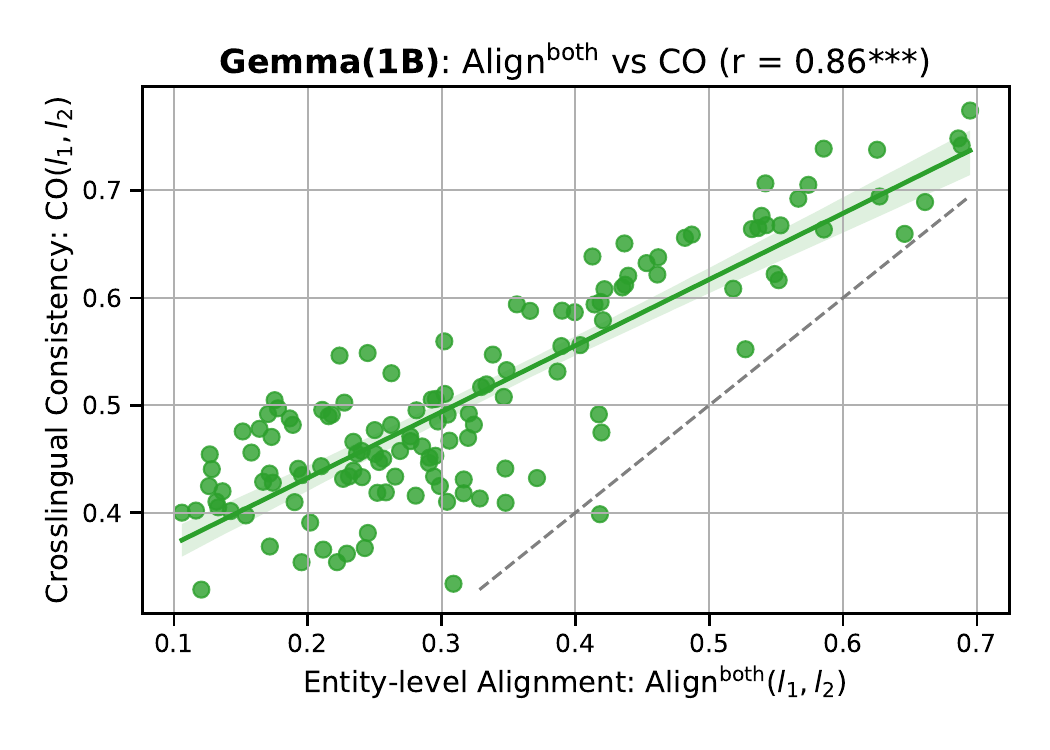}
    \includegraphics[width=0.15\textwidth]{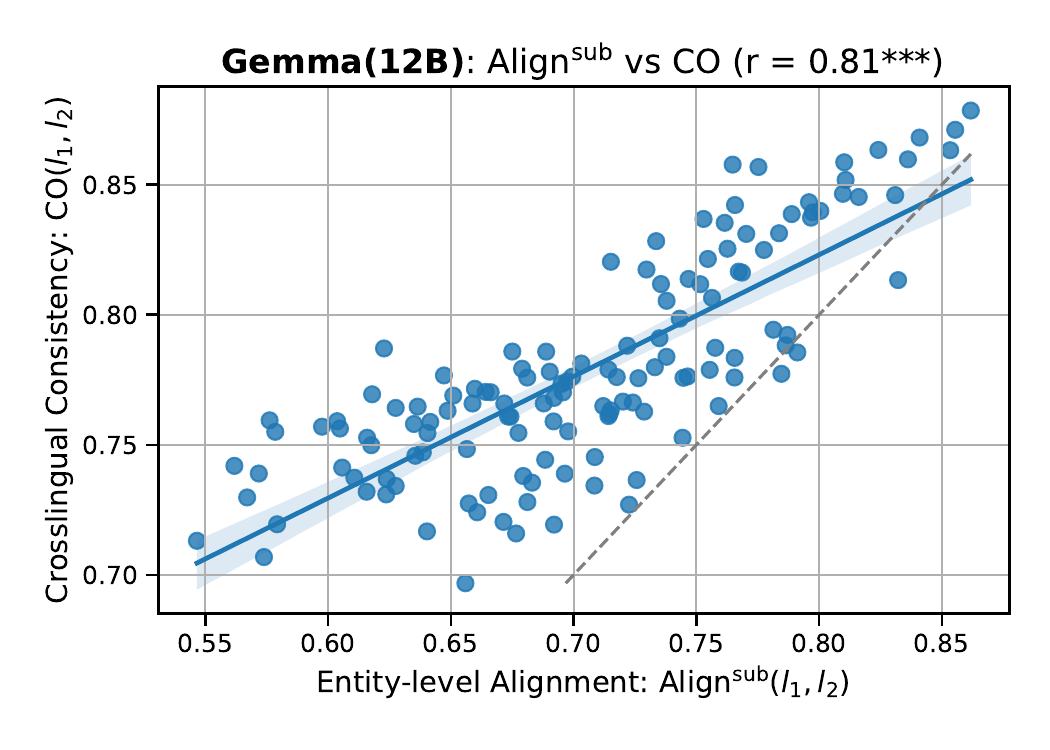}
    \includegraphics[width=0.15\textwidth]{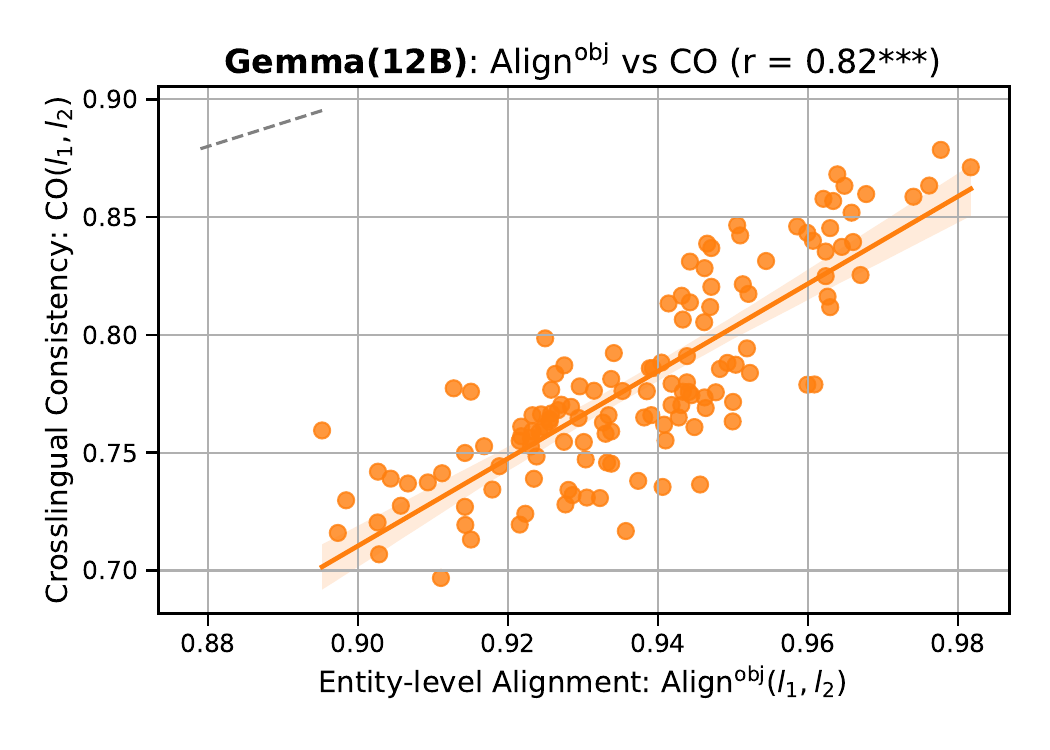}
    \includegraphics[width=0.15\textwidth]{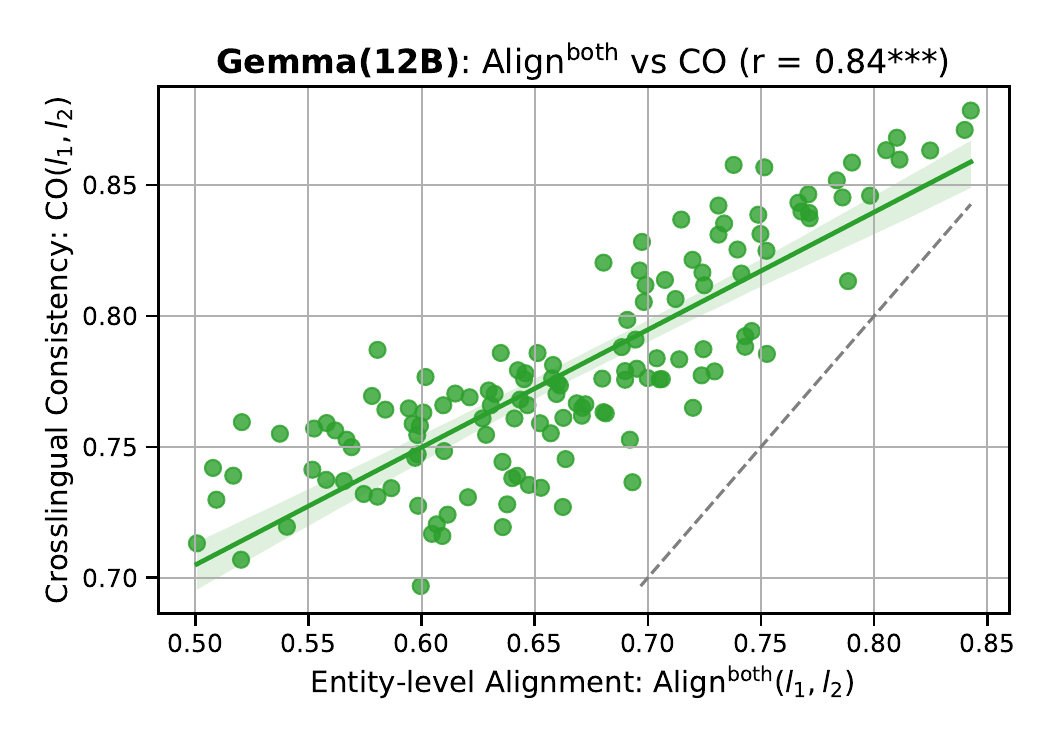}
    \includegraphics[width=0.15\textwidth]{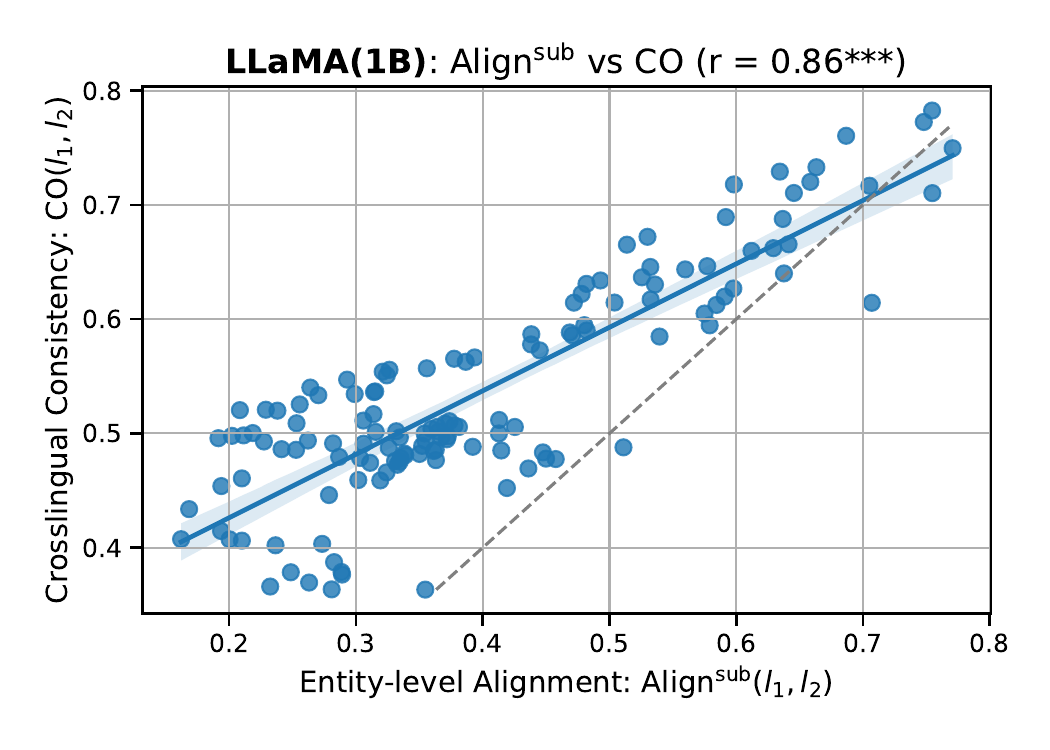}
    \includegraphics[width=0.15\textwidth]{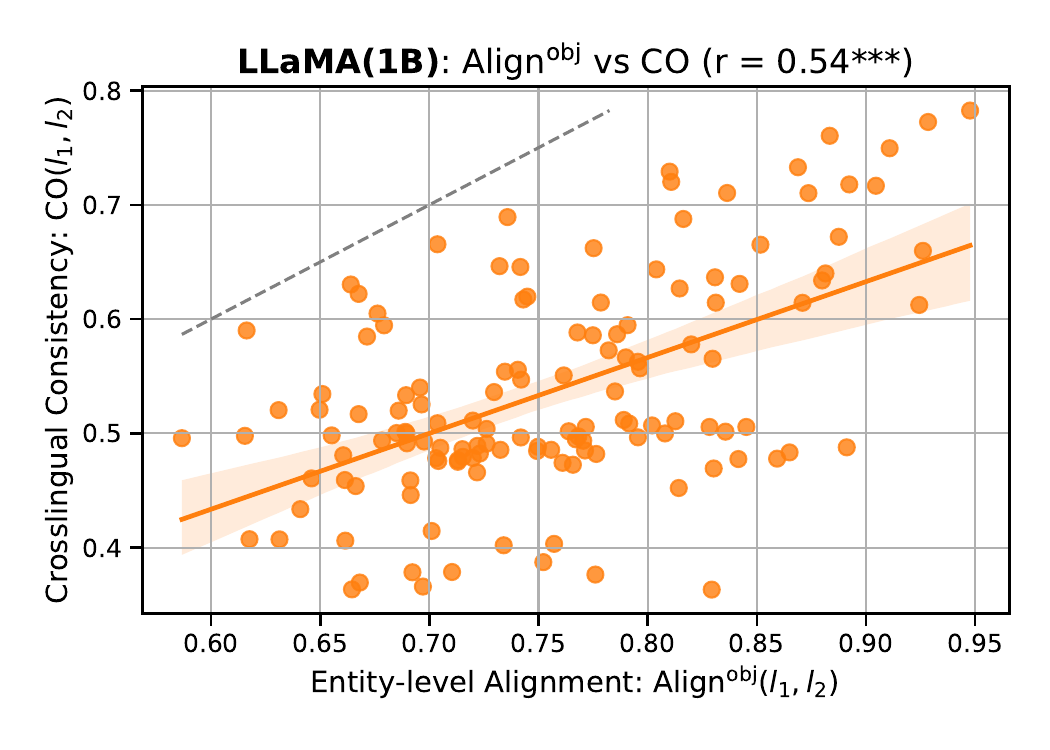}
    \includegraphics[width=0.15\textwidth]{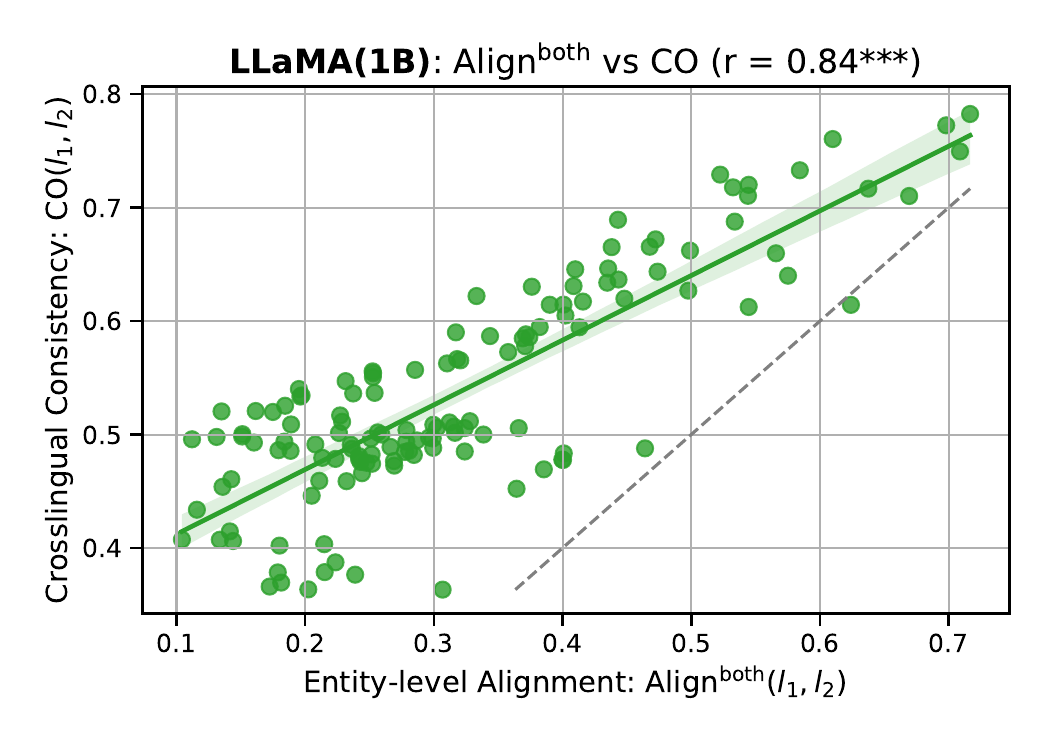}
    \includegraphics[width=0.15\textwidth]{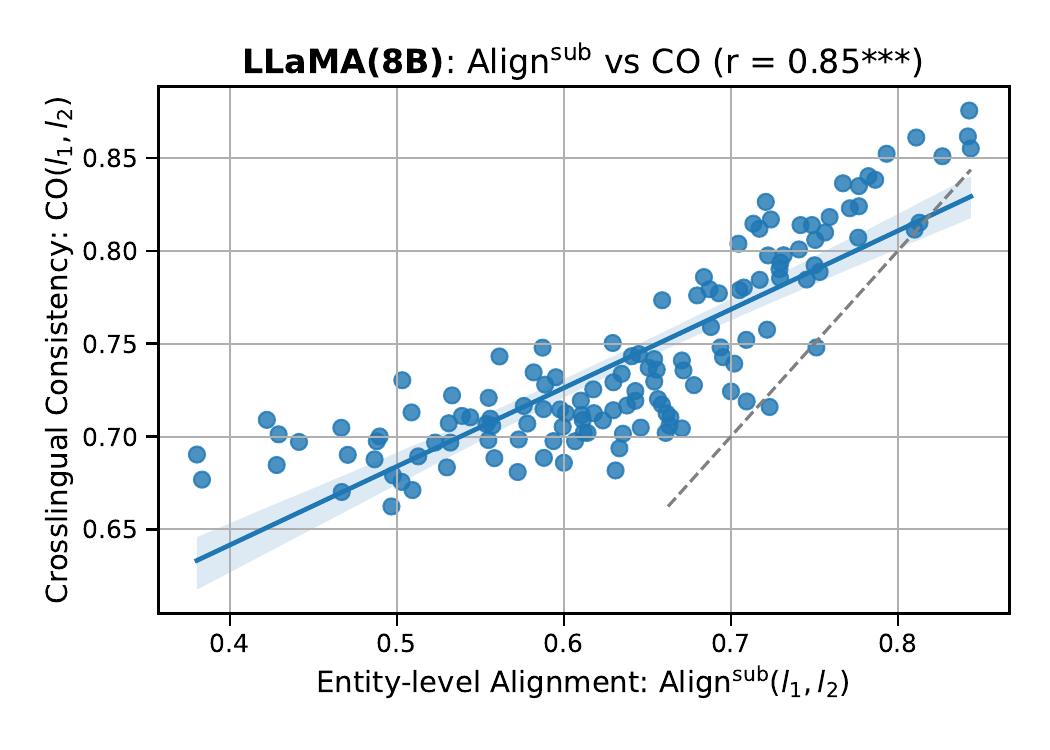}
    \includegraphics[width=0.15\textwidth]{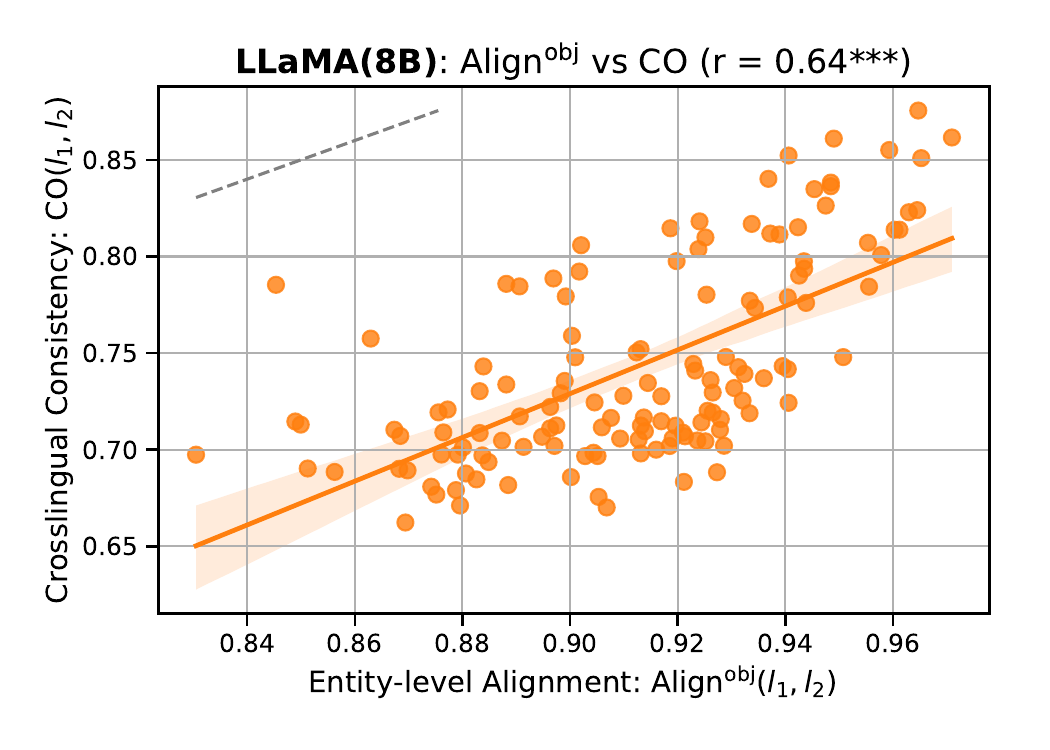}
    \includegraphics[width=0.15\textwidth]{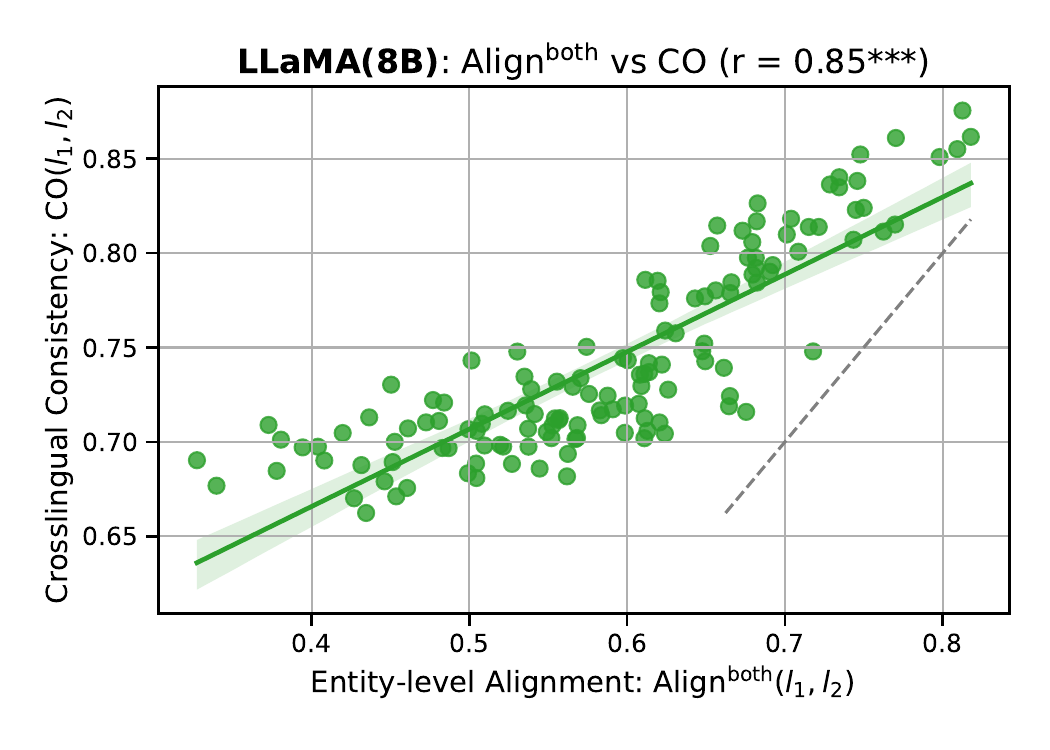}
    \includegraphics[width=0.15\textwidth]{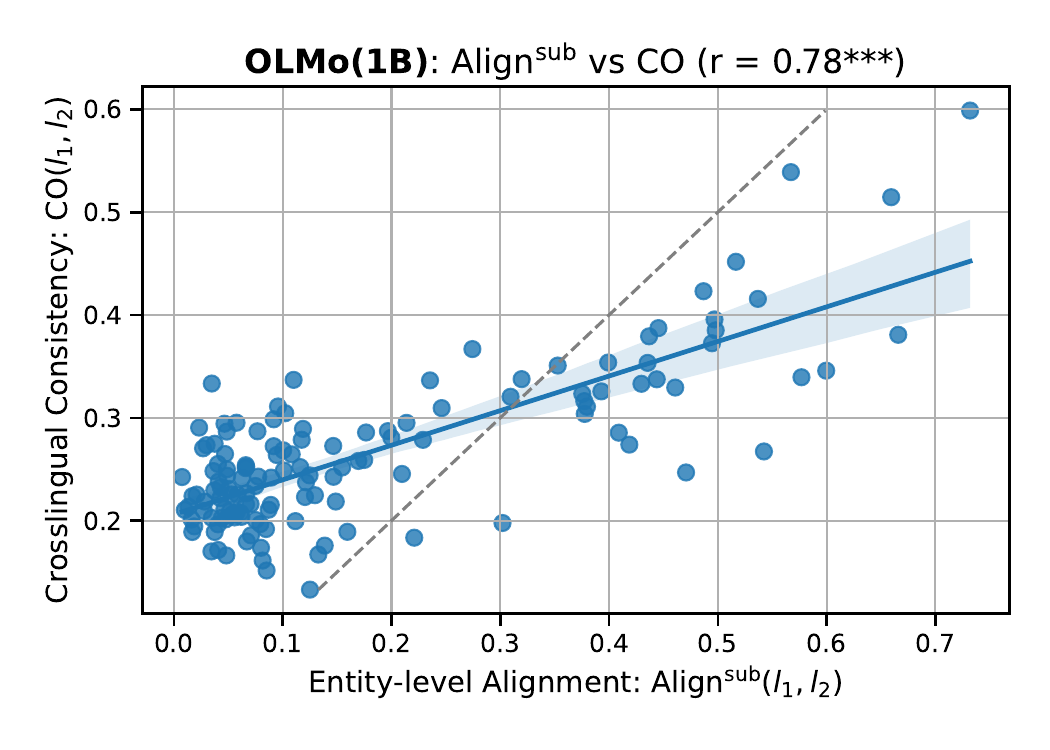}
    \includegraphics[width=0.15\textwidth]{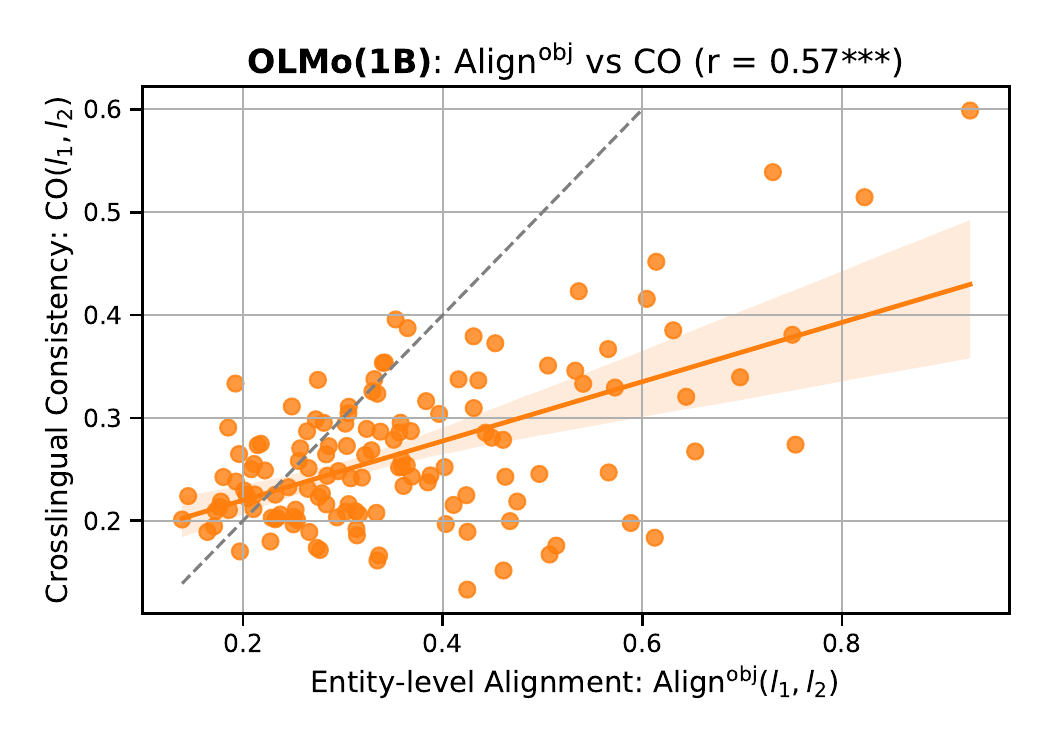}
    \includegraphics[width=0.15\textwidth]{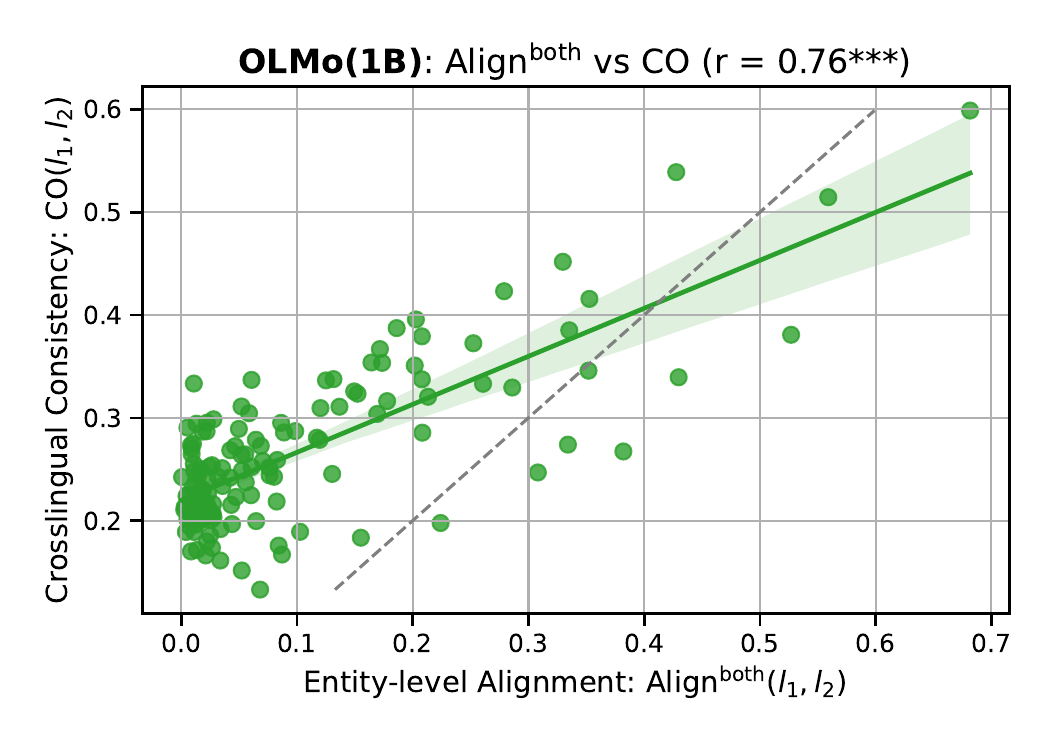}
    \includegraphics[width=0.15\textwidth]{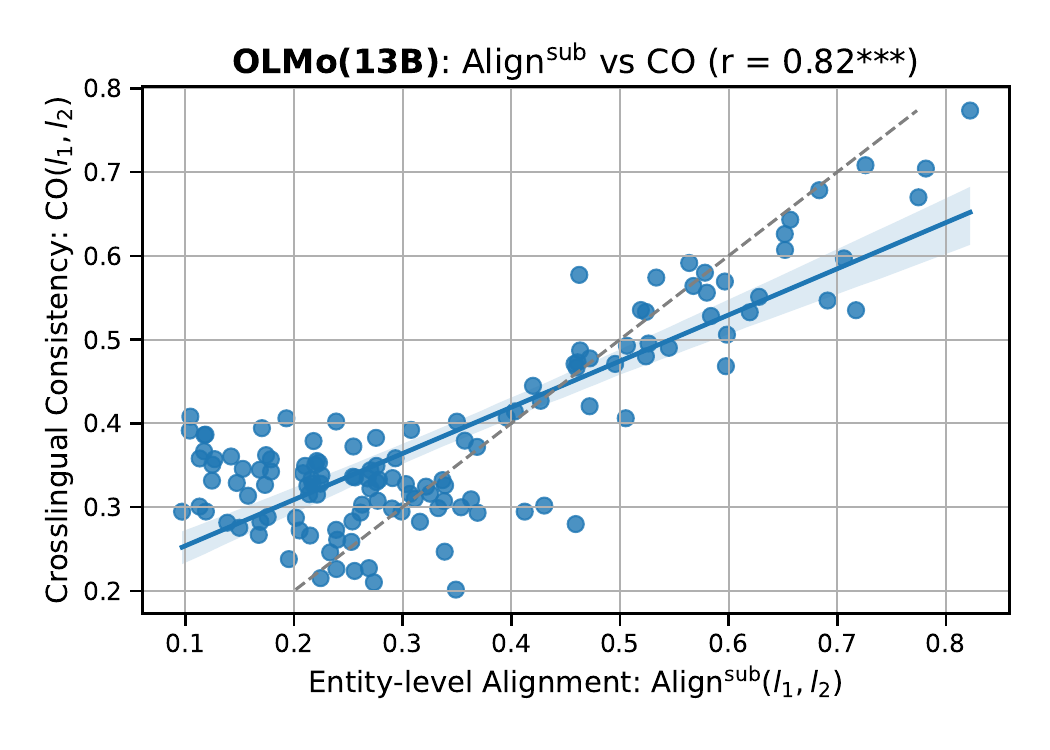}
    \includegraphics[width=0.15\textwidth]{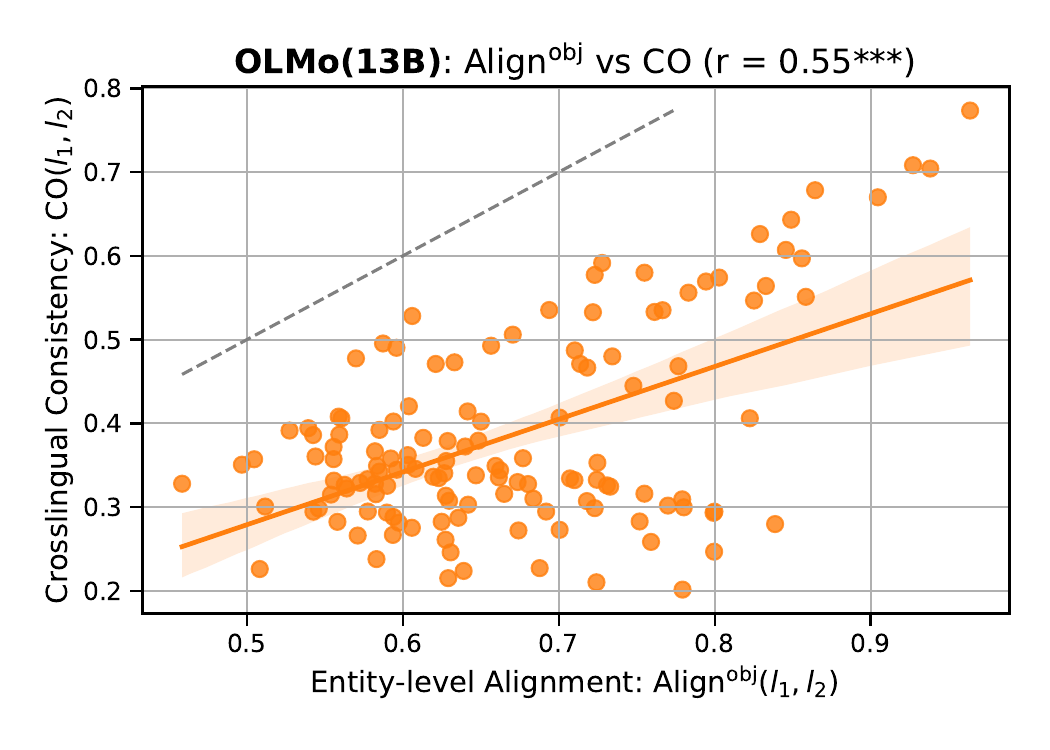}
    \includegraphics[width=0.15\textwidth]{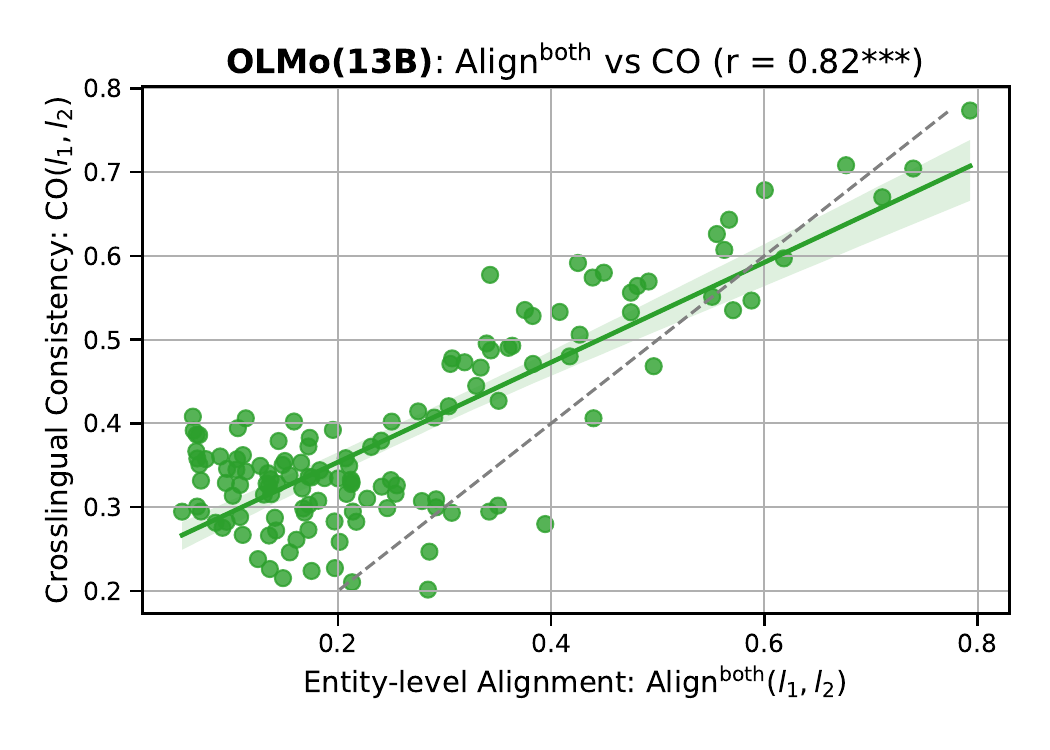}
    \includegraphics[width=0.15\textwidth]{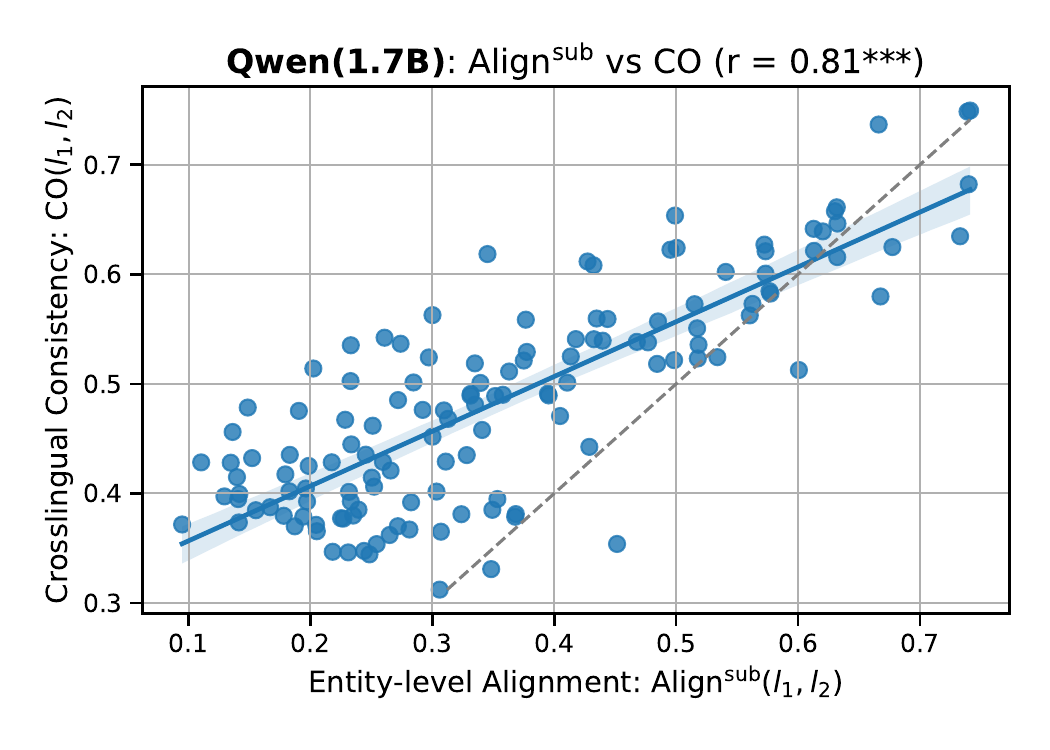}
    \includegraphics[width=0.15\textwidth]{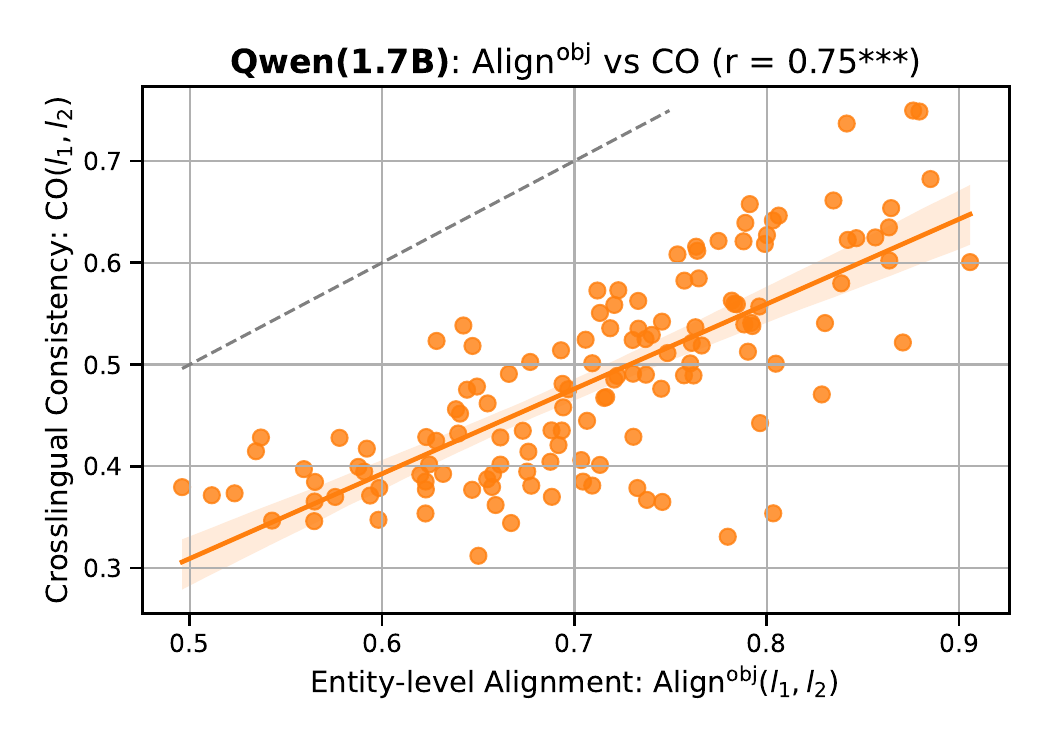}
    \includegraphics[width=0.15\textwidth]{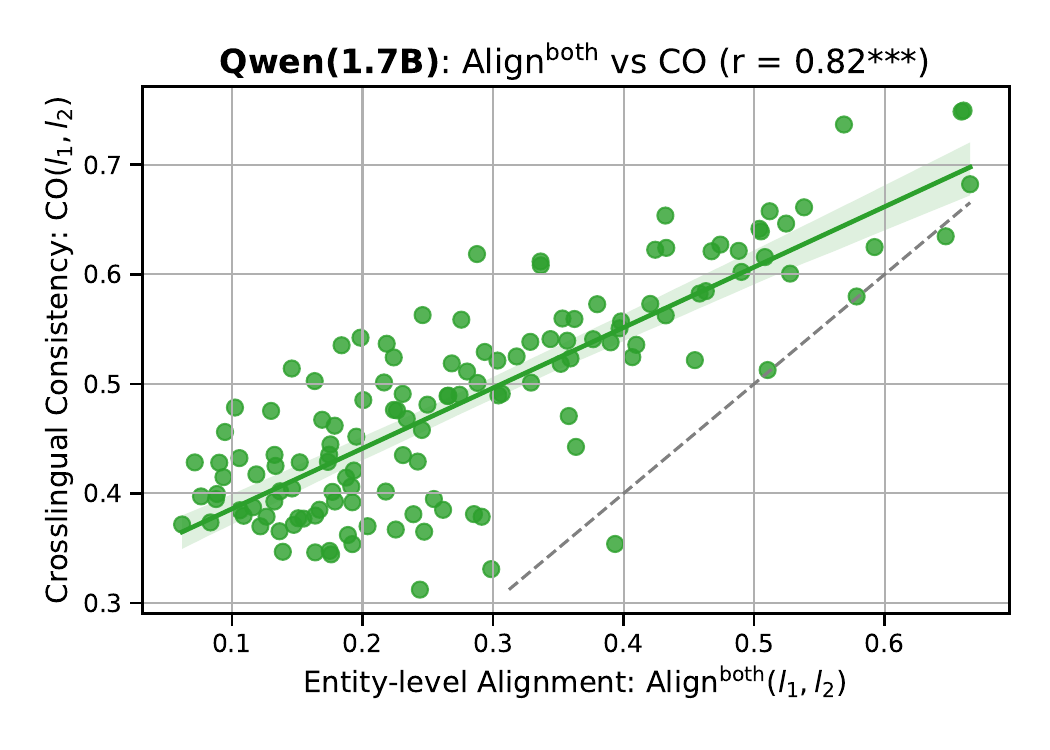}
    \includegraphics[width=0.15\textwidth]{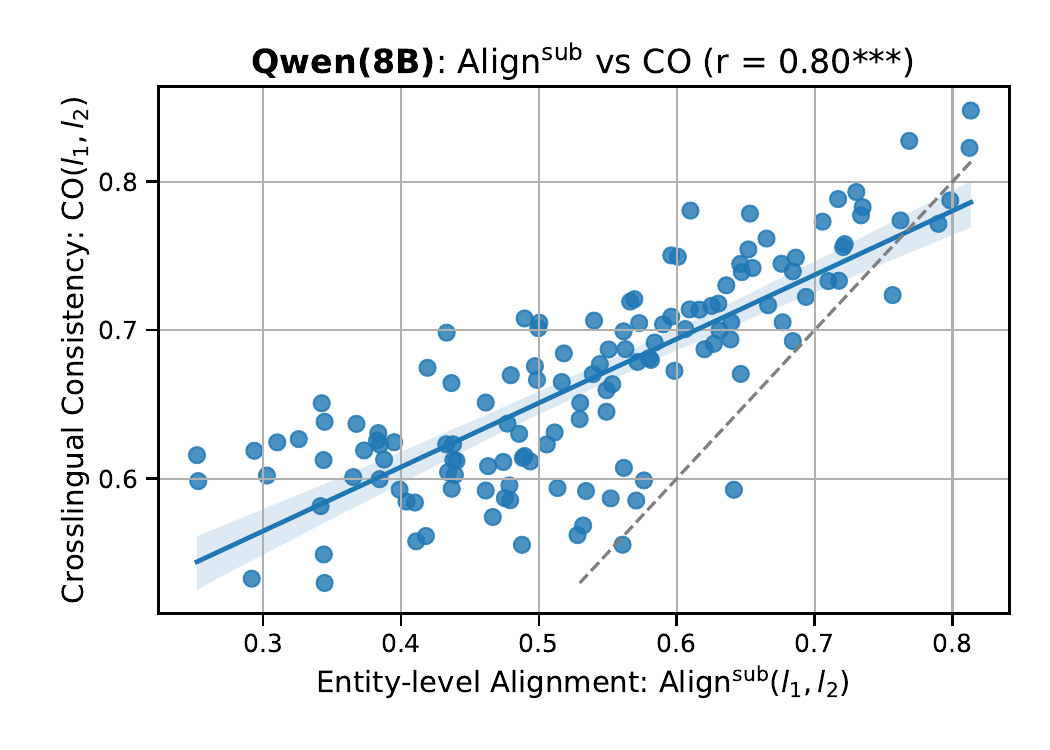}
    \includegraphics[width=0.15\textwidth]{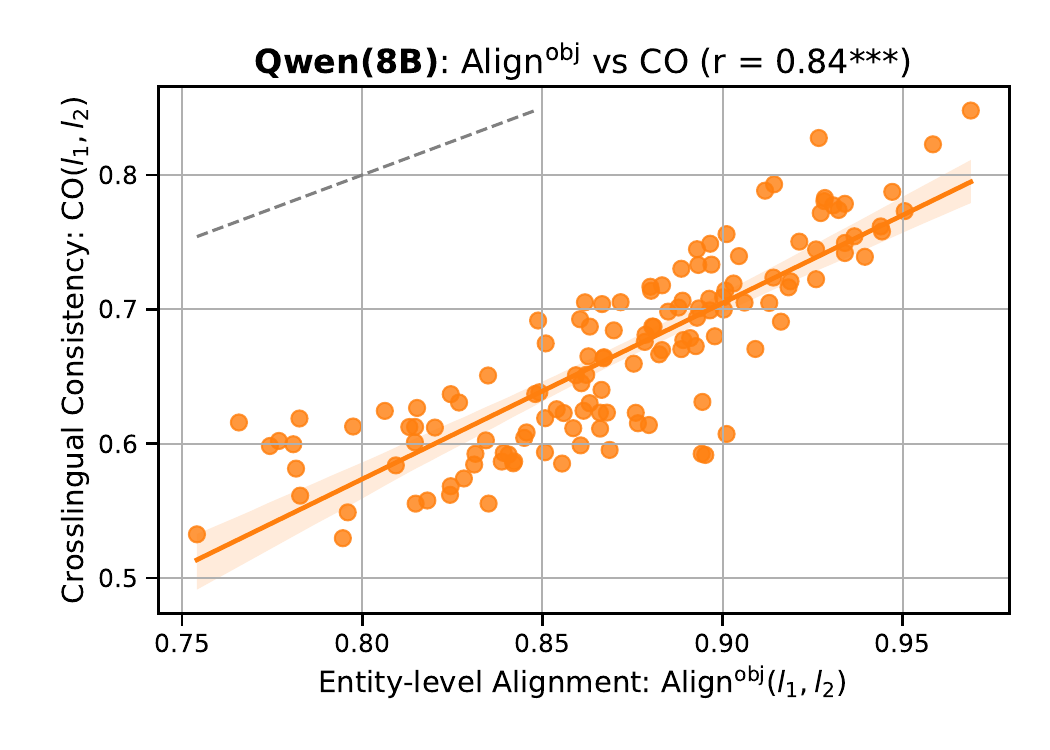}
    \includegraphics[width=0.15\textwidth]{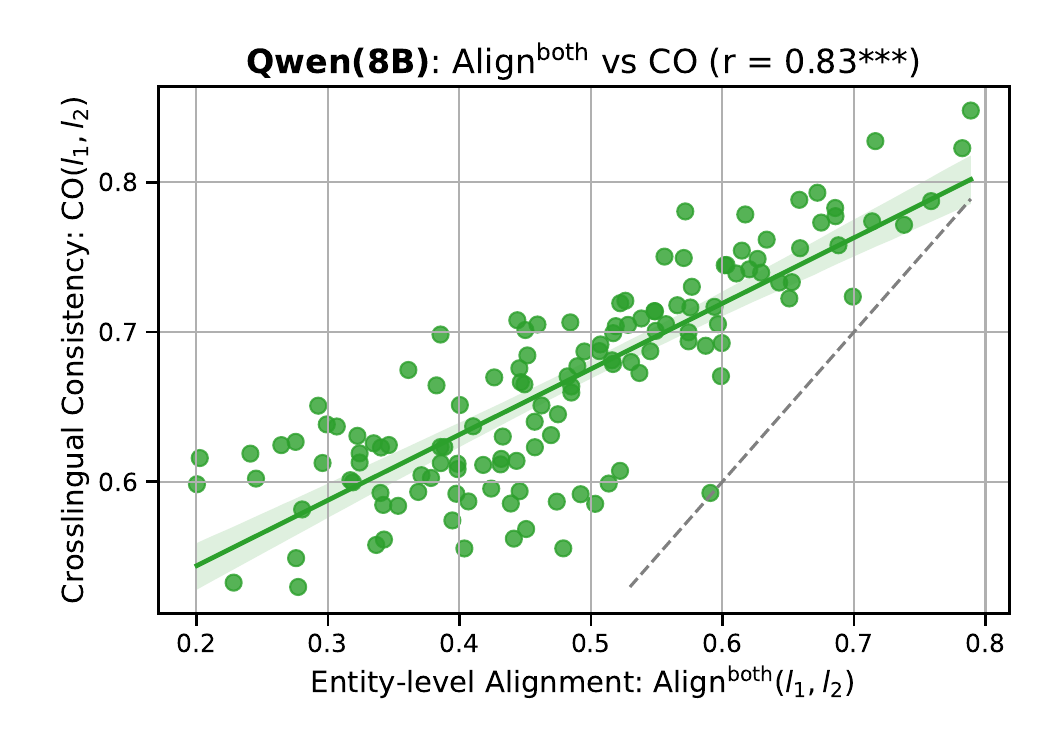}
    \caption{Correlation between entity-level alignment and crosslingual consistency. 
    Each subplot displays the relationship between one alignment metric, i.e., \blue{$\mathrm{Align}^{\mathrm{sub}}(l_1, l_2)$}, \orange{$\mathrm{Align}^{\mathrm{obj}}(l_1, l_2)$}, or \green{$\mathrm{Align}^{\mathrm{both}}(l_1, l_2)$}, and crosslingual consistency $\mathrm{CO}(l_1, l_2)$ for a given model. 
    Each point represents a language pair. 
    The \underline{gray dashed line} (the diagonal $y=x$)
    separates the region into one half where consistency is
    higher (above the line) and one half where alignment is
    higher (below the line).
    Strong and statistically significant correlations are
    observed across models, supporting our hypothesis that alignment is highly associated with consistency.}
    \label{fig:correlation_main}
\end{figure*}

\subsection{Consistency Correlated with Alignment}\seclabel{correlation}

We examine whether alignment is \emph{correlated} with consistency across language pairs in practice.
Specifically, we perform a correlation analysis between consistency $\mathrm{CO}(l_1, l_2)$ and each of the alignment metrics: $\mathrm{Align}^{\mathrm{sub}}(l_1, l_2)$, $\mathrm{Align}^{\mathrm{obj}}(l_1, l_2)$, and $\mathrm{Align}^{\mathrm{both}}(l_1, l_2)$.
For each model, we compute Pearson correlation coefficients over the 136 language pairs (cf. Figure~\ref{fig:correlation_main}).

\textbf{We observe strong and statistically significant correlations between consistency and alignment across models.} 
This result supports our hypothesis that crosslingual consistency is tightly linked to the model's ability to align entities across languages.
In all cases, the Pearson correlation is quite high, exceeding 0.7 for subject/both alignment and 0.5 for object alignment, significant at $p < 0.001$.
Notably, while high alignment does not always guarantee high consistency -- particularly in the case of object alignment (e.g., LLaMA (1B)) -- we rarely observe the opposite: high consistency with low alignment score.
This asymmetry suggests that alignment is closely associated with, and may be a prerequisite for, crosslingual consistency, though it is not by itself sufficient -- a pattern we examine further in \secref{groundness}. 

\textbf{Among the three alignment metrics, subject and joint alignment correlate more strongly with consistency than object alignment.} 
This asymmetry may reflect the task structure of factual recall: the subject entity is always present in the prompt and serves as the anchor for factual prediction.
If the model fails to align the subject of a fact across two languages in the first place, it is less likely to consistently recall facts containing the subject. 
Therefore, subject alignment may play an important role in facilitating consistency.
Based on this insight, we present two simple remedies for improving the factual recall in \secref{remedies}.

\textbf{OLMo exhibits weaker correlation, particularly with object-level alignment.}
Although OLMo still shows statistically significant correlations, they are generally weaker than those of multilingual models like Qwen or LLaMA. 
A likely cause is its limited exposure to factual content in non-English languages during pretraining \citep{liu2025tracingmultilingualfactualknowledge}, which impairs both factual recall and entity alignment. 
Additionally, English-centric models often rely on English as a pivot language \citep{wendler-etal-2024-llamas} in the language-agnostic conceptual space, but the prompts do not provide any explicit English signal, weakening the model's ability to bridge language-specific inputs and \emph{correct} concepts. 
As shown in \secref{remedies}, injecting English subjects into prompts can mitigate this limitation and substantially improve factual recall.

\begin{figure}[t]
  \centering
    \setlength{\abovecaptionskip}{-0.01cm}
    \setlength{\belowcaptionskip}{-0.4cm}
  \includegraphics[width=0.46\textwidth]{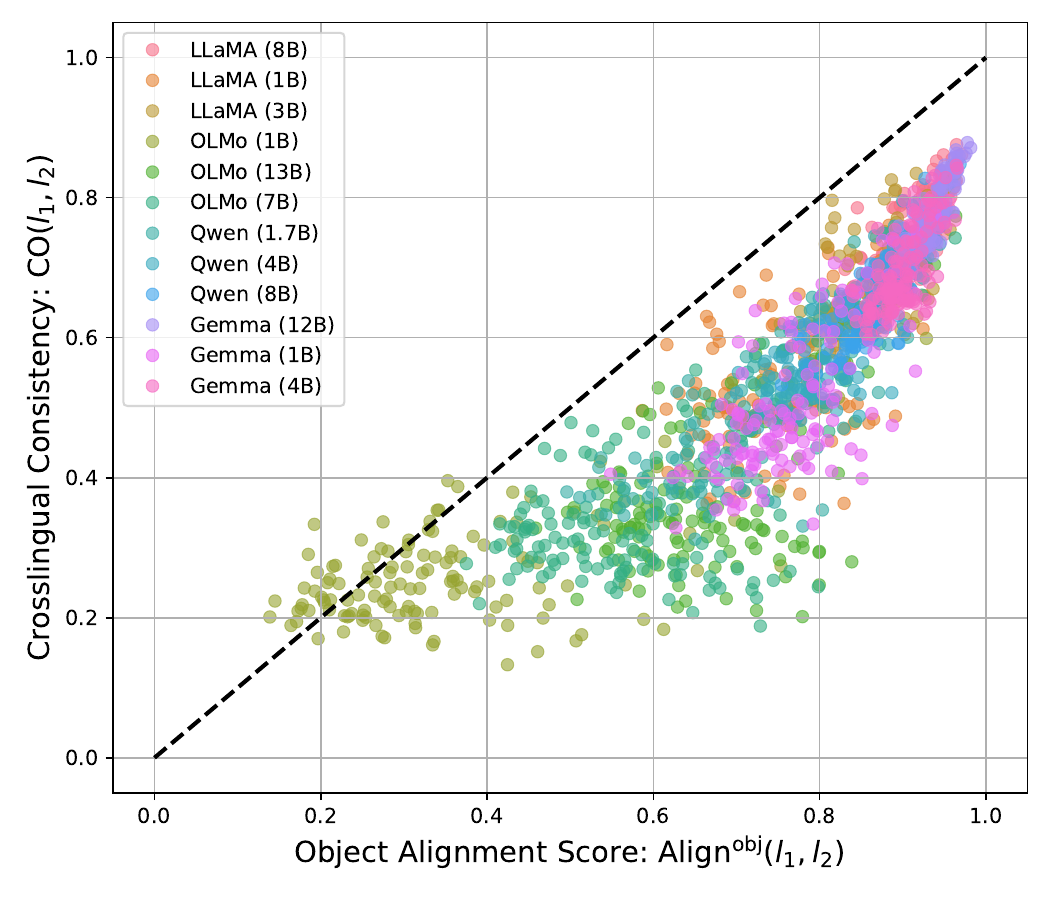}
  \caption{The boundedness relationship between consistency $\mathrm{CO}({l_1, l_2})$ and object alignment score $\mathrm{Align}^{\text{obj}}(l_1, l_2)$. 
  Each point indicates a language pair, while different colors indicate different models.
  $\mathrm{CO}({l_1, l_2})$ is almost always upper-bounded by $\mathrm{Align}^{\text{obj}}(l_1, l_2)$ except for OLMo (1B).}
  \label{fig:bounded}
\end{figure}

\textbf{Consistency is bounded by object alignment.}
We observe (Figure~\ref{fig:correlation_main}) that crosslingual consistency is almost always lower than object alignment scores across language pairs.
This is made explicit in Figure~\ref{fig:bounded}, which overlays consistency against object alignment scores for all language pairs across models. 
The vast majority of points lie below the diagonal, indicating that a model may not consistently recall a fact across two languages if it fails to align the object entity.
This pattern likely arises because both consistent factual recall and entity translation require the model to generate the correct object across languages.
The only exception is OLMo (1B), possibly due to its English-centric training data and weaker multilingual grounding, as discussed earlier.

Overall, these findings reinforce our core claim that good entity (subject and object) alignment facilitates crosslingual consistency. 


\subsection{Consistent Facts Are Entity-Aligned}\seclabel{groundness}

\begin{table*}[h]
\centering
\setlength{\belowcaptionskip}{-0.4cm}
\footnotesize
\resizebox{\textwidth}{!}{
\setlength{\tabcolsep}{1.5pt}
\begin{tabular}{lrrrrrrrrrrrr}
\toprule
  & Gemma (1B) & Gemma (4B) & Gemma (12B) & LLaMA (1B) & LLaMA (3B) & LLaMA (8B) & OLMo (1B) & OLMo (7B) & OLMo (13B) & Qwen (1.7B) & Qwen (4B) & Qwen (8B) \\
\midrule
$\mathcal{C}(l_1, l_2)$
&
124K (0.35) & 194K (0.54) & 229K (0.64) & 134K (0.38) & 188K (0.53) & 215K (0.60) &
42K (0.12) & 72K (0.20) & 87K (0.24) & 113K (0.32) & 159K (0.45) & 186K (0.52) \\
$\mathcal{I}(l_1, l_2)$
&
231K (0.65) & 162K (0.46) & 127K (0.36) & 222K (0.62) & 169K (0.47) & 142K (0.40) &
314K (0.88) & 284K (0.80) & 269K (0.76) & 243K (0.68) & 197K (0.55) & 171K (0.48) \\
$\mathcal{A}(l_1, l_2)$
&
335K (0.94) & 353K (0.99) & 355K (1.00) & 335K (0.94) & 350K (0.98) & 354K (0.99) &
211K (0.59) & 293K (0.82) & 315K (0.88) & 323K (0.91) & 344K (0.97) & 350K (0.98) \\
$\mathcal{N}(l_1, l_2)$
&
21K (0.06) & 3K (0.01) & 1K (0.00) & 22K (0.06) & 6K (0.02) & 2K (0.01) &
145K (0.41) & 64K (0.18) & 41K (0.12) & 33K (0.09) & 12K (0.03) & 6K (0.02) \\
\bottomrule
\end{tabular}
}
\caption{
Proportion of consistent ($\mathcal{C}(l_1, l_2)$), inconsistent ($\mathcal{I}(l_1, l_2)$), aligned ($\mathcal{A}(l_1, l_2)$), and non-aligned crosslingual facts ($\mathcal{N}(l_1, l_2)$) for each model.
Each fact in 2,619 factual queries is evaluated over all \(\binom{17}{2} = 136\) language pairs, 
resulting in a total of \(2619 \times 136 = 356{,}184\) evaluated facts per model.
}
\label{tab:proportion}
\end{table*}

We have shown that entity alignment and crosslingual consistency are strongly correlated at the aggregate level.
In this section, we aim to deepen our understanding of their relationship by analyzing individual fact instances.
Specifically, we ask two questions:
(1) \textit{To what extent does entity alignment support crosslingual consistency?}, and
(2) \textit{Can non-aligned facts still be consistently recalled across languages?}
To answer these questions, we examine each fact 
and determine whether it is crosslingually consistent or
entity-aligned.\footnote{Throughout this section, for
simplicity, a fact refers to $(\{l_1, l_2\}, f_i)$, i.e., a fact with respect to a given language pair.}
This fine-grained analysis allows us to characterize the role of entity alignment in enabling consistency, and to assess how often consistency emerges even in the absence of entity alignment.

\begin{figure}[t]
  \centering
    \setlength{\abovecaptionskip}{-0.01cm}
    \setlength{\belowcaptionskip}{-0.4cm}
  \includegraphics[width=0.46\textwidth]{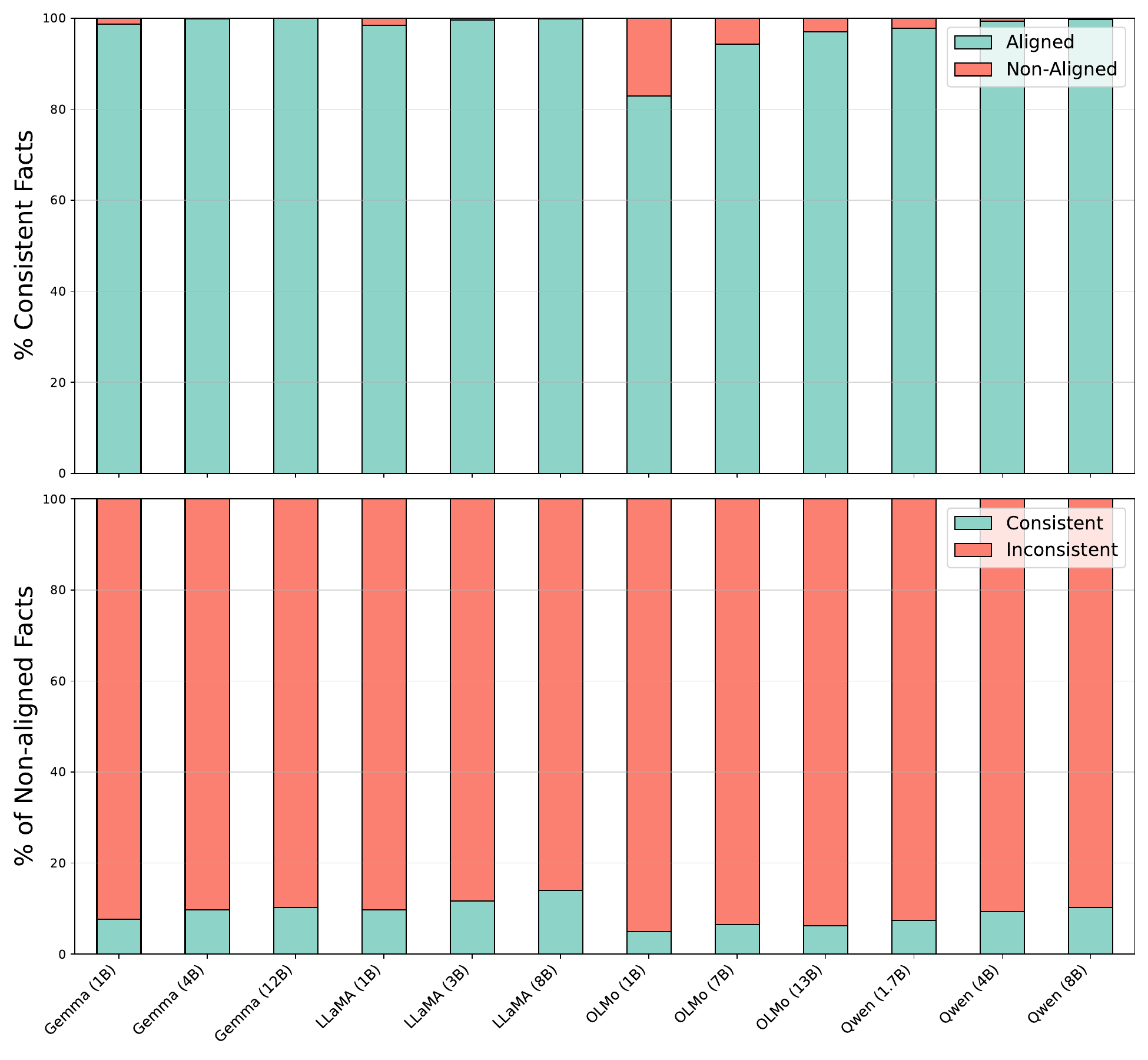}
  \caption{Entity alignment and consistency analysis across models.
    Most consistent facts are entity-aligned, and misalignment leads to inconsistency.
    }
  \label{fig:grounded-clc-barplot}
\end{figure}

We say that a fact $f_i$ is crosslingually
\textbf{\emph{consistent}} for languages $l_1$ and $l_2$ if
the model generates outputs that contain the expected
answers in both languages. Formally, this mirrors the
definition used in our consistency metric
(cf.\ \secref{fact_recall_evaluation}):
\enote{hs}{do you have to motivate this definition? my
  understanding is that a fact is aligned if at least one of
  the four possible translations is correct. why do you
  define mathcal A this way?}
\enote{yl}{yes, a fact is defined to be aligned if at least one of the four possible translations is correct. The intuition of this "loose" definition (as opposed to requiring every translation to be correct) is that we could account for some model's inability because of low-resource languages -- for example, even if the subject in l1 and l2 is aligned in the conceptual space, because of the limited capability, the model might be able to translate in one direction (e.g., l1 -> l2) while fails on the other direction (e.g., l2 -> l1).}
\[
\mathcal{C}(l_1, l_2) = \left\{ f_i \ \middle| \ o_i^{l_1} \subseteq \mathcal{M}(q_i^{l_1}) \ \land\  o_i^{l_2} \subseteq \mathcal{M}(q_i^{l_2}) \right\}
\]
Similarly, we define crosslingually \textbf{\emph{inconsistent}} facts for language $l_1$ and $l_2$ as the complement of $\mathcal{C}(l_1, l_2)$, which is referred to as $\mathcal{I}(l_1, l_2)$.

We further categorize each fact $f_i$ based on whether its subject or object entities are aligned between two languages, i.e., whether the entities can be correctly translated between $l_1$ and $l_2$. We define the set of \textbf{\emph{aligned}} facts as:
\[
\begin{aligned}
\mathcal{A}(l_1, l_2) = \{ f_i\ | \
& (s_i^{l_2} \subseteq \mathcal{M}(s_i^{l_1}) \lor 
   s_i^{l_1} \subseteq \mathcal{M}(s_i^{l_2})) \\
\lor\ 
& (o_i^{l_2} \subseteq \mathcal{M}(o_i^{l_1}) \lor 
   o_i^{l_1} \subseteq \mathcal{M}(o_i^{l_2}))
\}
\end{aligned}
\]

That is, a fact is considered aligned if at least one of the four translation tasks succeeds.\footnote{Here we use \emph{alignment} at the \emph{instance level}, i.e., whether the subject or object entity of a given fact can be correctly translated between $l_1$ and $l_2$. 
This differs from the usage in \secref{correlation}, where alignment refers to \emph{aggregated alignment scores} over translation performance across all facts.}
This relaxed formulation reflects two intuitions: 
(1) correct alignment of either the subject or object may already provide a strong conceptual anchor to access the underlying fact and 
(2) asymmetric translations (e.g., correct for $l_1 \to l_2$ but not for $l_2 \to l_1$) are common, especially when involving low-resource languages.
The set of \textbf{\emph{non-aligned}} facts $\mathcal{N}(l_1, l_2)$ is then the complement of $\mathcal{A}(l_1, l_2)$.

We report the proportion of consistent facts $\mathcal{C}(l_1, l_2)$, inconsistent facts $\mathcal{I}(l_1, l_2)$, aligned facts $\mathcal{A}(l_1, l_2)$, and non-aligned facts $\mathcal{N}(l_1, l_2)$ for each model in Table~\ref{tab:proportion}.
Figure~\ref{fig:grounded-clc-barplot} provides a more detailed breakdown:
(1) among all consistent facts $\mathcal{C}(l_1, l_2)$, what fraction are aligned ($\mathcal{A}(l_1, l_2)$) vs.\  non-aligned ($\mathcal{N}(l_1, l_2)$), and
(2) among all non-aligned facts $\mathcal{N}(l_1, l_2)$, what proportion are consistent ($\mathcal{C}(l_1, l_2)$) vs.\  inconsistent ($\mathcal{I}(l_1, l_2)$).
Together, they illustrate how much consistency can be attributed to entity alignment, and whether models can still achieve consistency without it.

\textbf{Models exhibit stronger entity alignment than crosslingual consistency.}
As shown in Table~\ref{tab:proportion}, nearly all models achieve high rates of entity alignment across language pairs, often aligning over 95\% of facts. 
For example, Gemma (12B) and LLaMA (8B) align more than 99\% of all evaluated crosslingual facts.
In contrast, their consistency rates remain notably lower (64\% and 60\%, respectively). 
This discrepancy suggests that entity alignment is a foundational skill models acquire more easily during pretraining, whereas achieving consistency demands additional capabilities such as relation representation and knowledge retrieval (required in the processing stage in Figure~\ref{fig:conceptual_model}).
With a larger model size, such capabilities are enhanced; therefore, the consistency is substantially improved (e.g., from 0.35 to 0.64 when comparing Gemma (1B) and Gemma (12B)).
The OLMo family, in particular, shows weak alignment and consistency, likely due to its English-centric pretraining data. 

\textbf{Crosslingual consistency is tightly conditioned on entity alignment.}
Figure~\ref{fig:grounded-clc-barplot} shows that the vast majority of consistent facts are entity-aligned, particularly for highly multilingual models like Gemma and Qwen, where over 99\% of consistent facts involve successful entity alignment (either the subject, object, or both). 
In contrast, consistent recall among non-aligned facts is
rare: fewer than 10\% of such cases achieve crosslingual
consistency, regardless of model family or
size.\footnote{These non-aligned but consistent facts may be
a form of ``memorization'', where the model directly memorizes the fact as a whole in the respective language instead of transferring the knowledge across languages through a shared conceptual space, as shown by \citet{liu2025tracingmultilingualfactualknowledge}.}
This asymmetry suggests that consistency across languages does not arise from chance or superficial pattern matching, but instead critically depends on the model's ability to semantically align entities. 
Entity alignment, therefore, acts as a prerequisite or bottleneck for achieving consistency. 
These findings underscore the importance of entity-level alignment and imply that enhancing alignment -- particularly for low-resource languages -- may directly improve a model's ability to maintain consistent factual knowledge across linguistic boundaries.

\begin{table*}[t]
\setlength{\abovecaptionskip}{-0.001cm}
\setlength{\belowcaptionskip}{-0.5cm}
\centering
\footnotesize
\setlength{\tabcolsep}{2.5pt}
\begin{tabular}{lrrrrr|rrrrr}
\toprule
\multirow{2}{*}{Model} & \multicolumn{5}{c}{Recall (ACC)} & \multicolumn{5}{c}{Consistency (CO)} \\
 & Base & \textsc{SubSub} & \textsc{SubInj} & $\uparrow_{\textsc{SubSub}}$ (\%) & $\uparrow_{\textsc{SubInj}}$ (\%) & Base & \textsc{SubSub} & \textsc{SubInj} & $\uparrow_{\textsc{SubSub}}$ (\%) & $\uparrow_{\textsc{SubInj}}$ (\%) \\
\midrule
Gemma (1B) & 0.51 & \underline{0.56} & \textbf{0.58} & 9.0 & 13.7 & 0.51 & \underline{0.56} & \textbf{0.60} & 10.7 & 17.3 \\
Gemma (4B) & 0.66 & \underline{0.71} & \textbf{0.72} & 6.5 & 8.3 & 0.70 & \underline{0.75} & \textbf{0.77} & 8.0 & 10.6 \\
Gemma (12B) & 0.73 & \underline{0.75} & \textbf{0.76} & 2.8 & 4.1 & 0.78 & \underline{0.81} & \textbf{0.83} & 4.4 & 6.2 \\
LLaMA (1B) & 0.53 & \underline{0.60} & \textbf{0.61} & 11.7 & 14.9 & 0.54 & \underline{0.61} & \textbf{0.63} & 13.2 & 17.8 \\
LLaMA (3B) & 0.65 & \underline{0.71} & \textbf{0.72} & 8.6 & 10.7 & 0.68 & \underline{0.75} & \textbf{0.77} & 11.1 & 14.3 \\
LLaMA (8B) & 0.71 & \underline{0.75} & \textbf{0.76} & 5.6 & 7.2 & 0.74 & \underline{0.80} & \textbf{0.82} & 7.7 & 10.0 \\
OLMo (1B) & 0.27 & \underline{0.28} & \textbf{0.31} & 2.4 & 11.2 & \underline{0.27} & {0.24} & \textbf{0.28} & -8.7 & 4.8 \\
OLMo (7B) & 0.38 & \underline{0.45} & \textbf{0.47} & 19.5 & 25.1 & 0.35 & \underline{0.43} & \textbf{0.46} & 23.3 & 31.6 \\
OLMo (13B) & 0.43 & \underline{0.54} & \textbf{0.56} & 25.9 & 31.6 & 0.39 & \underline{0.52} & \textbf{0.55} & 35.3 & 43.9 \\
Qwen (1.7B) & 0.48 & \underline{0.51} & \textbf{0.54} & 6.3 & 11.7 & 0.49 & \underline{0.52} & \textbf{0.56} & 7.3 & 14.5 \\
Qwen (4B) & 0.59 & \underline{0.63} & \textbf{0.64} & 5.9 & 8.4 & 0.60 & \underline{0.66} & \textbf{0.68} & 8.7 & 12.4 \\
Qwen (8B) & 0.65 & \underline{0.68} & \textbf{0.69} & 4.0 & 6.3 & 0.67 & \underline{0.71} & \textbf{0.73} & 6.0 & 9.2 \\
\bottomrule
\end{tabular}
\caption{
  Performance of  \textsc{SubSub} and \textsc{SubInj}
compared to Base. 
Both methods outperform the baseline in terms of factual recall (ACC) and crosslingual consistency (CO).
\textsc{SubInj} yields stronger improvements than \textsc{SubSub} across all models. 
Percentages indicate relative improvements over the baseline.
}
\label{tab:subinj_perf}
\end{table*}

\section{Remedies: \textsc{SubSub} and \textsc{SubInj}}\seclabel{remedies}

In \secref{relationship}, we show that crosslingual consistency is closely tied to a model’s ability to align entities across languages.
However, many LLMs, particularly English-centric ones, struggle with this alignment, reflecting weaknesses in the transition from language-specific surface forms to language-agnostic conceptual representations. 
This observation motivates a simple yet effective idea: if a model fails to map the subject into a shared conceptual space, why not provide it with a reliable lexical anchor?
Since most LLMs are extensively trained on English and often use English as an implicit pivot language in the conceptual space \citep{wendler-etal-2024-llamas,wang2025lostmultilingualitydissectingcrosslingual}, integrating English subject information can help the model bypass the subject mapping step and thereby improve factual recall across languages.

\subsection{Prompt Formulation}

We propose two prompting-based remedies to implement this idea. 
\textbf{\textsc{SubSub}} replaces the subject in the original-language prompt with its English equivalent, enforcing direct entity alignment.
This idea is similar to the substitution method proposed by \citet{yang-etal-2024-large-language-models} to facilitate the multi-hop reasoning in factual recall.
For example, we replace ``\begin{CJK}{UTF8}{min}フランス\end{CJK}'' with ``France'' and the final prompt in Japanese would then be ``{\begin{CJK}{UTF8}{min}Franceの首都はどこにありますか？答えは：\end{CJK}}'' (translation: \emph{Where is France’s capital located? The answer is:}).
  \textbf{\textsc{SubInj}} appends the English subject alongside the native-language form, preserving natural phrasing while encouraging the model to align the entities (e.g., ``{\begin{CJK}{UTF8}{min}フランス (France) の首都はどこにありますか？答えは：\end{CJK}}'').
The two proposed remedies align with recent ideas of leveraging \textit{code-switching} for better comprehension of the original text \citep{mohamed2025lostmixevaluatingllm}.

These remedies are conceptually related to the \emph{vector interventions} of \citet{lu2025pathstakenunderstandingmending}, who show that factual inconsistencies often arise because models fail to sufficiently engage their reliable \emph{English-centric} factual recall pipeline. 
By injecting carefully constructed vectors into intermediate layers, they steer models to re-activate these latent pathways, thereby improving factual recall across languages.
Our approach can be seen as a \emph{prompting-level analogue}: instead of modifying hidden states directly, we inject an explicit English signal into the input. 
This encourages the model to follow a similar trajectory as vector interventions -- activating English-centric recall mechanisms -- while remaining lightweight and training-free.


\subsection{Results and Discussion}

\begin{figure*}
    \centering
    \setlength{\abovecaptionskip}{-0.05cm}
    \setlength{\belowcaptionskip}{-0.4cm}
    \includegraphics[width=0.15\textwidth]{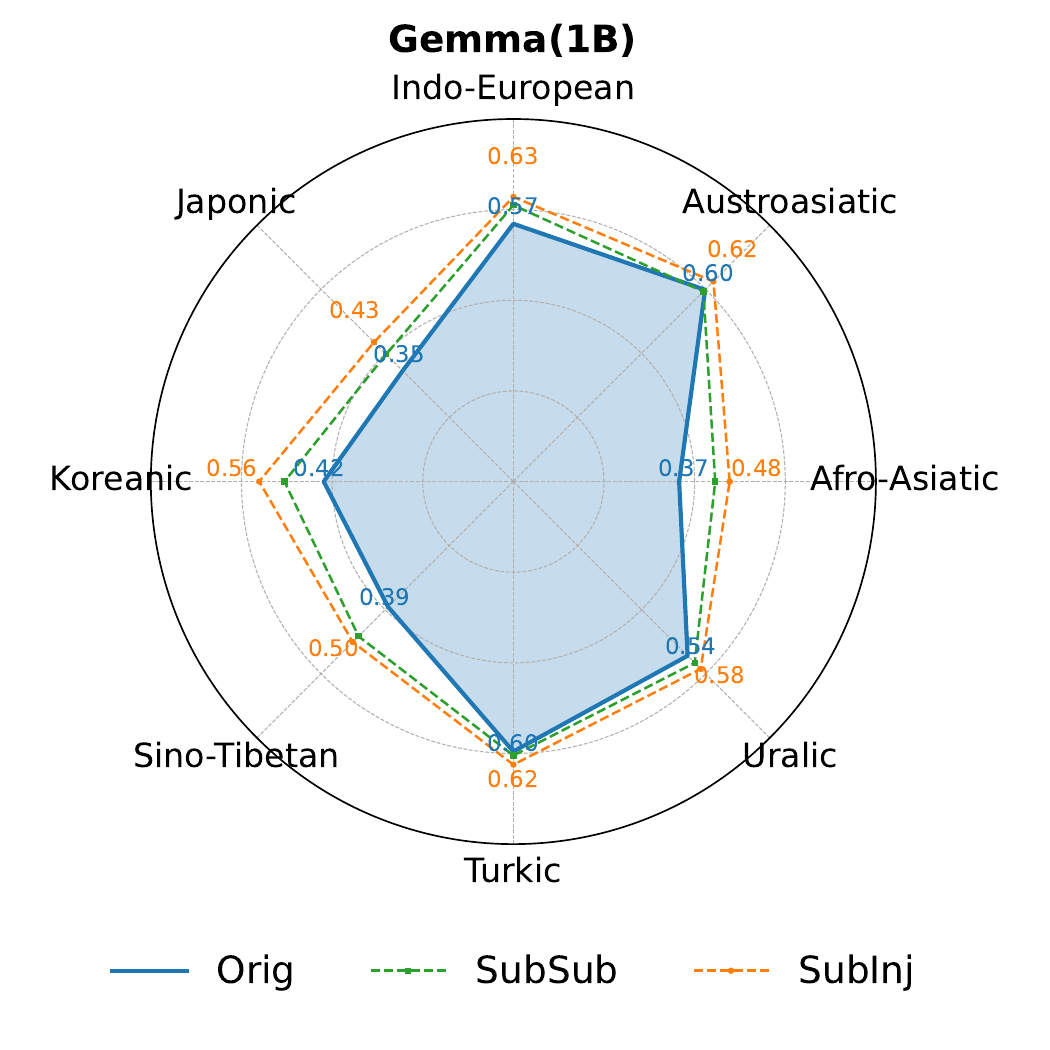}
    \includegraphics[width=0.15\textwidth]{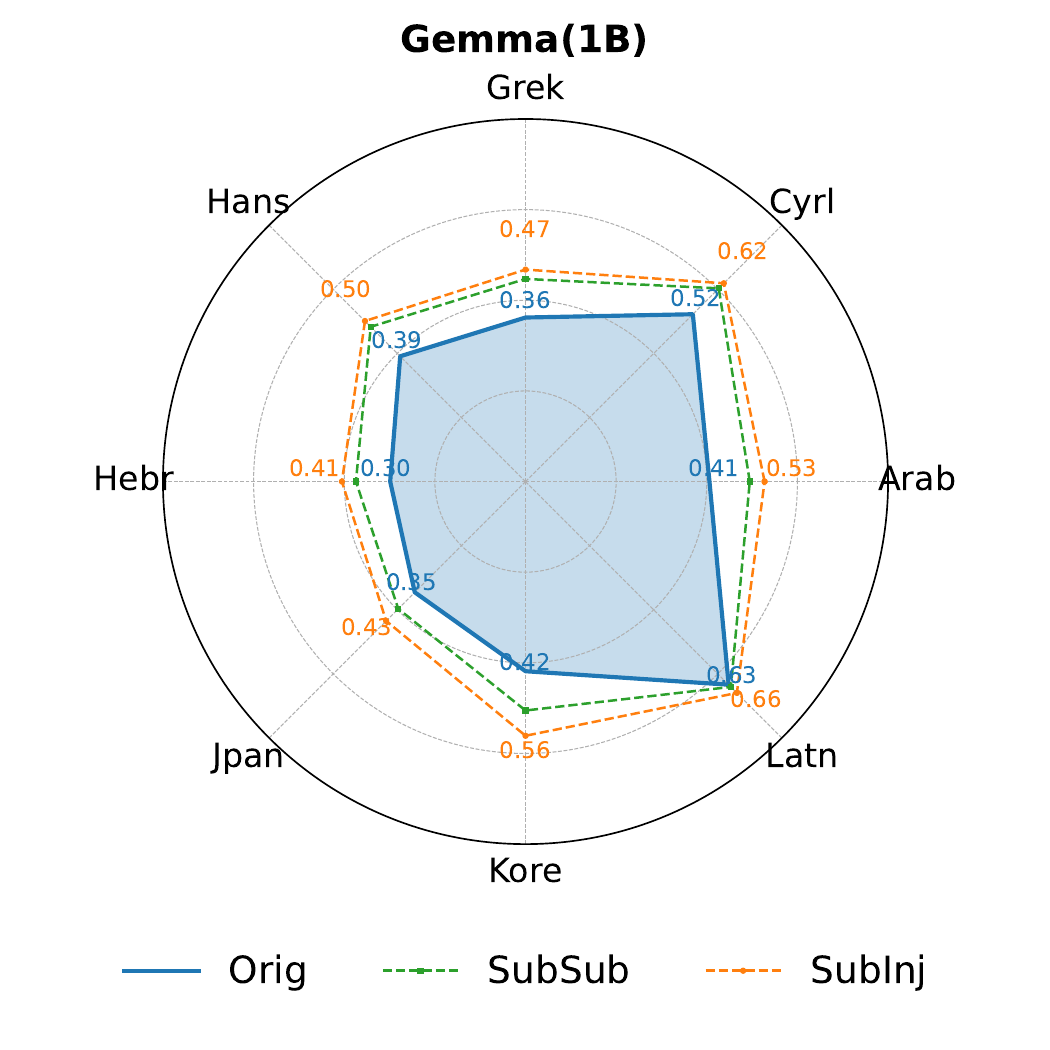}
    \includegraphics[width=0.15\textwidth]{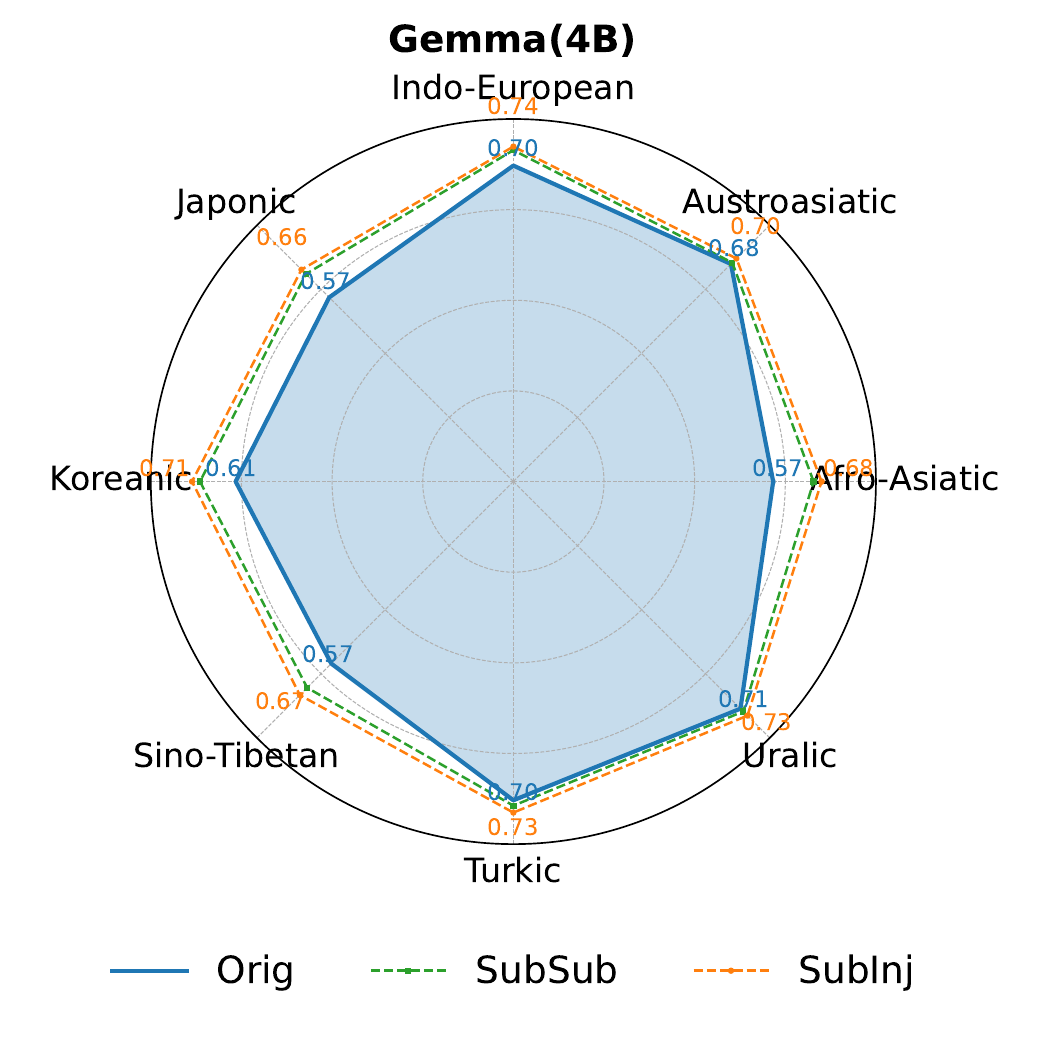}
    \includegraphics[width=0.15\textwidth]{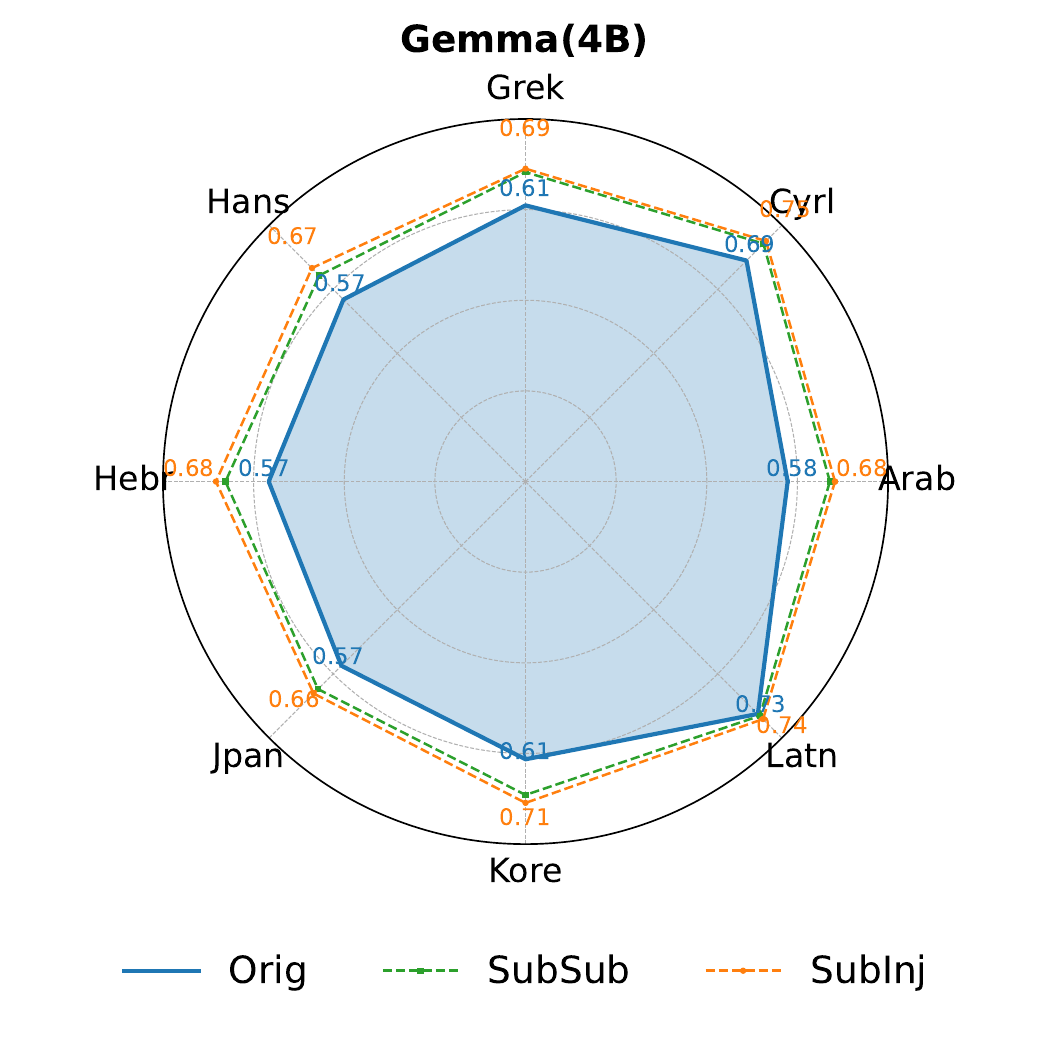}
    \includegraphics[width=0.15\textwidth]{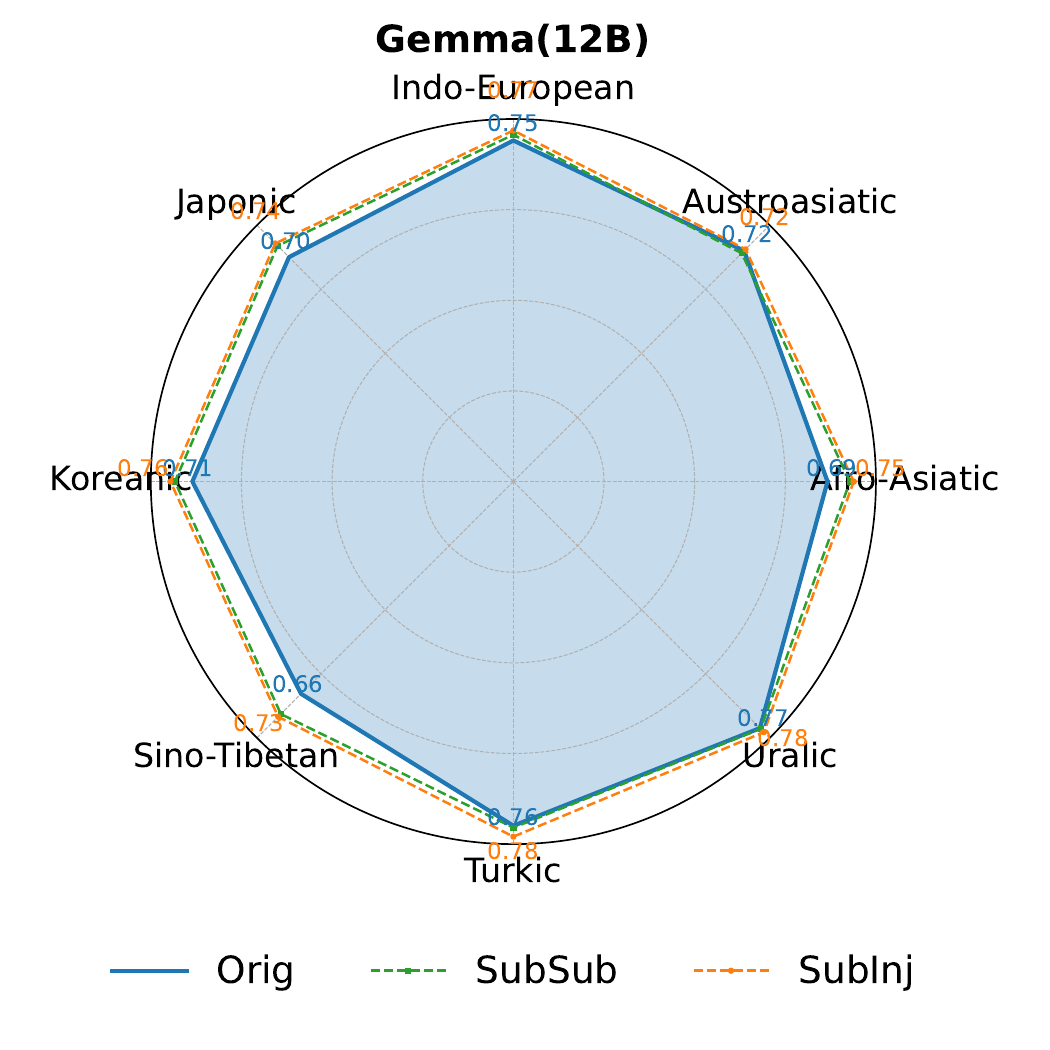}
    \includegraphics[width=0.15\textwidth]{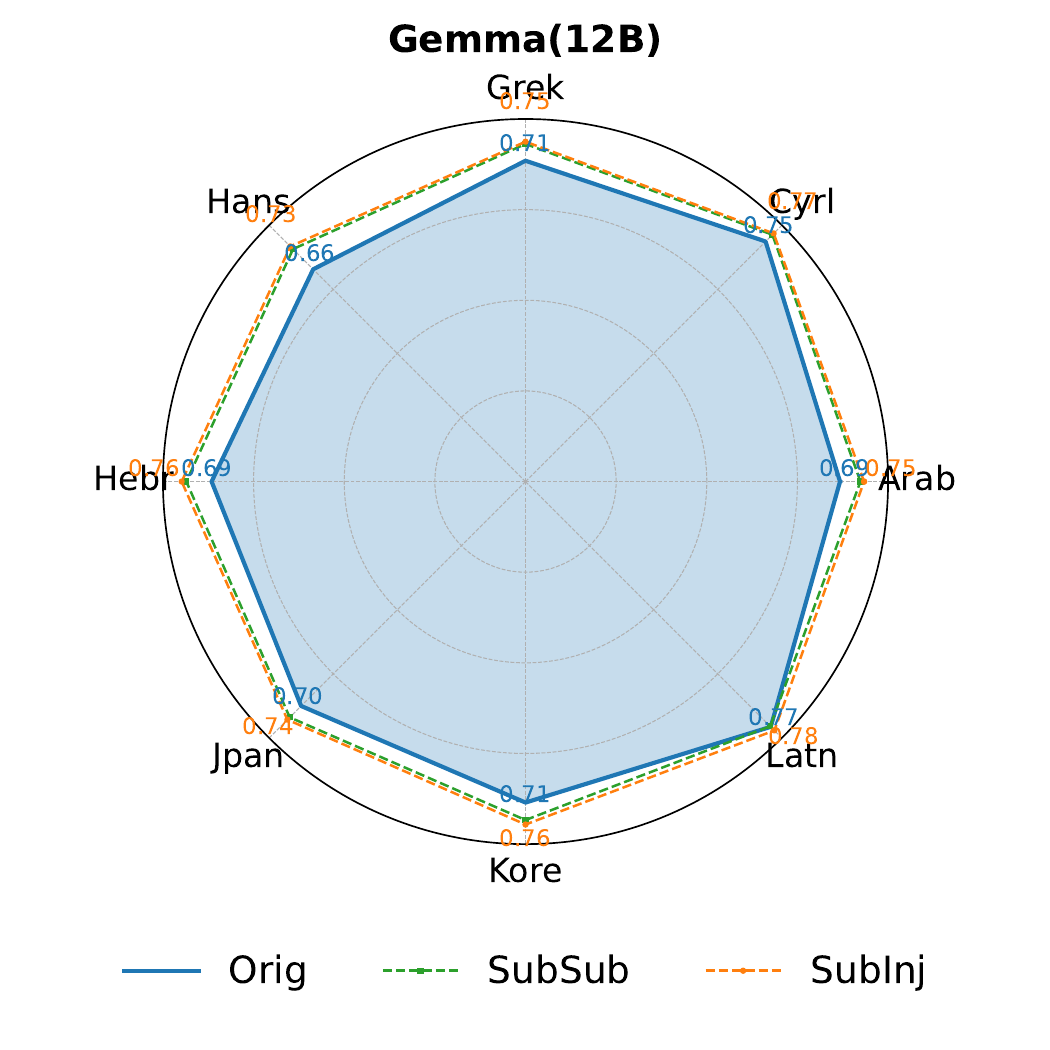}
    
    \includegraphics[width=0.15\textwidth]{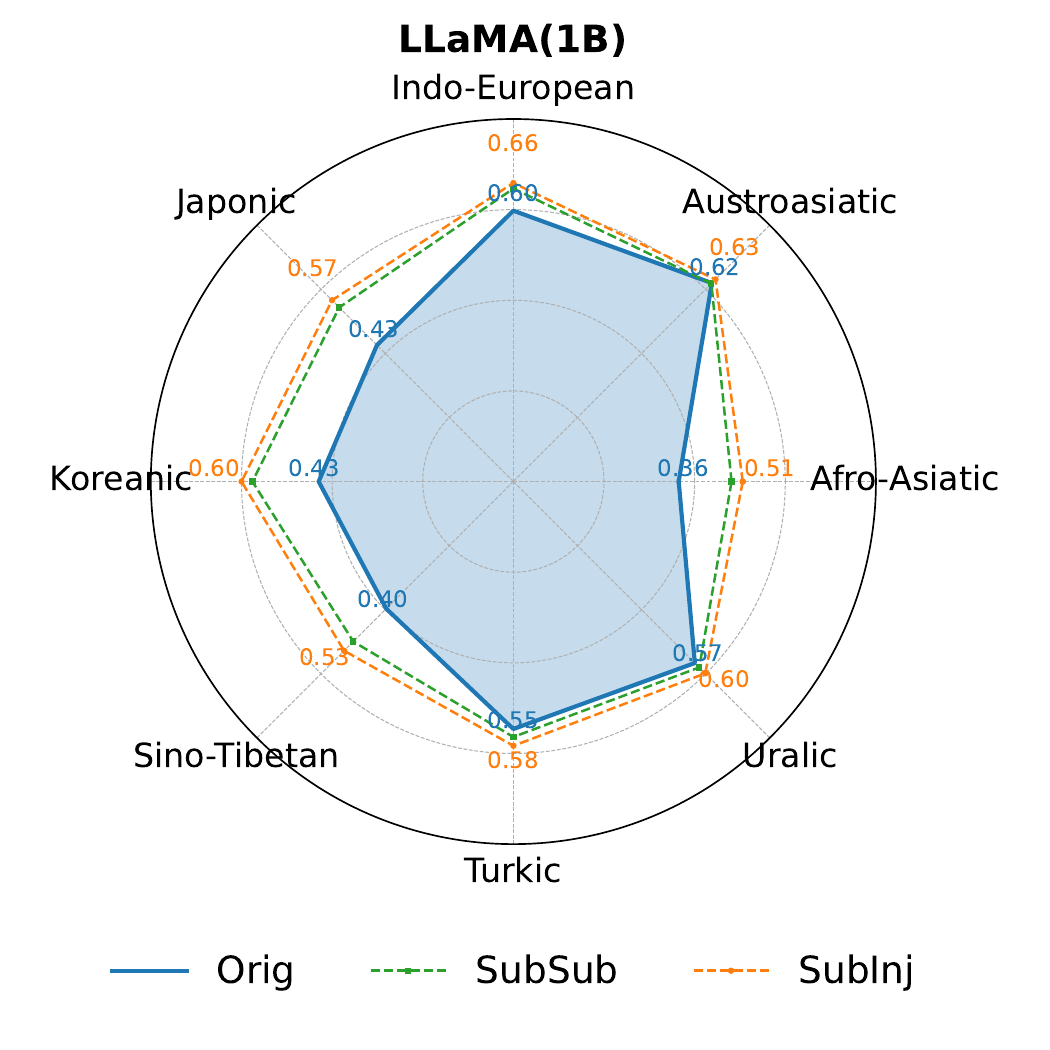}
    \includegraphics[width=0.15\textwidth]{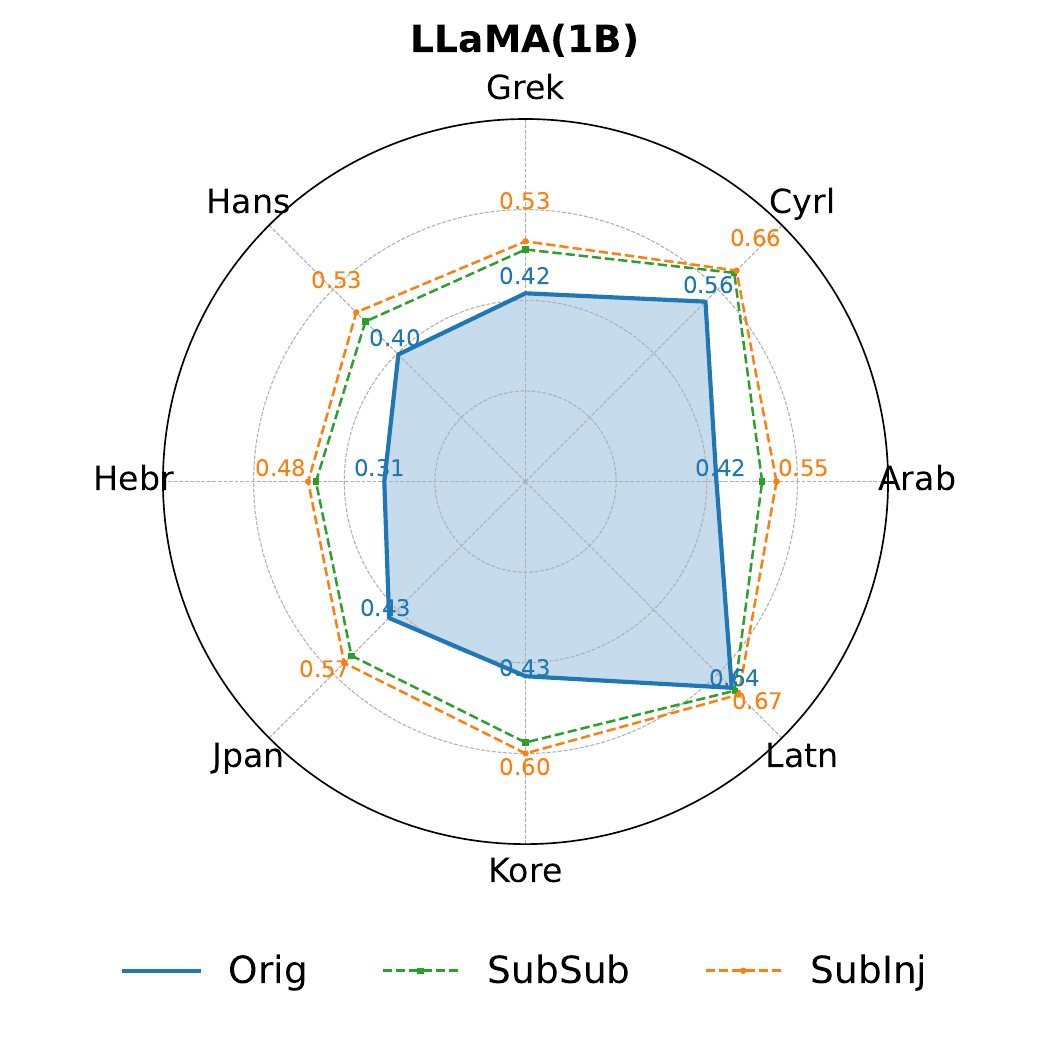}
    \includegraphics[width=0.15\textwidth]{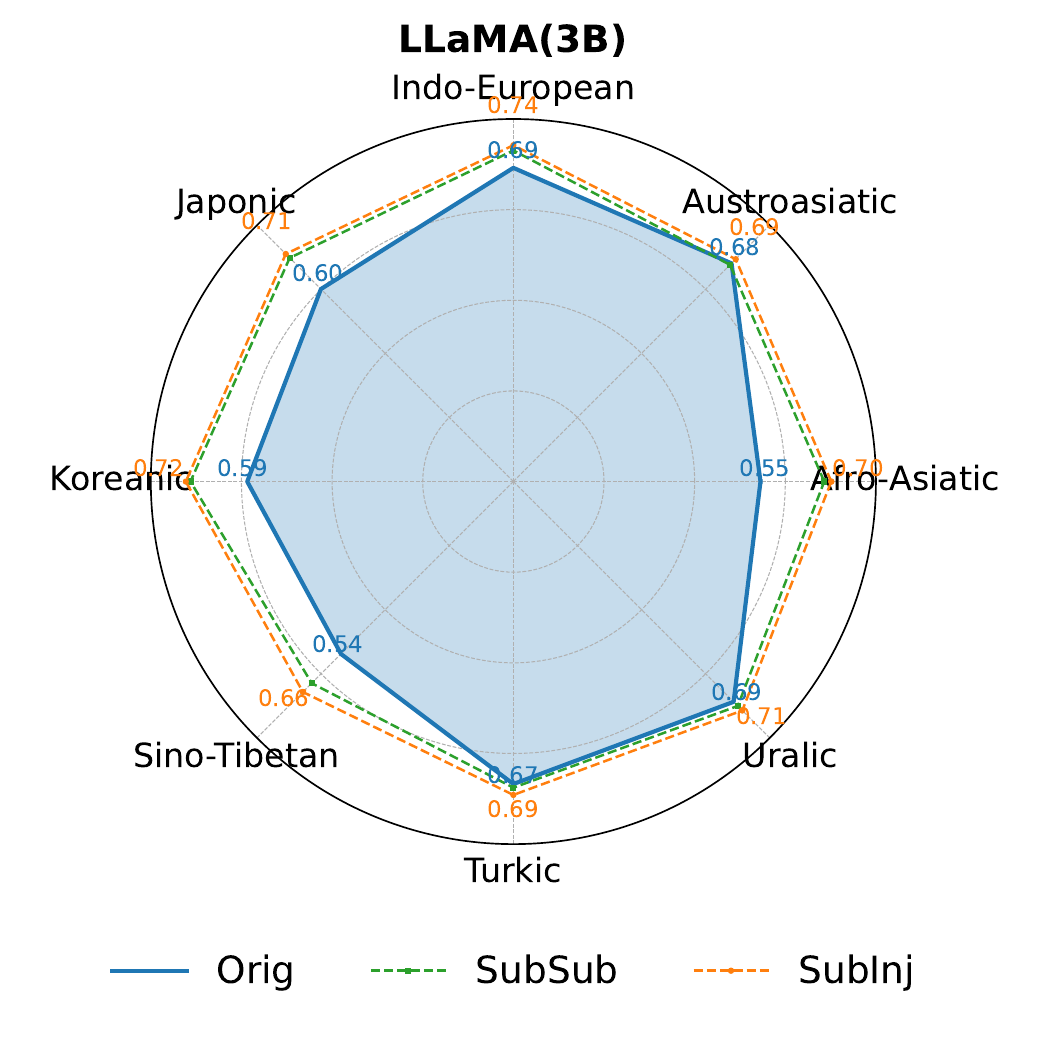}
    \includegraphics[width=0.15\textwidth]{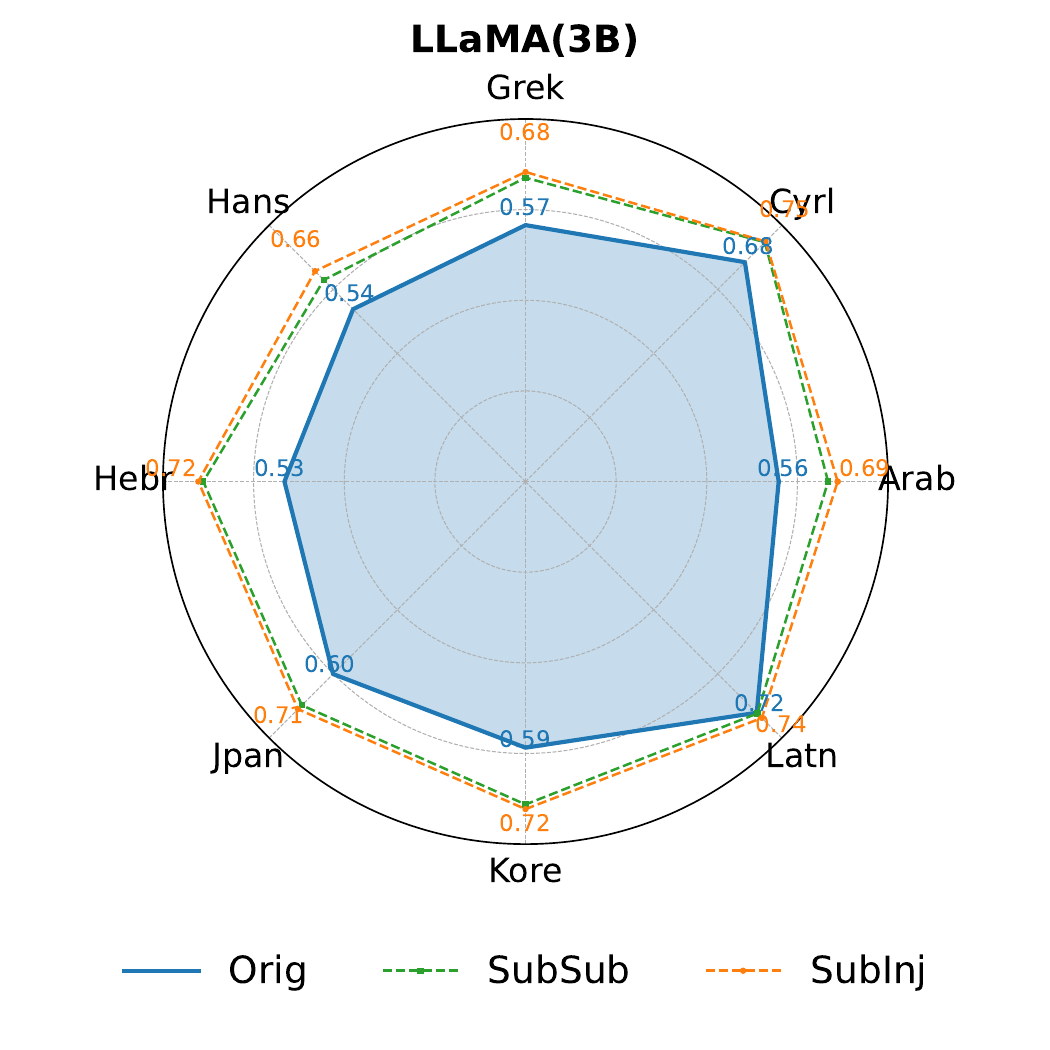}
    \includegraphics[width=0.15\textwidth]{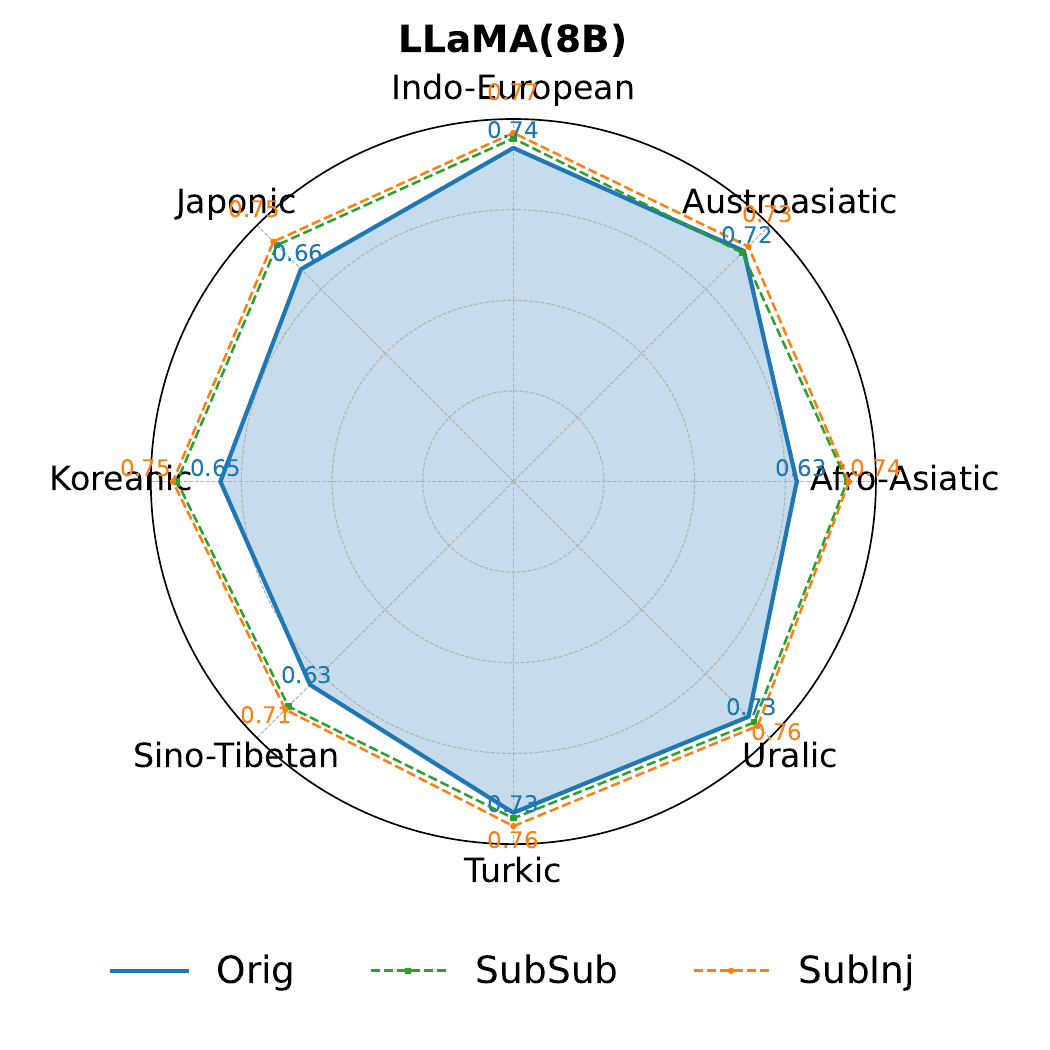}
    \includegraphics[width=0.15\textwidth]{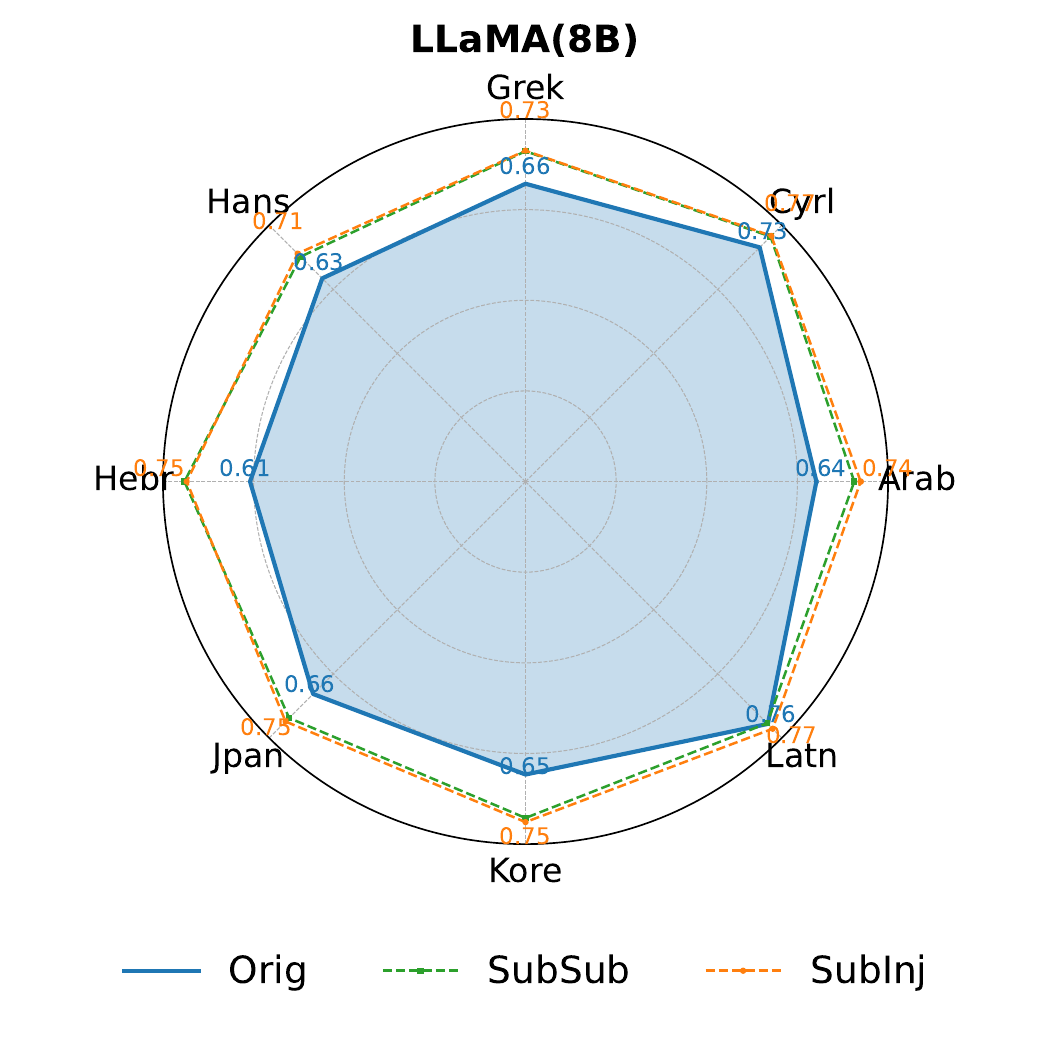}

    \includegraphics[width=0.15\textwidth]{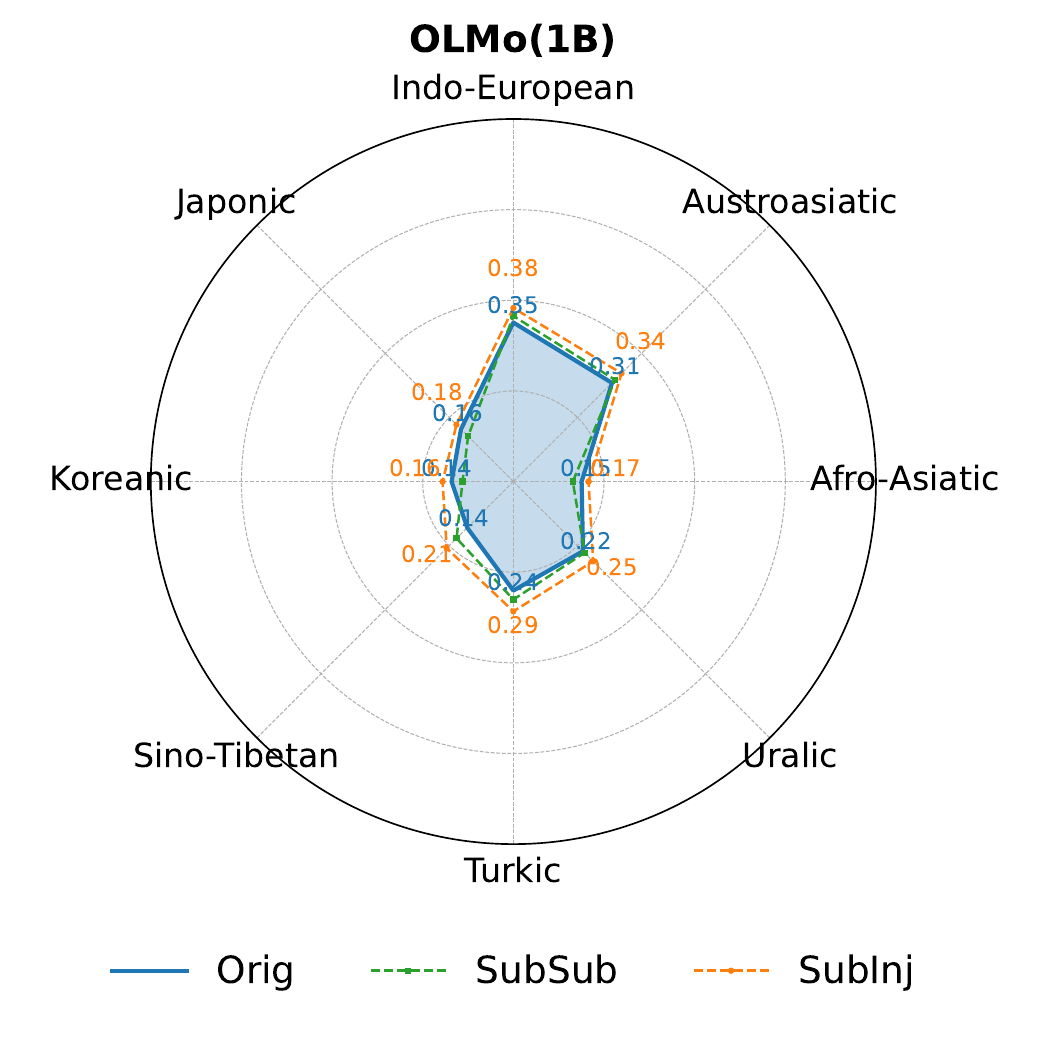}
    \includegraphics[width=0.15\textwidth]{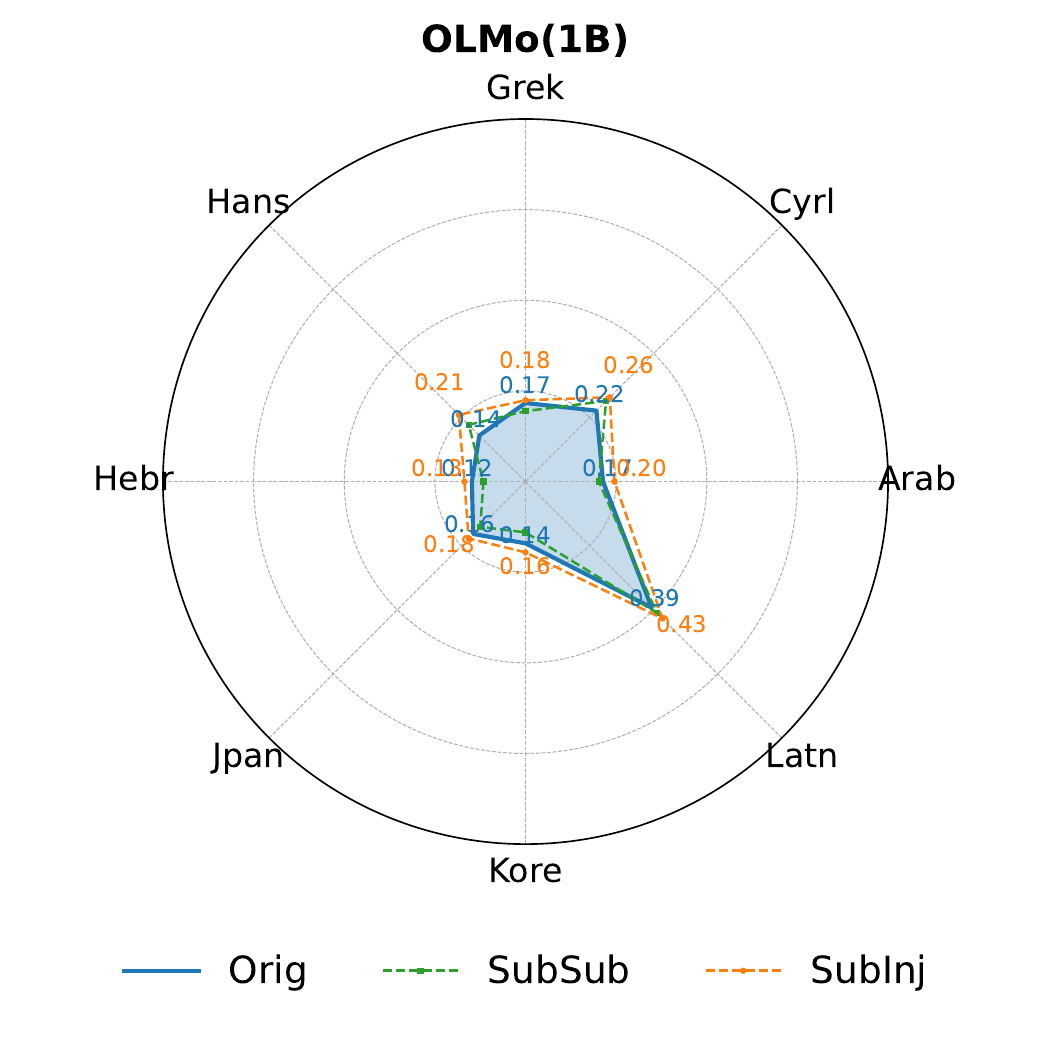}
    \includegraphics[width=0.15\textwidth]{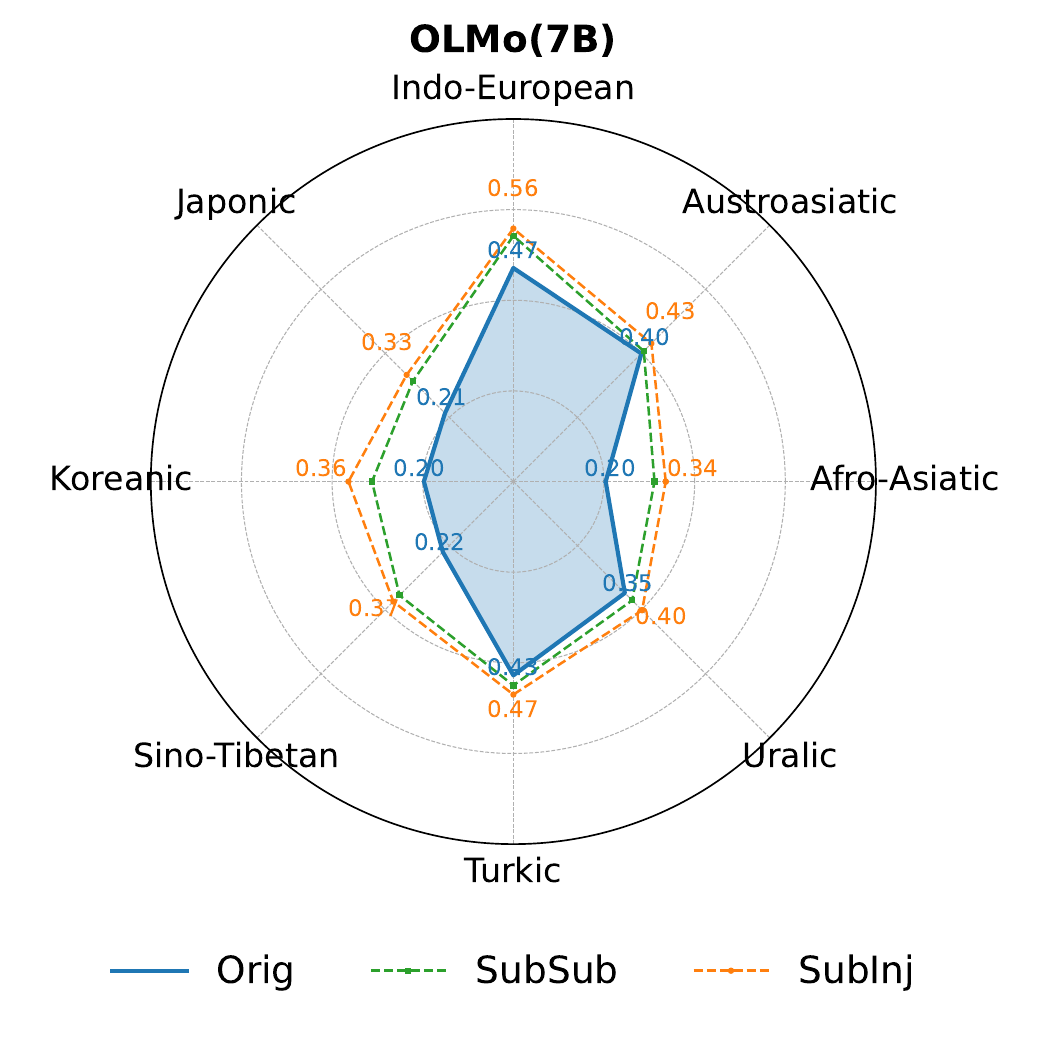}
    \includegraphics[width=0.15\textwidth]{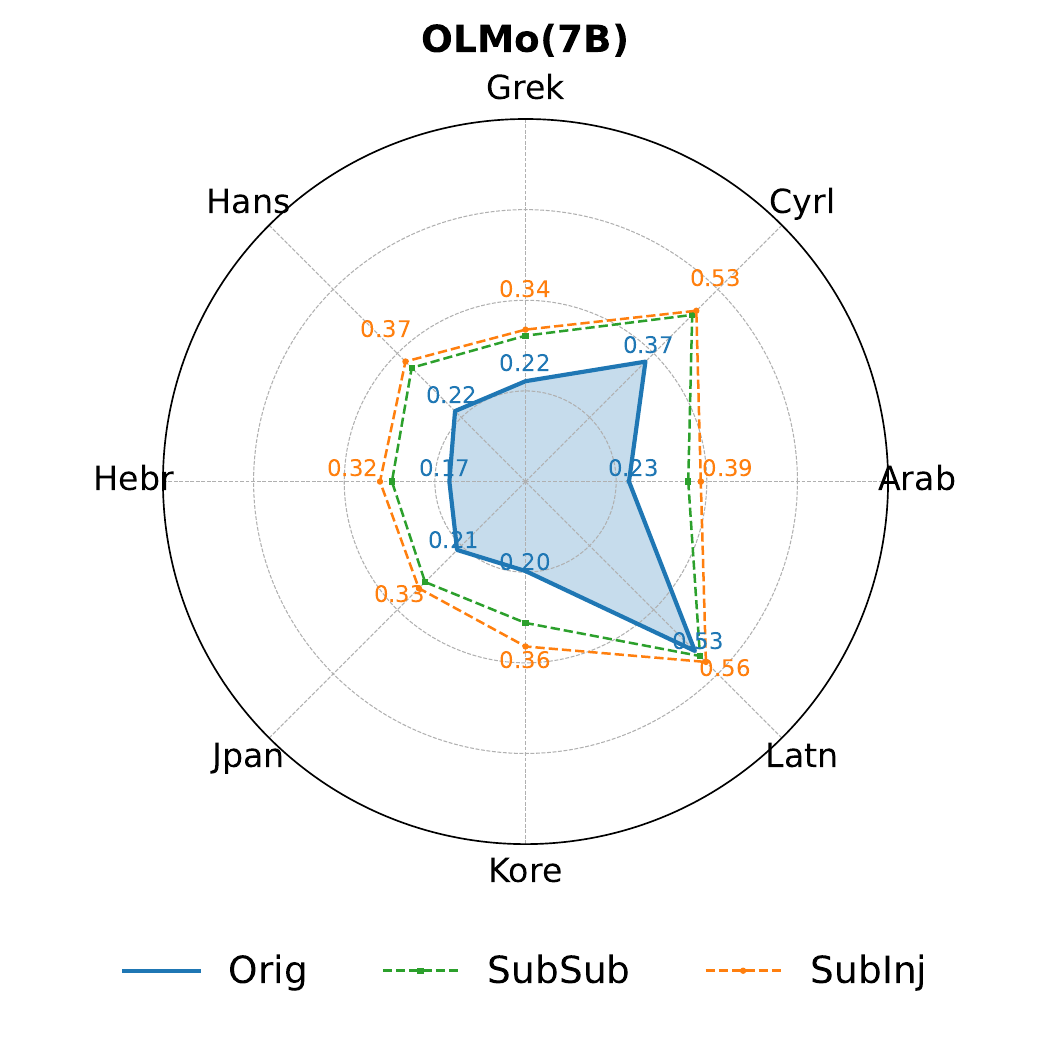}
    \includegraphics[width=0.15\textwidth]{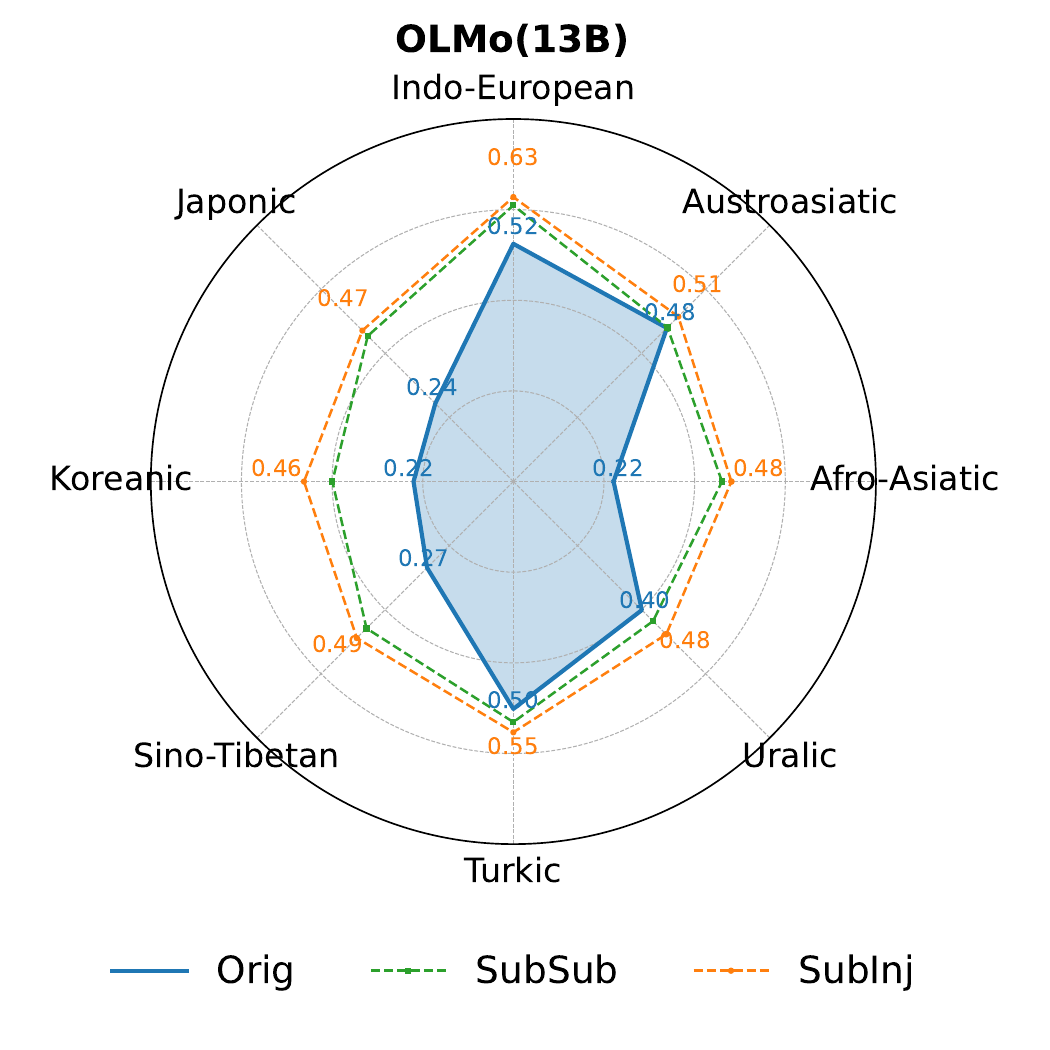}
    \includegraphics[width=0.15\textwidth]{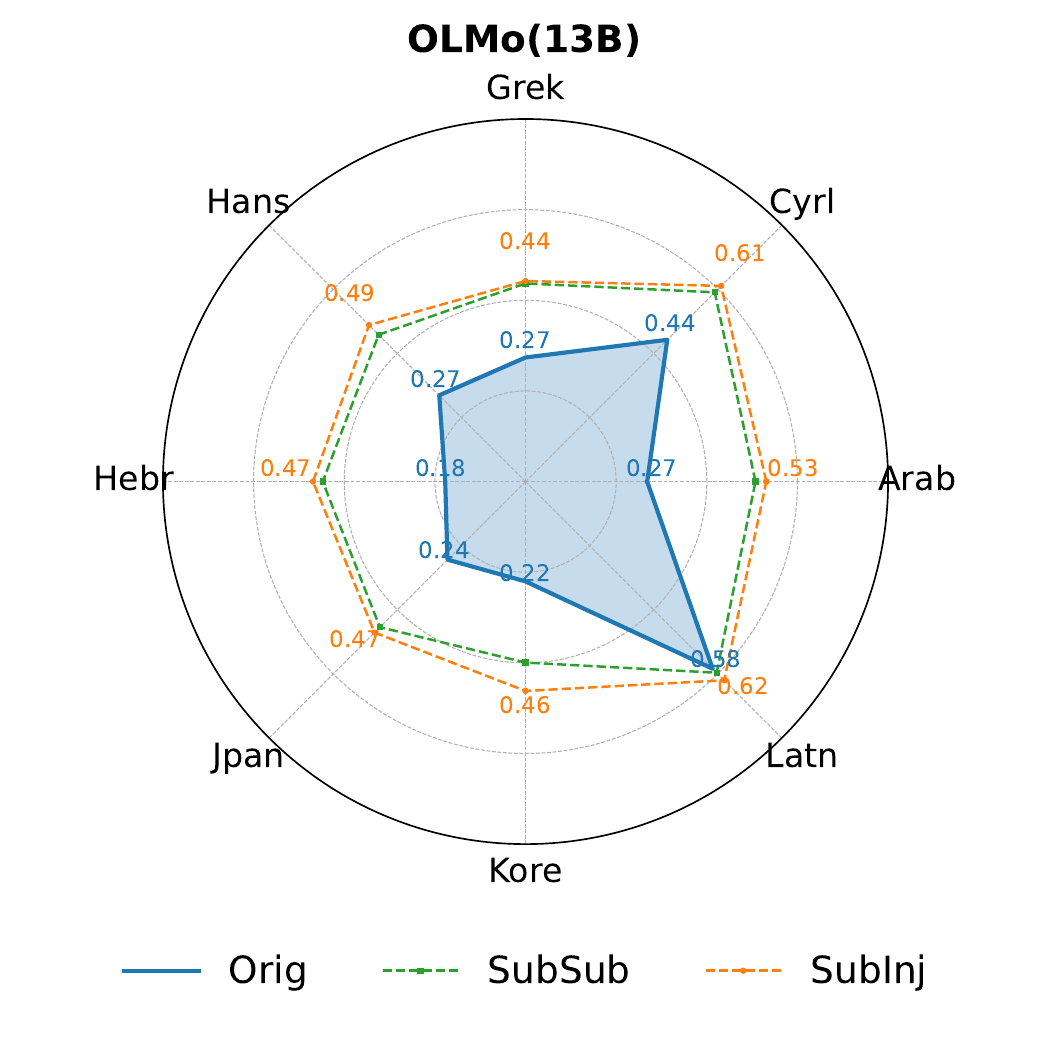}

    \includegraphics[width=0.15\textwidth]{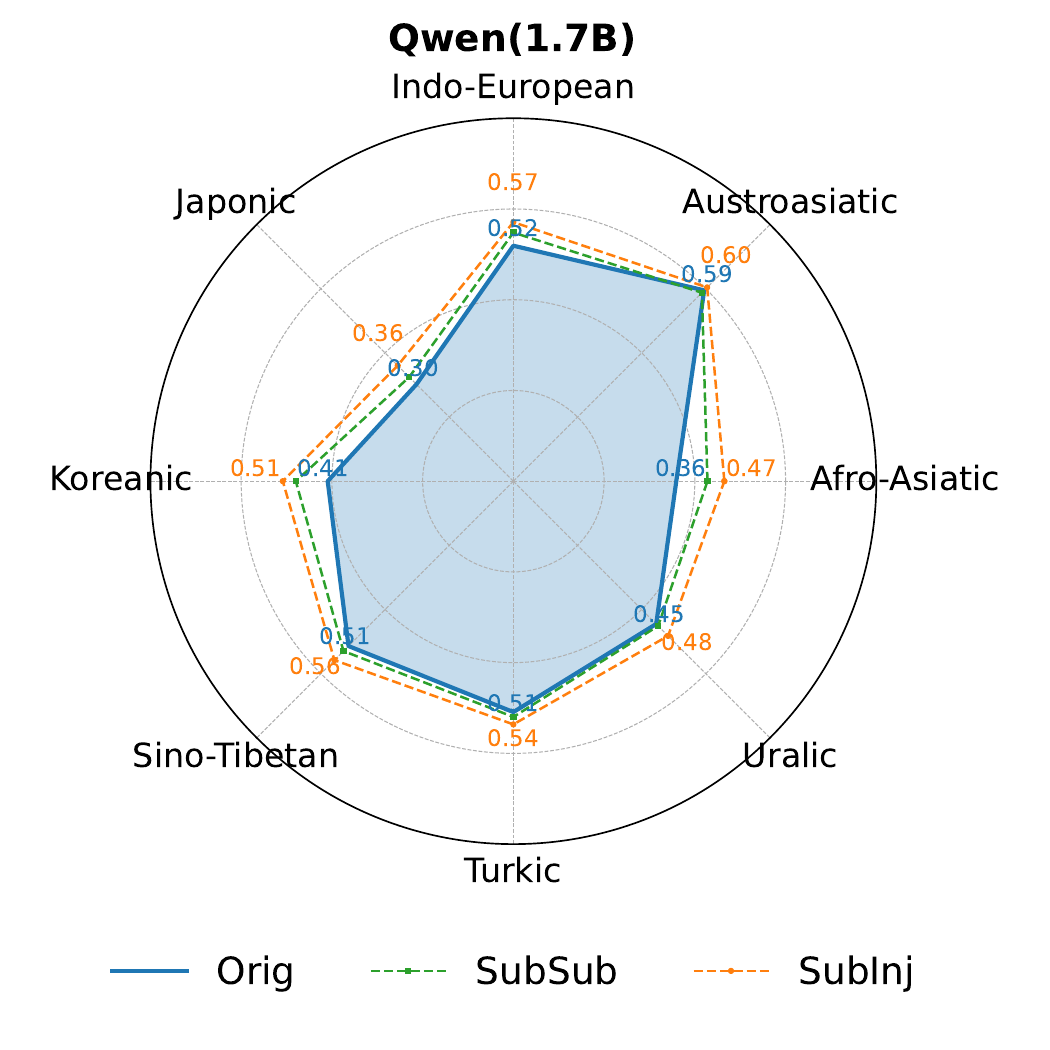}
    \includegraphics[width=0.15\textwidth]{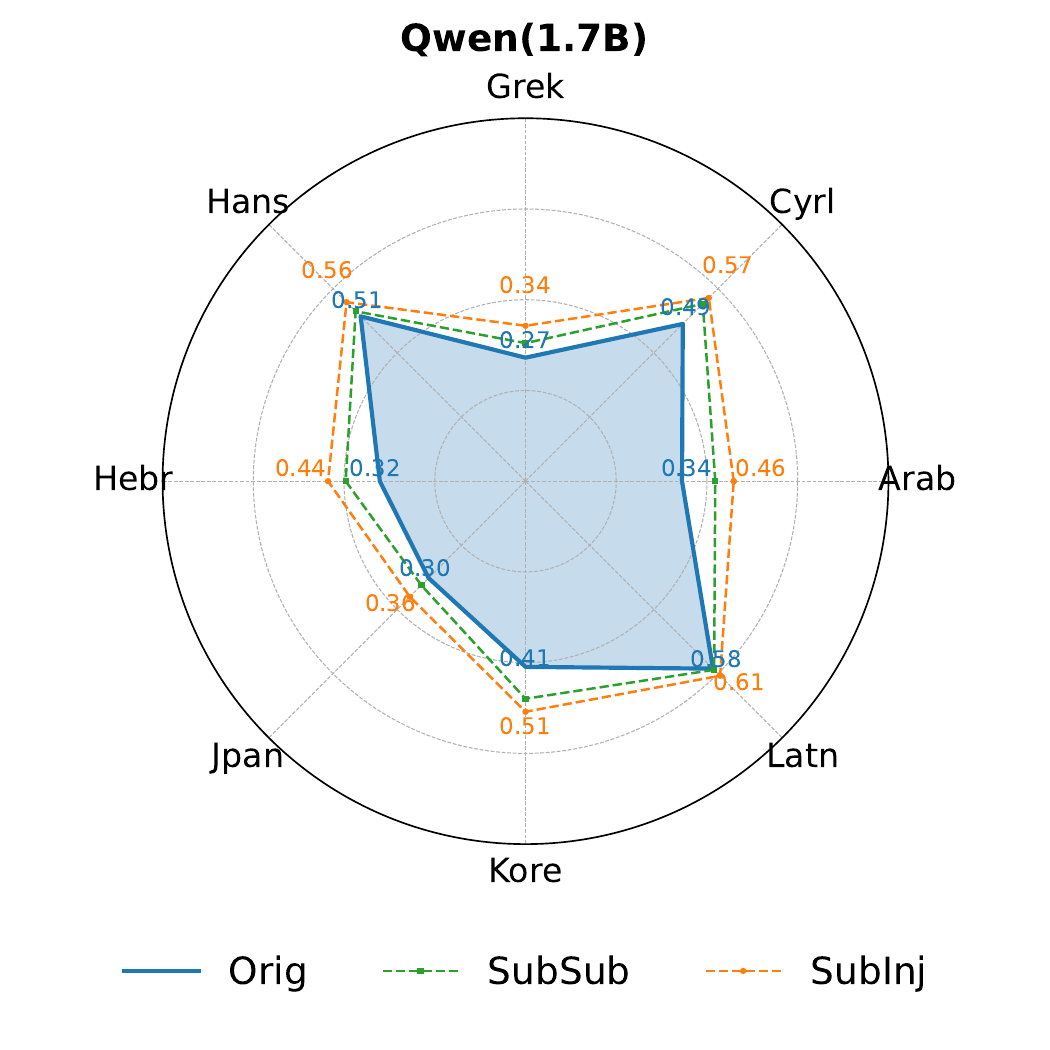}
    \includegraphics[width=0.15\textwidth]{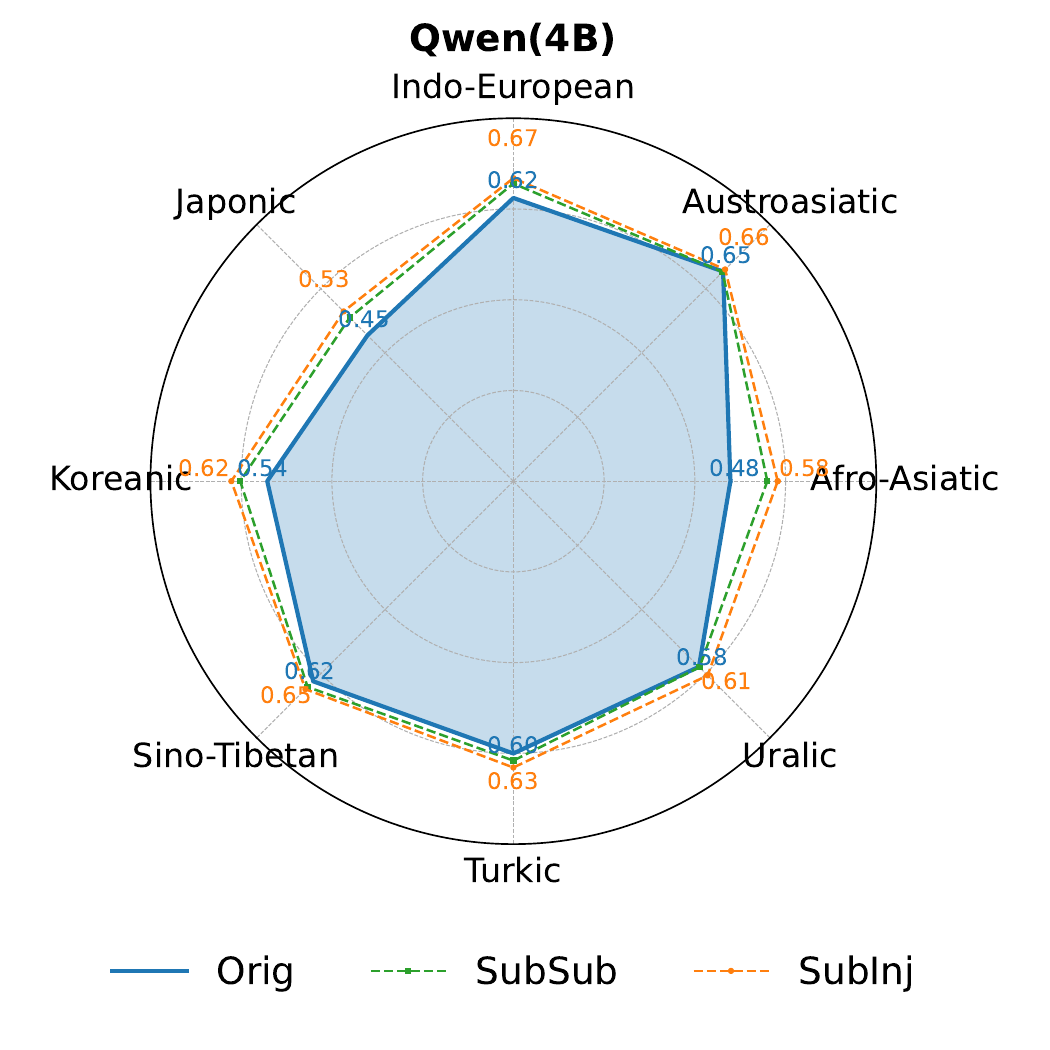}
    \includegraphics[width=0.15\textwidth]{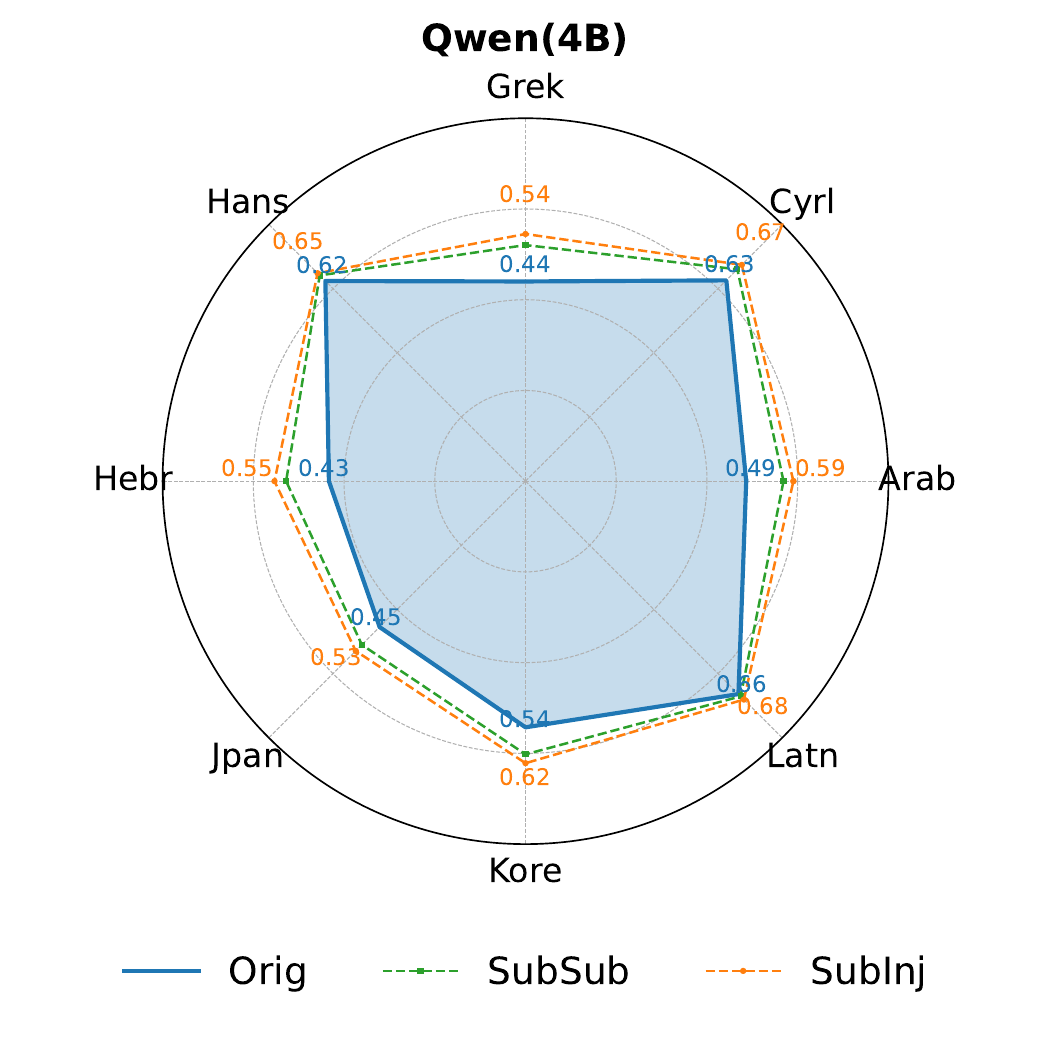}
    \includegraphics[width=0.15\textwidth]{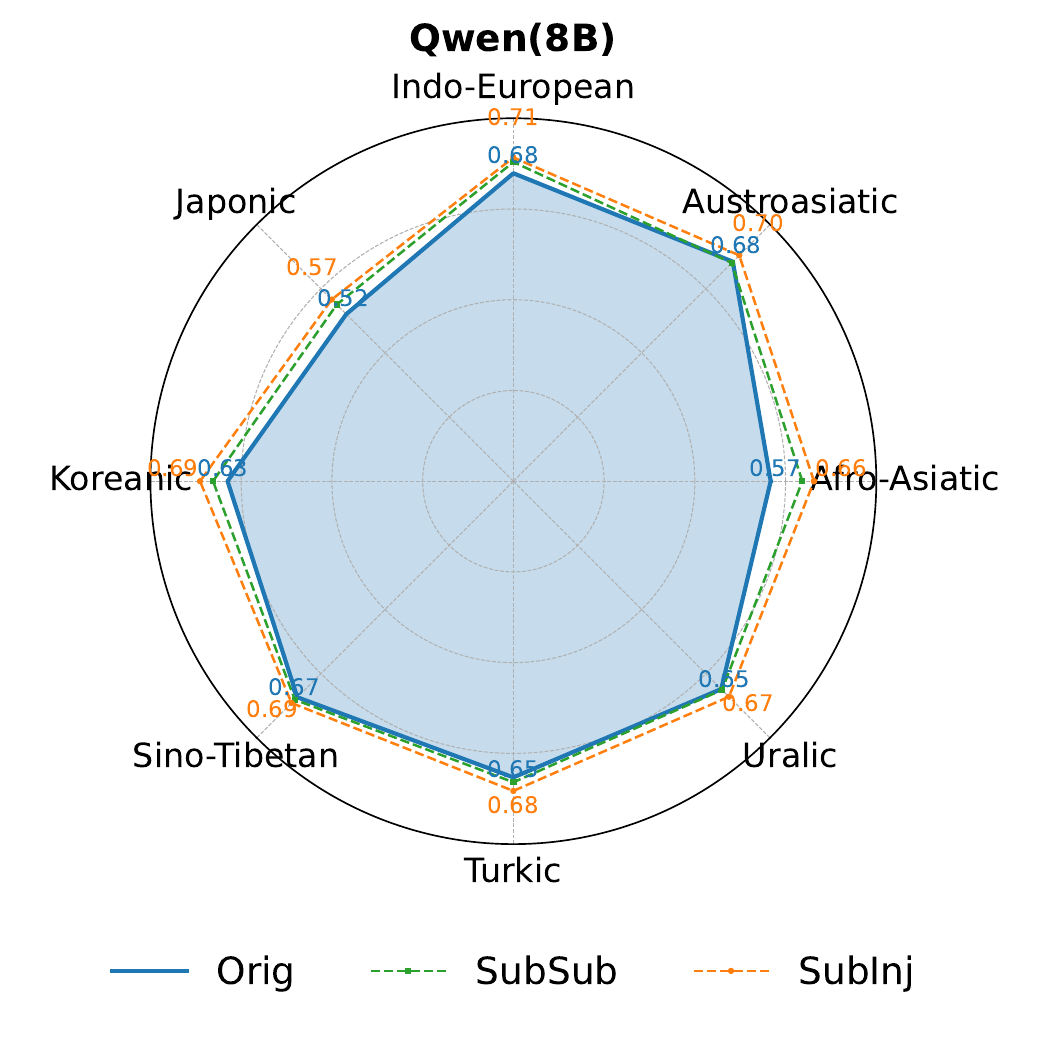}
    \includegraphics[width=0.15\textwidth]{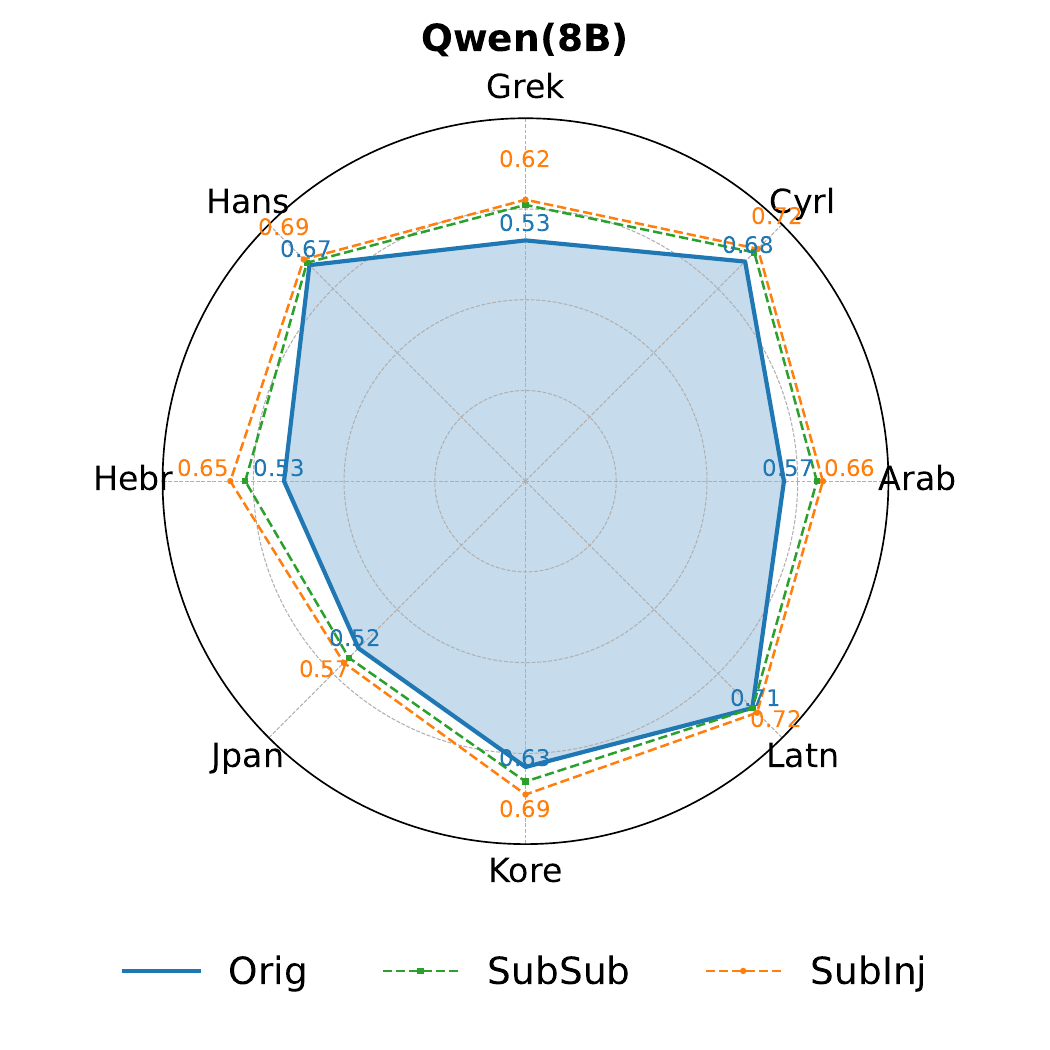}

    \caption{
    Radar plots comparing Baseline (Orig), \textsc{SubSub}, and \textsc{SubInj} factual recall performance (ACC) across language families and script groups.
    \textsc{SubInj} consistently improves  recall across all categories, especially in English-centric models (i.e., OLMo model families) and in non-Latin scripts.
    }
    \label{fig:radar_perf}
\end{figure*}

Table~\ref{tab:subinj_perf} summarizes the recall and consistency gains of both methods across models. 
Additionally, Figure~\ref{fig:radar_perf} presents a fine-grained analysis by language family and script.

\textbf{Both \textsc{SubSub} and \textsc{SubInj} yield consistent improvements in factual recall and consistency.}
As shown in Table~\ref{tab:subinj_perf}, both strategies lead to performance gains across all model families and sizes. 
\textsc{SubInj} consistently outperforms \textsc{SubSub}, albeit by a modest margin, indicating that appending the English subject preserves the information of the original prompt while providing a strong alignment cue.
For multilingual LLMs, the improvements are most pronounced in smaller models.
For example, \texttt{gemma-3-1b-pt} shows an improvement of 13.7\% in ACC and 17.3\% in CO under \textsc{SubInj}. 
However, the gains become small as the size increases, suggesting that larger models already possess stronger internal alignment capabilities.
This aligns with recent findings from \citet{lim2025languagespecificlatentprocesshinders} that larger models are more capable of retrieving knowledge embedded across typologically different languages.\footnote{\citet{lim2025languagespecificlatentprocesshinders} found that while larger models have stronger multilingual capabilities, e.g., crosslingual language understanding, hidden states of different languages are more likely to dissociate from the shared conceptual space compared to smaller models.}

\enote{hs}{last sentence: doesn't this undermine the
  analysis in terms of a common conceptual space for larger models?}
\enote{yl}{yes, i agree, i changed this part and added a footnote.}

\textbf{English-centric models benefit substantially from the remedies, especially for larger models.}
Unlike multilingual models, where gains drop with increasing model size, English-centric models, i.e., OLMo, exhibit the opposite trend: performance improvements from \textsc{SubInj} and \textsc{SubSub} become more pronounced as the model scales up. 
For instance, \texttt{OLMo-2-1124-13B} shows a remarkable increase of over 30\% in ACC and over 40\% in CO.
This reverse pattern reflects a key difference in how multilingual and English-centric models process crosslingual input.
While multilingual models already maintain robust entity alignment across languages, English-centric models lack such alignment and rely heavily on English as a pivot language in their latent conceptual space.
The primary challenge thus lies in transitioning from language-specific inputs into this shared latent space.
This transition depends critically on both strong entity alignment and sufficient model capacity, explaining why the effectiveness of both methods grows with model size.

\textbf{\textsc{SubSub} and \textsc{SubInj} consistently improve factual recall across language families and scripts.}
As illustrated in Figure~\ref{fig:radar_perf}, both methods yield consistent gains across all language families and script groups. 
Improvements are particularly pronounced for non-Latin scripts such as Cyrillic, Arabic, and Chinese (Hans), likely because these languages lack surface-level token overlap with English. 
This makes crosslingual entity alignment, especially subject alignment, more challenging.
Our approach, by explicitly leveraging English subjects, encourages direct entity alignment and thereby facilitates consistency, likely by addressing cases where subject alignment is missing or suboptimal.
Substantial gains are also observed in typologically diverse language families such as Semitic and Slavic, showing the robustness of our methods across linguistic boundaries. 
Notably, English-centric models benefit the most, with marked improvements in nearly all non-Latin scripts and language families. 
This underscores the effectiveness of integrating English subjects as a way to facilitate entity alignment, which further reinforces the role of English as a conceptual pivot language in multilingual factual recall.

\begin{figure*}[h!]
    \centering
\setlength{\belowcaptionskip}{-0.1cm}
    \begin{subfigure}{\linewidth}
        \centering
        \includegraphics[width=0.45\linewidth]{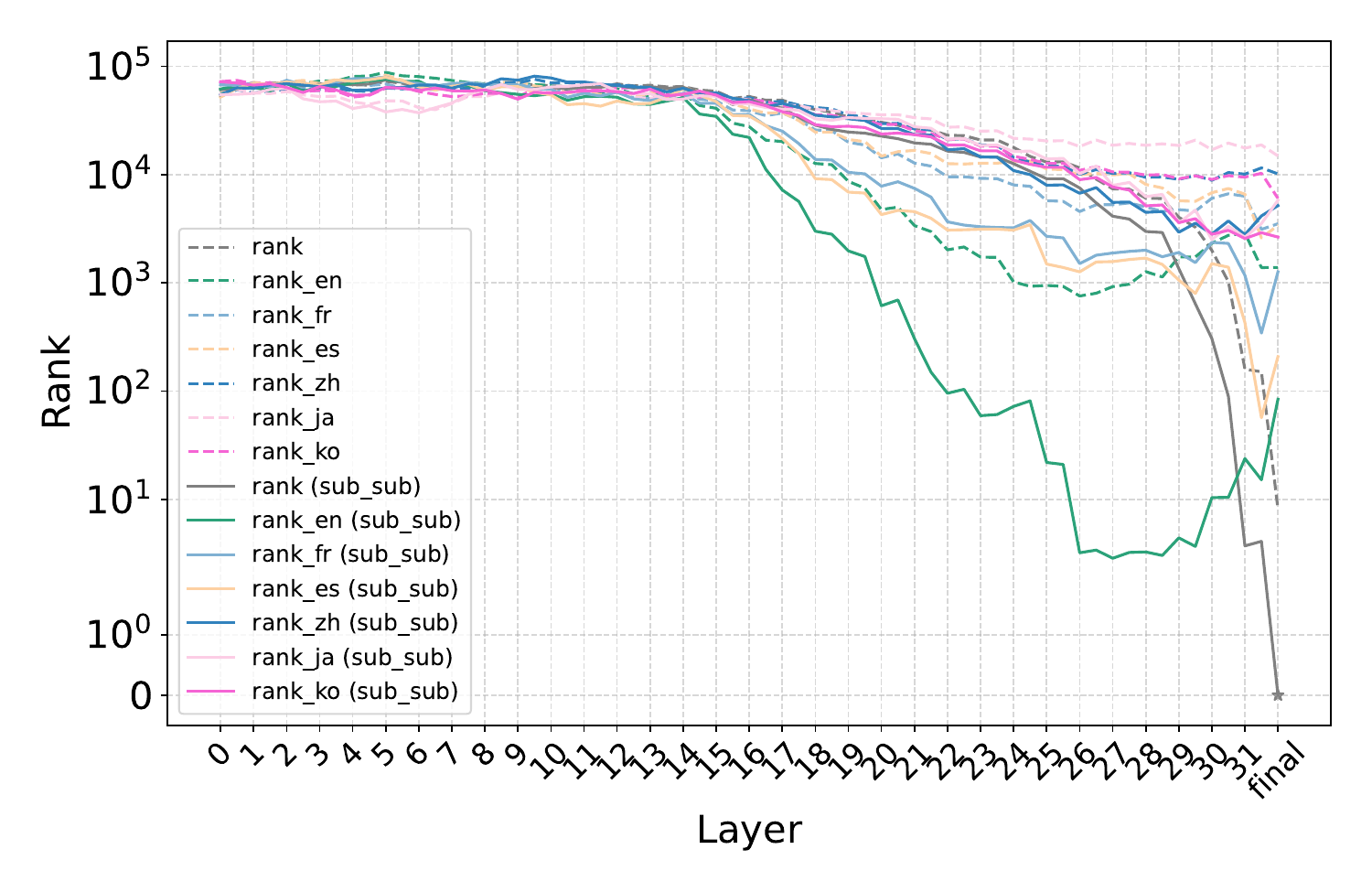}
        \includegraphics[width=0.45\linewidth]{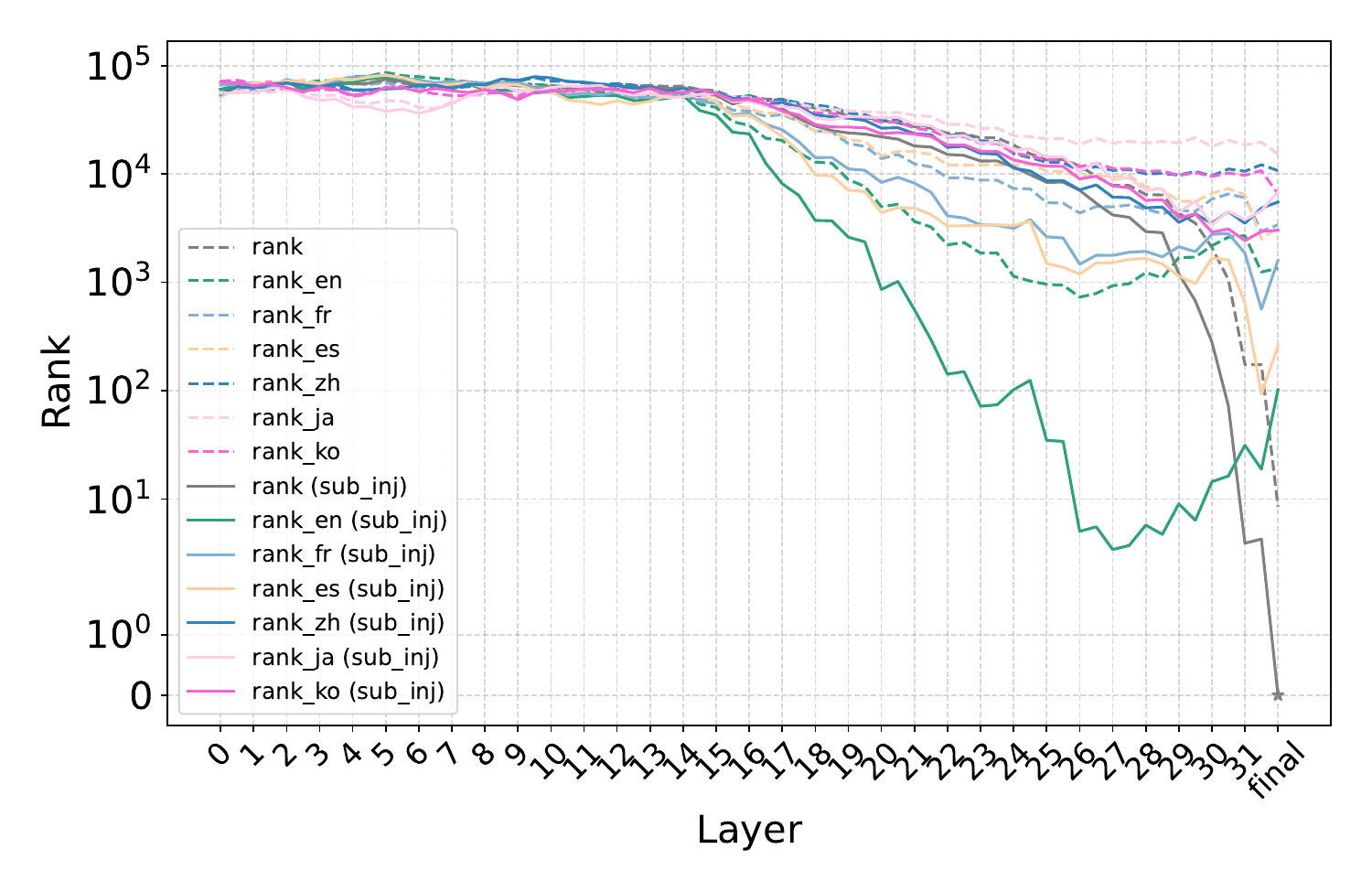}
        \caption{\textbf{LLaMA (8B)}: Left -- with \textsc{SubSub}; Right -- with \textsc{SubInj}.}
        \label{fig:logit-lens-rank-llama3-8b}
    \end{subfigure}

    \begin{subfigure}{\linewidth}
        \centering
        \includegraphics[width=0.45\linewidth]{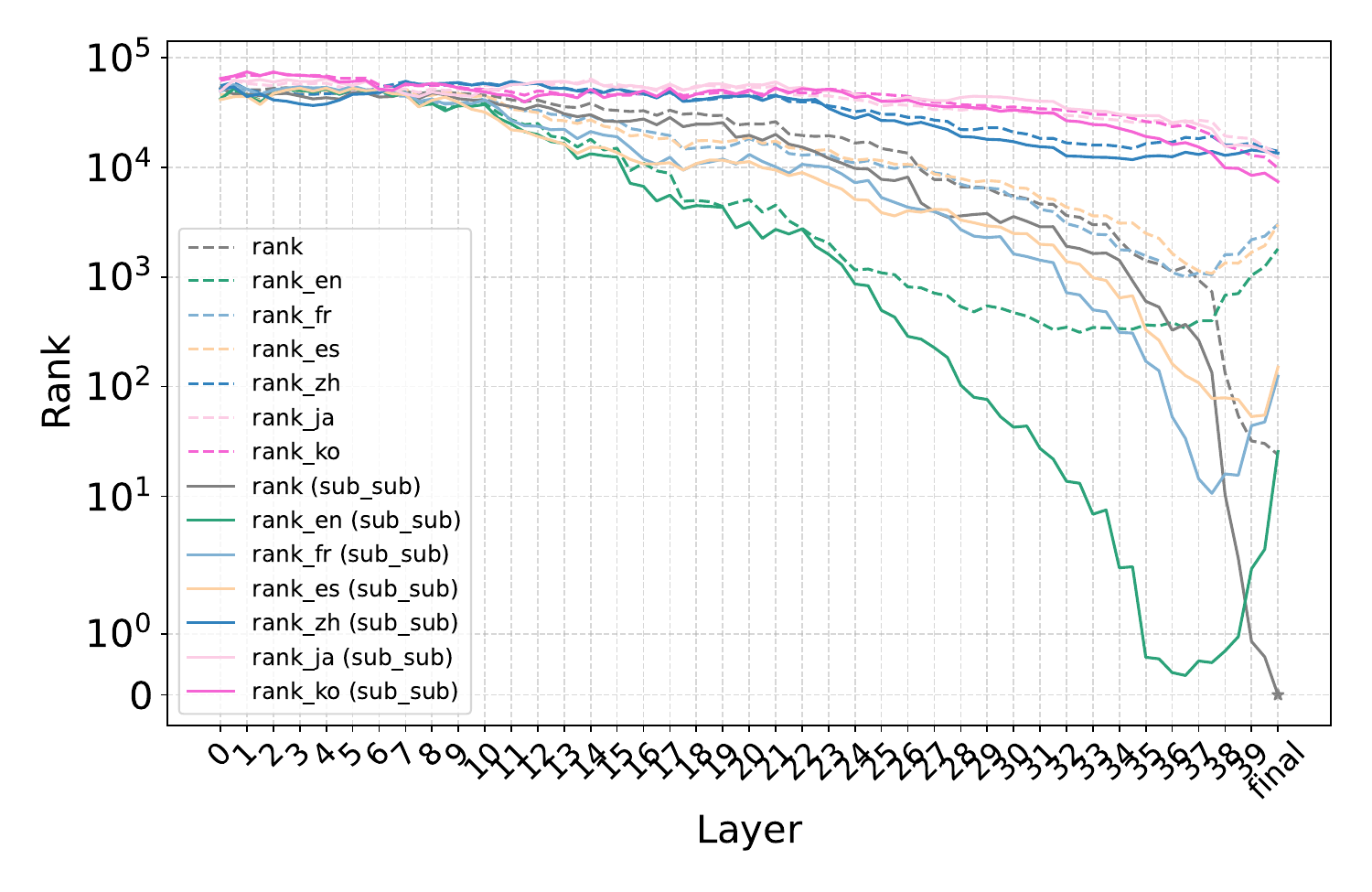}
        \includegraphics[width=0.45\linewidth]{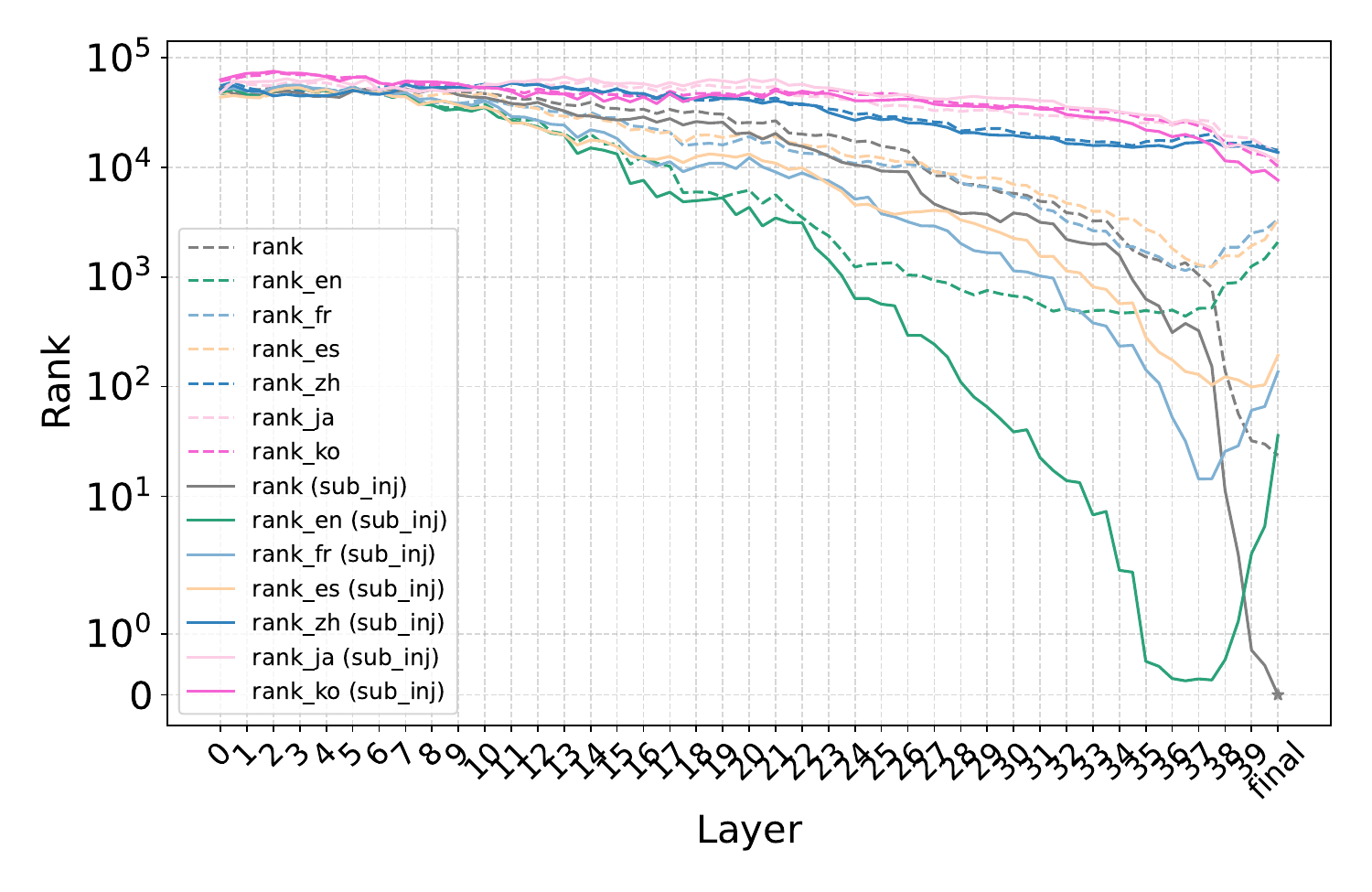}
        \caption{\textbf{OLMo (13B)}: Left -- with \textsc{SubSub}; Right -- with \textsc{SubInj}.}
        \label{fig:logit-lens-rank-olmo-13b}
    \end{subfigure}

    \caption{\textbf{Logit Lens analysis of object token ranks across model layers.} We plot the rank of the target object token (lower is better) in its input language, along with its equivalents in six other languages, using lines of different colors. Dashed lines represent the original prompts, while solid lines correspond to prompts modified with \textsc{SubSub} and \textsc{SubInj}. 
    Both interventions consistently result in lower ranks across languages compared to the original prompts, indicating that entity representations are more aligned in the common conceptual space, resulting in more consistent crosslingual factual prediction.
    }
    \label{fig:logit-lens-rank}
\end{figure*}

\section{Mechanistic Interpretability}\seclabel{mechanistic}
To better understand why \textsc{SubSub} and \textsc{SubInj} improve crosslingual factual recall, we perform a mechanistic analysis of how these interventions affect internal model representations. Specifically, we apply Logit Lens \citep{logit-lens} to project intermediate hidden states at each layer onto the output vocabulary and track the rank of the target object token across layers. A lower rank indicates that the correct answer is more readily accessible from that representation, thus reflecting stronger factual grounding.

We analyze how the concept of the target object entity is represented and decoded across languages by plotting its rank curves in the model’s intermediate layers. For each prompt in a given input language, we track the rank of (i) the correct object token in that language, and (ii) its equivalents in six other languages: English (en), French (fr), Spanish (es), Chinese (zh), Japanese (ja), and Korean (ko). This allows us to examine how well the model encodes a language-agnostic representation of the object entity. Figure~\ref{fig:logit-lens-rank} displays the results for LLaMA (8B) and OLMo (13B), with additional models included in Appendix~\secref{mech-interp-additional}.

As we observe in Figure~\ref{fig:logit-lens-rank}, the solid lines representing \textsc{SubSub} and \textsc{SubInj} consistently yield lower ranks across layers compared to the dashed lines from the original prompts. 
This trend is especially noticeable in later layers, where factual decoding typically occurs. 
These observations indicate that both prompting strategies strengthen entity representation alignment across languages.
The improved alignment of objects suggests that subject alignment -- achieved through input prompts with substitutions/injections -- engages reliable pivot-language processing in the shared conceptual space and then propagates to better object alignment, reducing inconsistencies that possibly arise from misaligned subjects.

Overall, these results confirm that \textsc{SubSub} and \textsc{SubInj} enhance crosslingual entity alignment at the \emph{representation level}, therefore facilitating more accurate and consistent factual recall. 




\section{Conclusion}

In this work, we report a set of converging observations that support the hypothesis that crosslingual factual consistency in multilingual LLMs is highly associated with entity-level alignment -- the model's ability to map subject and object entities into a shared conceptual space across languages. 
In our experiments, we observe that when entity alignment
fails, consistency rarely emerges, revealing alignment as a key factor for consistent multilingual factual recall. 
To address the inconsistency problem, we introduce two simple prompting strategies \textsc{SubSub} and \textsc{SubInj} that integrate English-translated subjects into queries, yielding substantial improvements in both factual recall accuracy and crosslingual consistency.
Finally, our mechanistic analysis suggests that these prompt-based interventions help models better align entities within a shared conceptual space biased toward high-resource languages, e.g., English, thereby facilitating consistent factual recall.


\bibliography{tacl2021}
\bibliographystyle{acl_natbib}

\appendix

\newpage

\section{KLAR Dataset Details}\seclabel{klar}

The statistics of the KLAR dataset \citep{wang2025lostmultilingualitydissectingcrosslingual} are presented in Table~\ref{tab:relation_fact_counts}.
KLAR is based on BMLAMA17 \citep{qi-etal-2023-cross} with some minor modifications to improve the applicability to autoregressive models. 
We use \textbf{2,619} facts grouped into \textbf{20}~relation categories.

\begin{table}[t]
\small
\setlength{\belowcaptionskip}{-0.4cm}
\setlength{\tabcolsep}{0.1mm}
\centering
\resizebox{\linewidth}{!}{%
\begin{tabular}{l r}
\hline
\textbf{Relation} & \textbf{Number of Facts} \\
\hline
\texttt{applies\_to\_jurisdiction} & 79 \\
\texttt{capital} & 336 \\
\texttt{capital\_of} & 212 \\
\texttt{continent} & 212 \\
\texttt{country\_of\_citizenship} & 60 \\
\texttt{developer} & 76 \\
\texttt{field\_of\_work} & 167 \\
\texttt{headquarters\_location} & 51 \\
\texttt{instrument} & 46 \\
\texttt{language\_of\_work\_or\_name} & 108 \\
\texttt{languages\_spoken} & 104 \\
\texttt{location\_of\_formation} & 66 \\
\texttt{manufacturer} & 35 \\
\texttt{native\_language} & 130 \\
\texttt{occupation} & 46 \\
\texttt{official\_language} & 602 \\
\texttt{owned\_by} & 50 \\
\texttt{place\_of\_birth} & 35 \\
\texttt{place\_of\_death} & 79 \\
\texttt{religion} & 125 \\
\hline
\textbf{Total} & \textbf{2,619} \\
\hline
\end{tabular}
}
\caption{Number of facts per relation type.}
\label{tab:relation_fact_counts}
\end{table}

\section{Supplementary Results}\seclabel{sumpplementary}

\begin{figure*}[h]
    \centering
    \subcaptionbox{\scriptsize LLaMA (1B) \textsc{SubSub}}{%
        \includegraphics[width=0.22\textwidth]{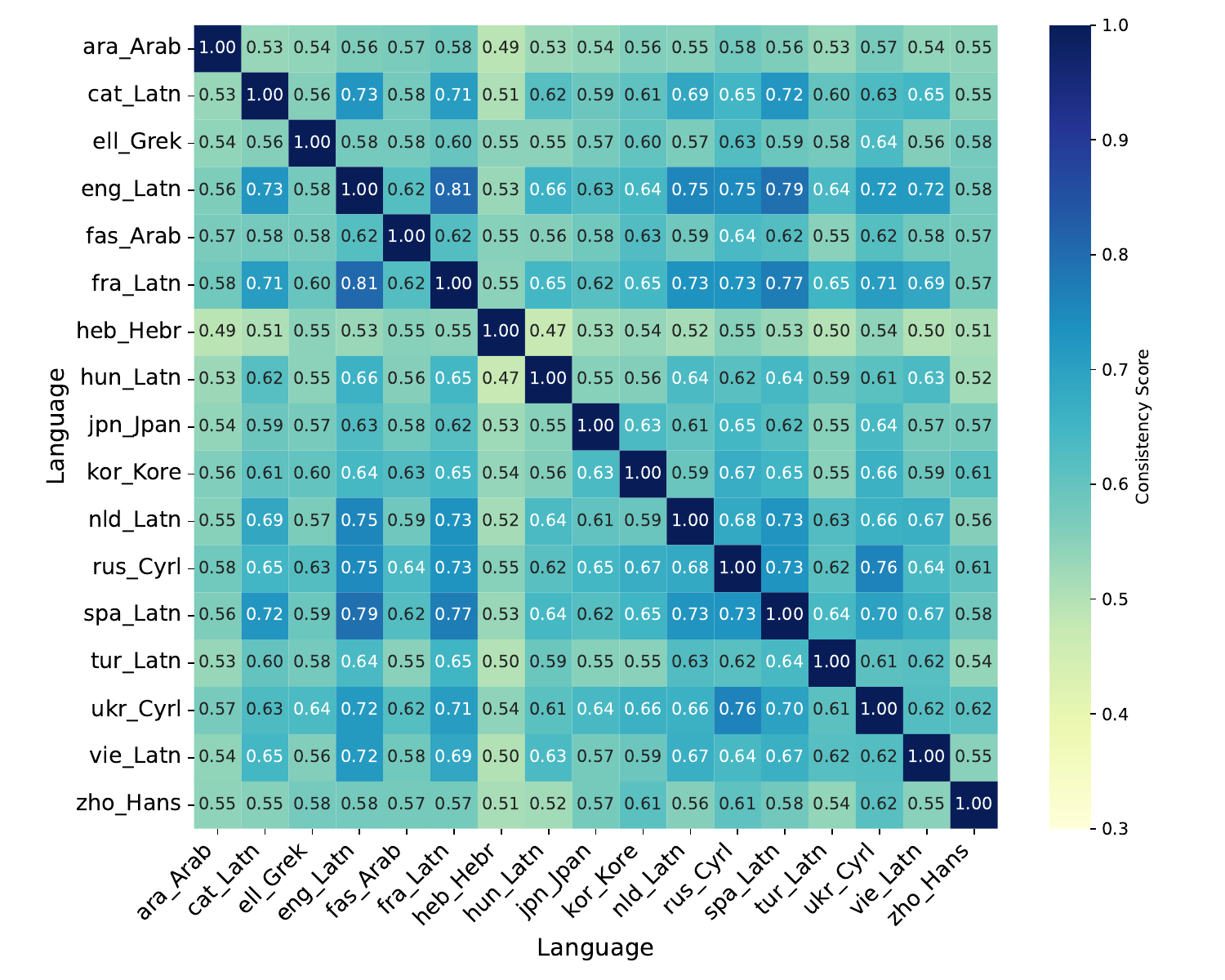}
    }
    \subcaptionbox{\scriptsize LLaMA (1B) \textsc{SubInj}}{%
        \includegraphics[width=0.22\textwidth]{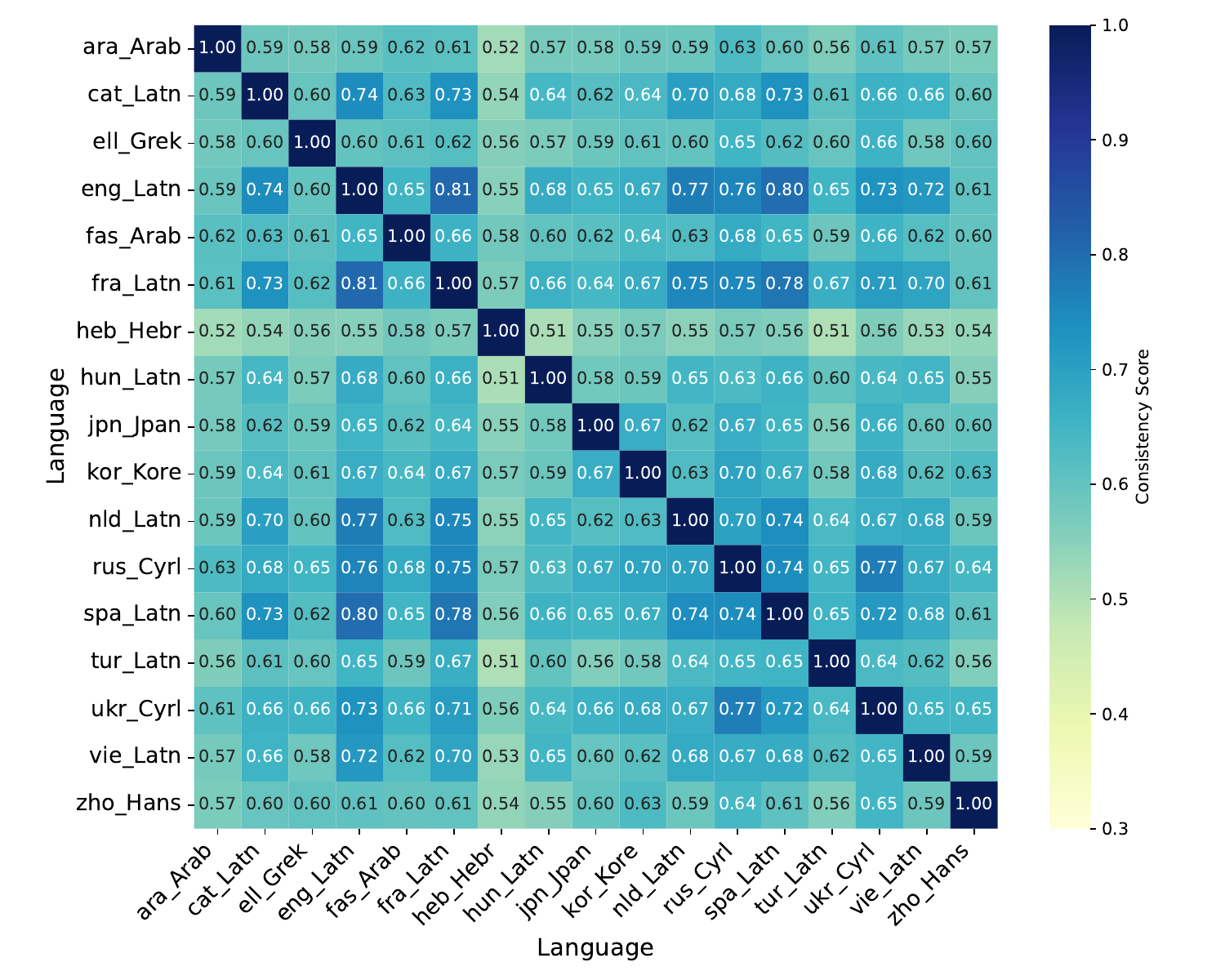}
    }
    \subcaptionbox{\scriptsize LLaMA (8B) \textsc{SubSub}}{%
        \includegraphics[width=0.22\textwidth]{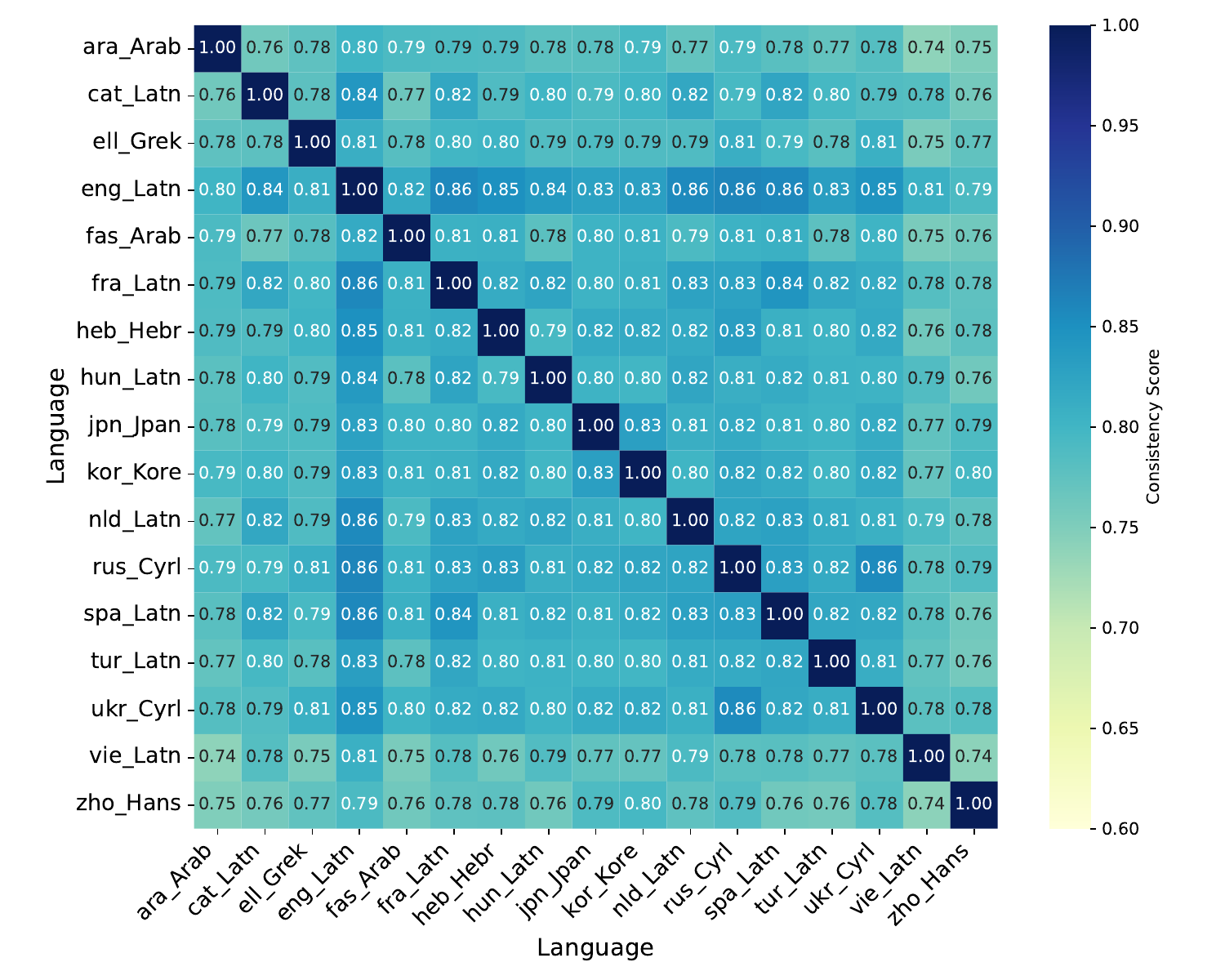}
    }
    \subcaptionbox{\scriptsize LLaMA (8B) \textsc{SubInj}}{%
        \includegraphics[width=0.22\textwidth]{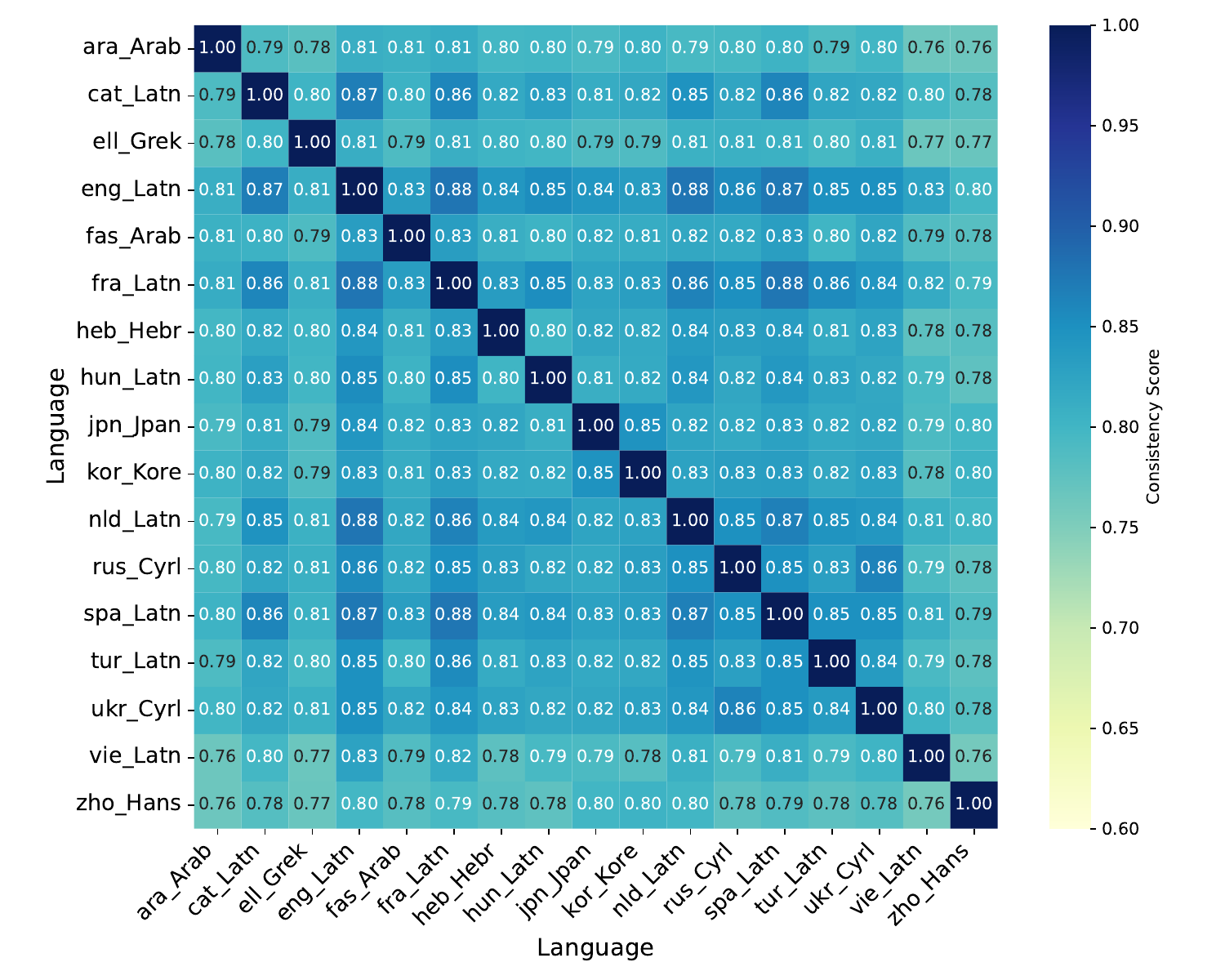}
    }

    \subcaptionbox{\scriptsize OLMo (1B) \textsc{SubSub}}{%
        \includegraphics[width=0.22\textwidth]{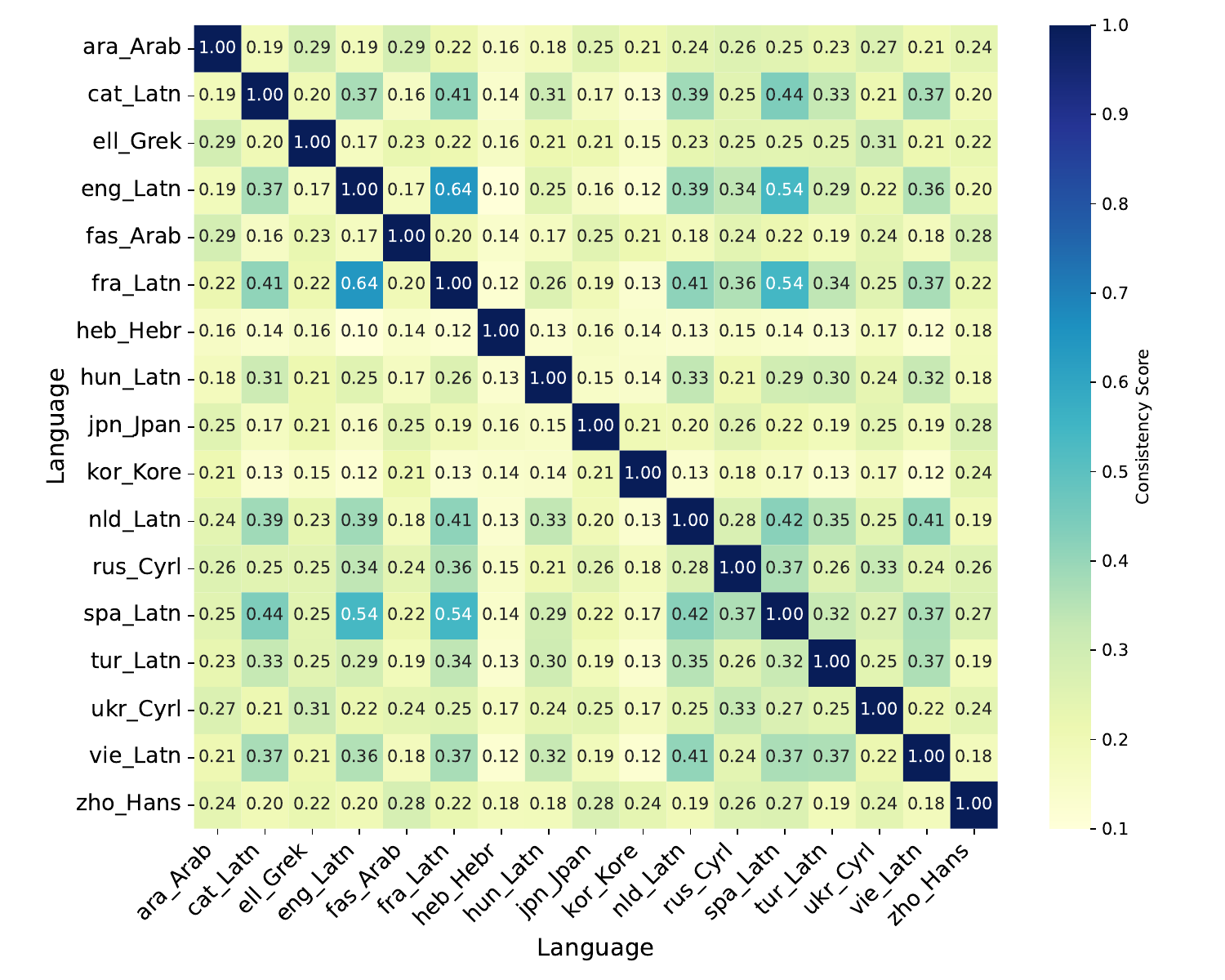}
    }
    \subcaptionbox{\scriptsize OLMo (1B) \textsc{SubInj}}{%
        \includegraphics[width=0.22\textwidth]{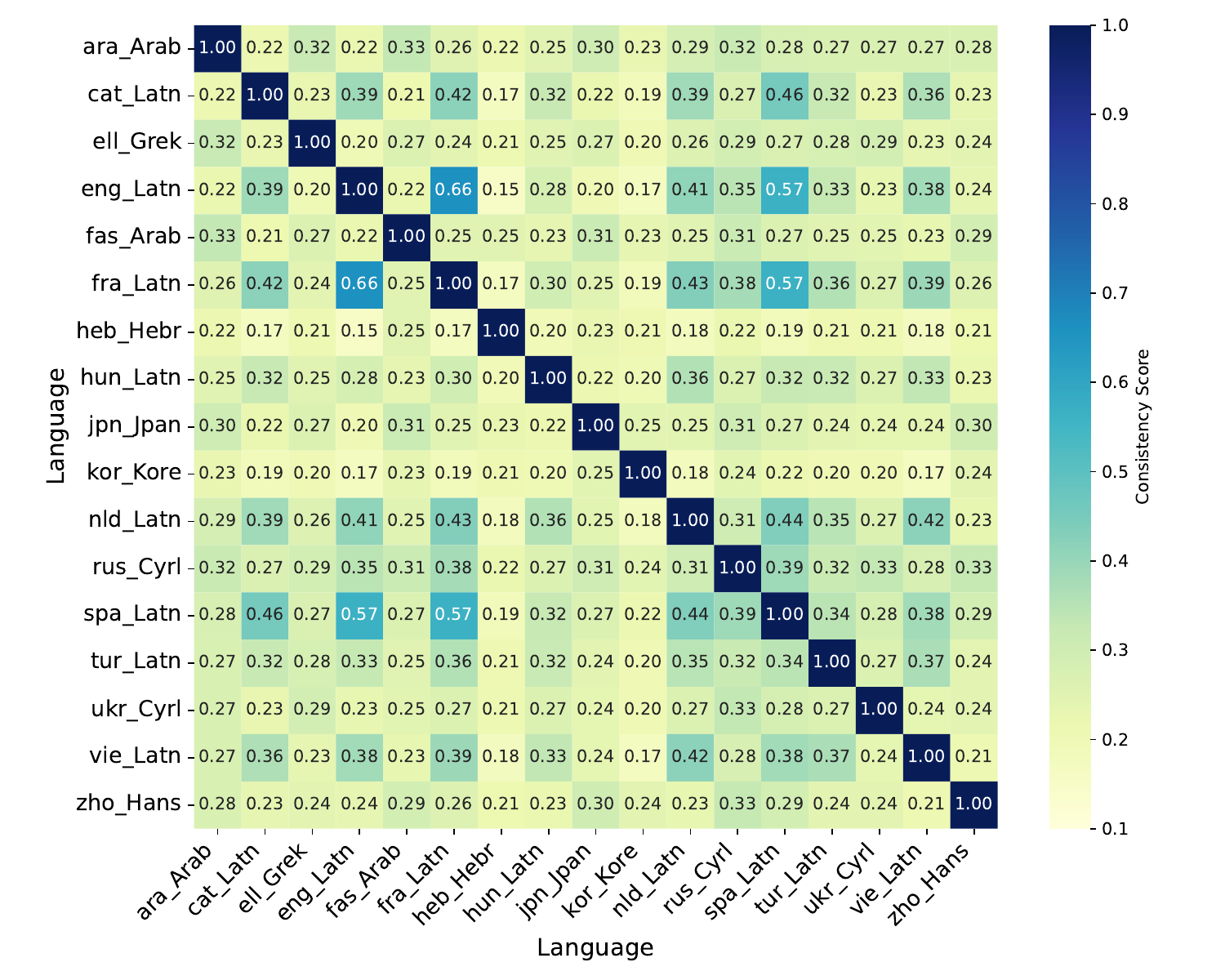}
    }
    \subcaptionbox{\scriptsize OLMo (13B) \textsc{SubSub}}{%
        \includegraphics[width=0.22\textwidth]{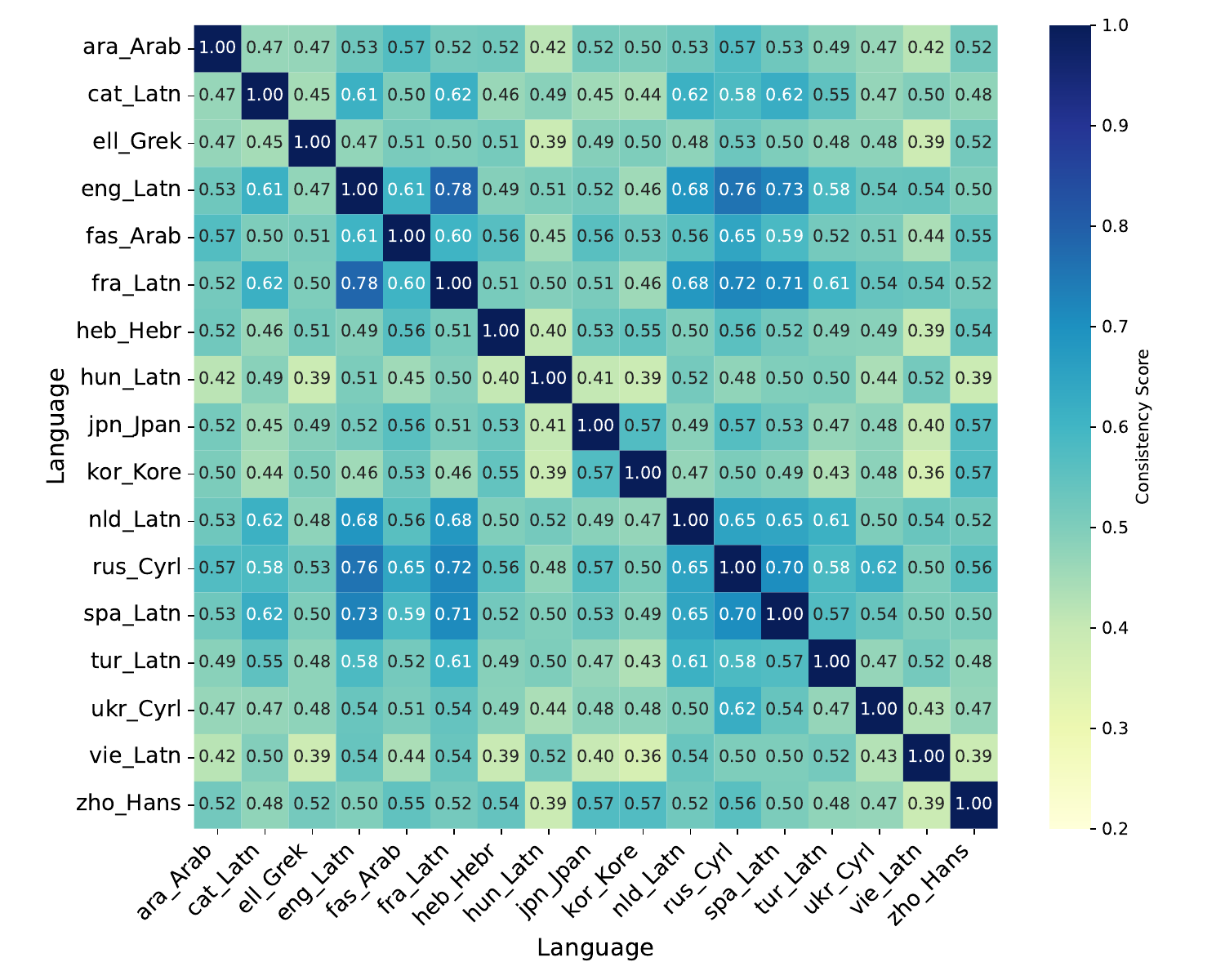}
    }
    \subcaptionbox{\scriptsize OLMo (13B) \textsc{SubInj}}{%
        \includegraphics[width=0.22\textwidth]{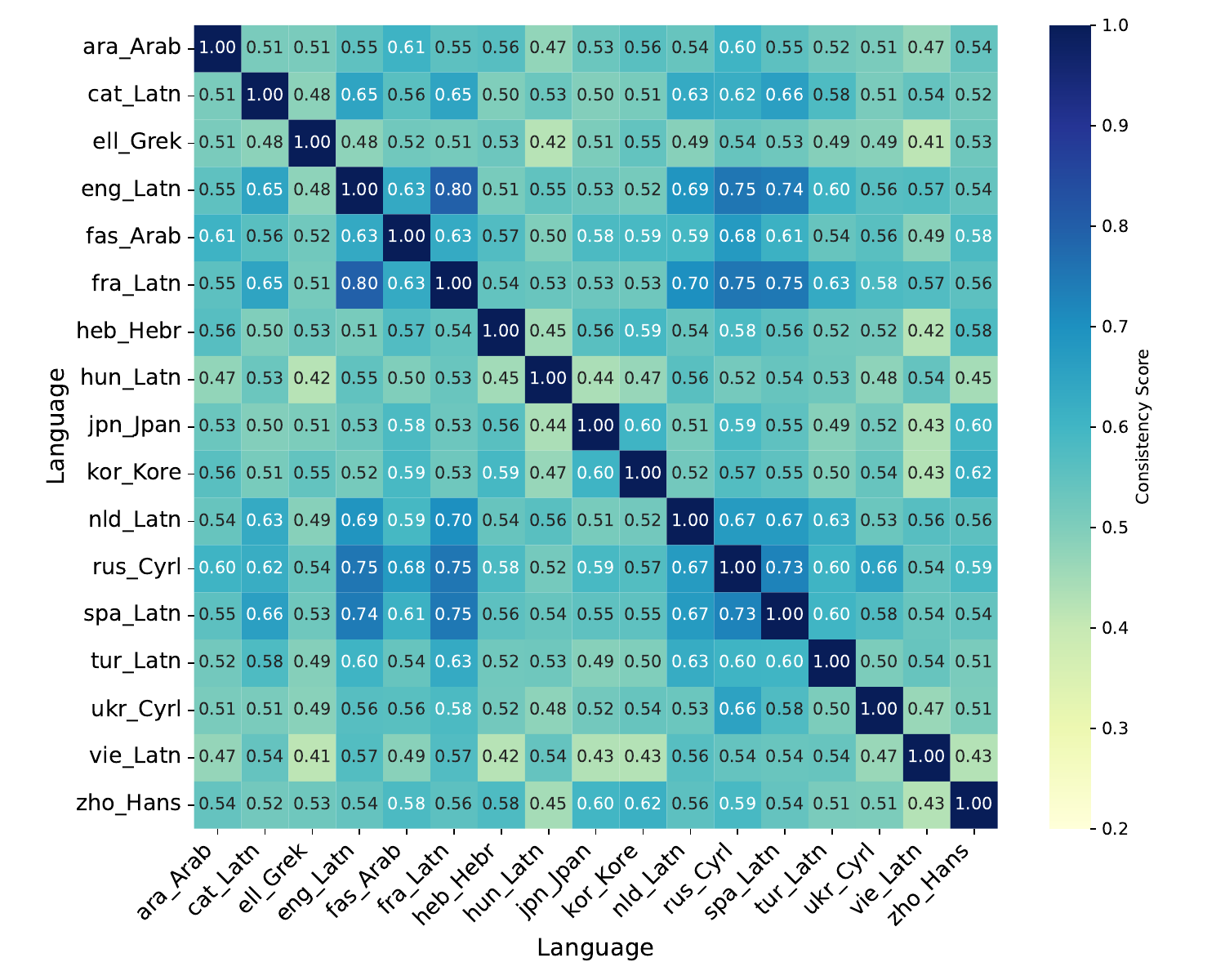}
    }
    \caption{Factual recall consistency across language pairs using \textbf{English} translations of subjects in \textsc{SubSub} and \textsc{SubInj}.}
    \label{fig:perf_en}
\end{figure*}

\begin{figure*}[h]
    \centering
    \subcaptionbox{\scriptsize LLaMA (1B) \textsc{SubSub}}{%
        \includegraphics[width=0.22\textwidth]{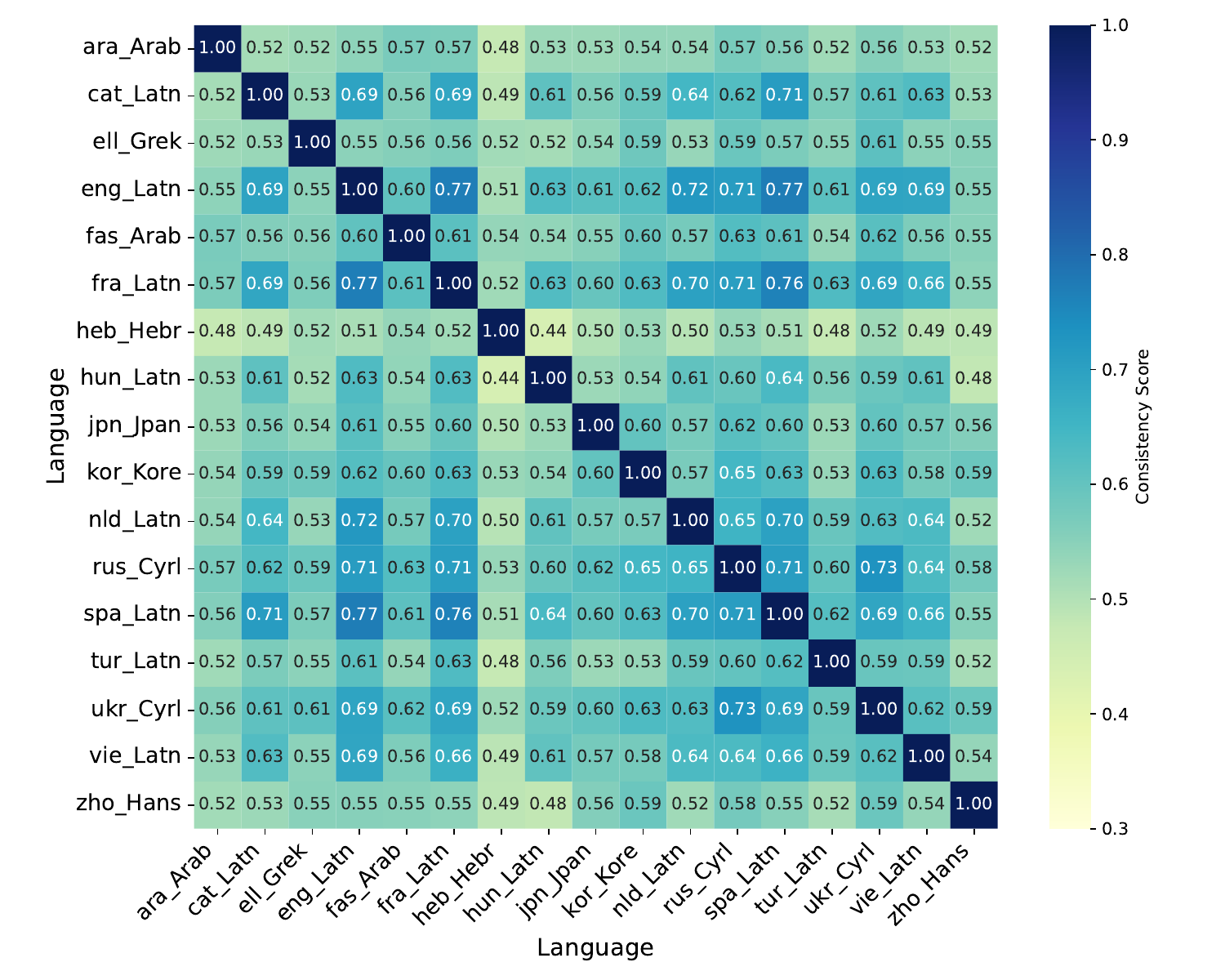}
    }
    \subcaptionbox{\scriptsize LLaMA (1B) \textsc{SubInj}}{%
        \includegraphics[width=0.22\textwidth]{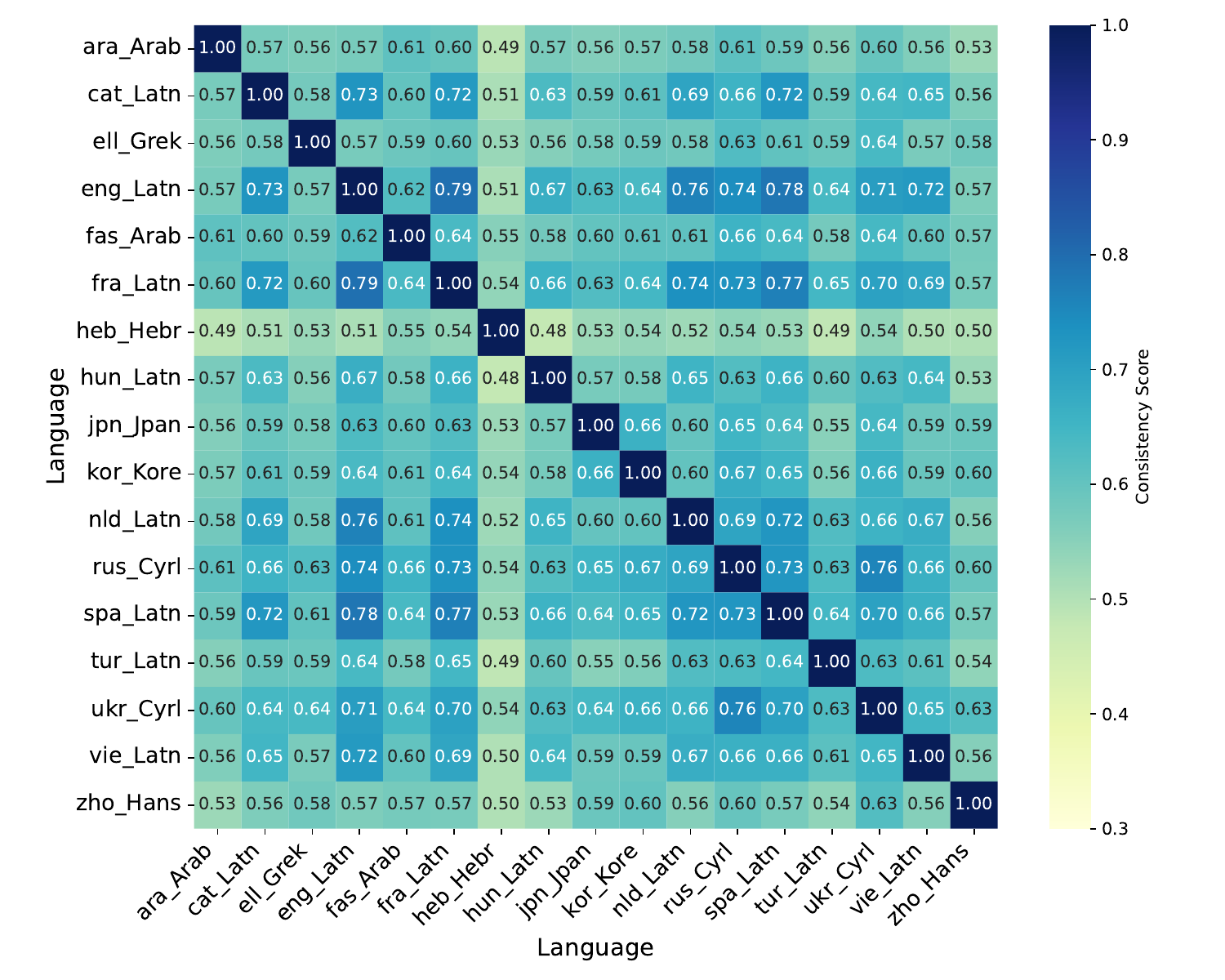}
    }
    \subcaptionbox{\scriptsize LLaMA (8B) \textsc{SubSub}}{%
        \includegraphics[width=0.22\textwidth]{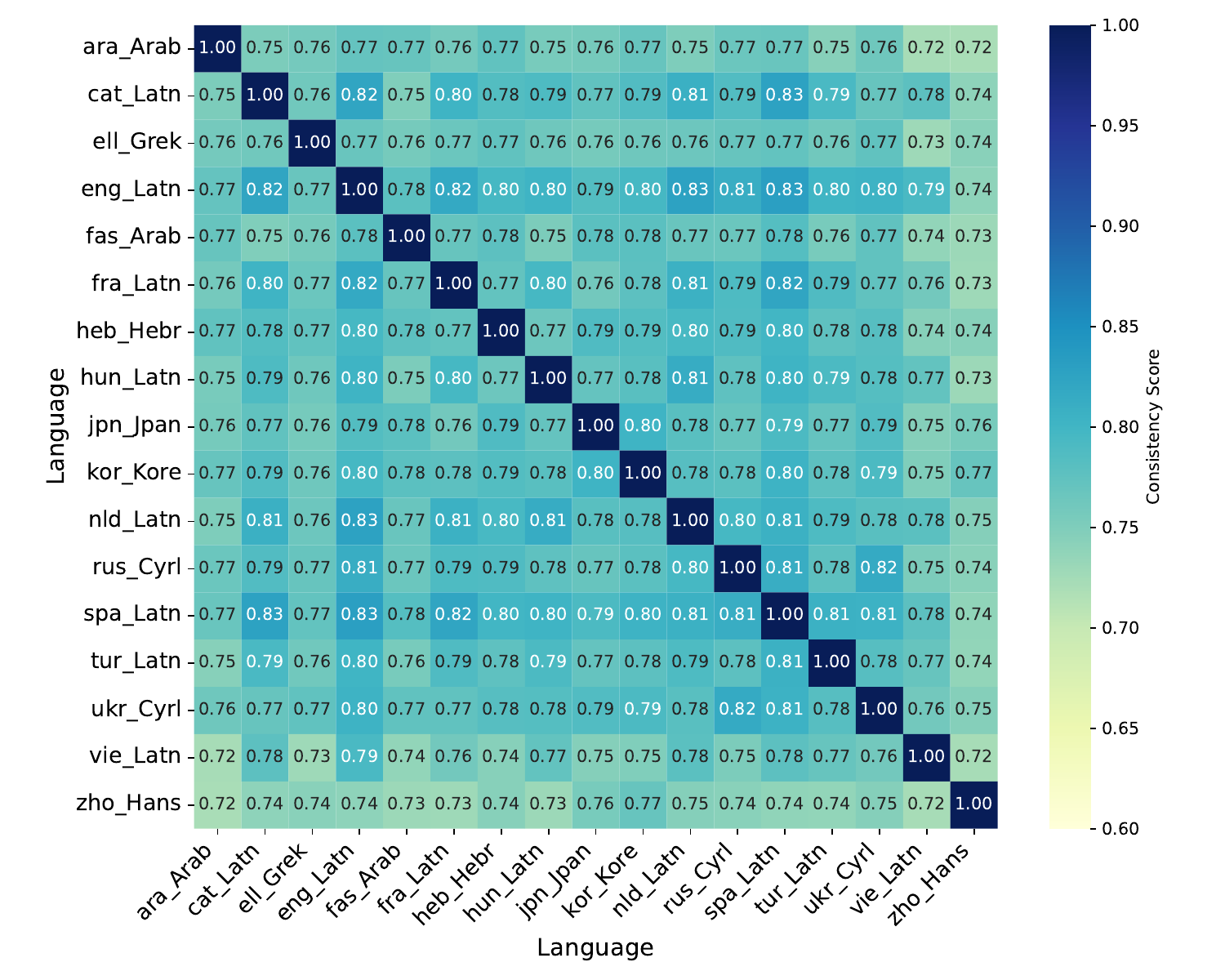}
    }
    \subcaptionbox{\scriptsize LLaMA (8B) \textsc{SubInj}}{%
        \includegraphics[width=0.22\textwidth]{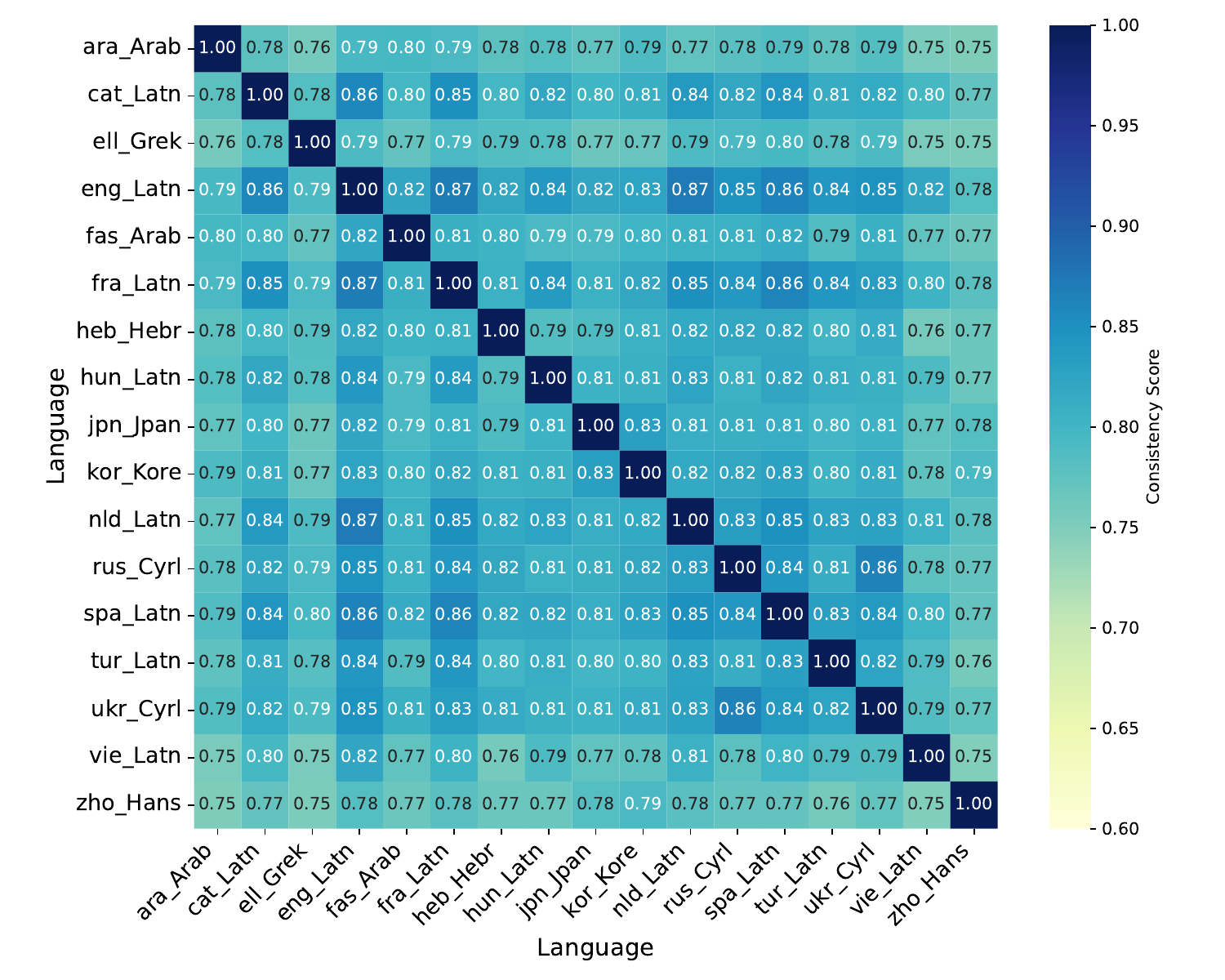}
    }

    \subcaptionbox{\scriptsize OLMo (1B) \textsc{SubSub}}{%
        \includegraphics[width=0.22\textwidth]{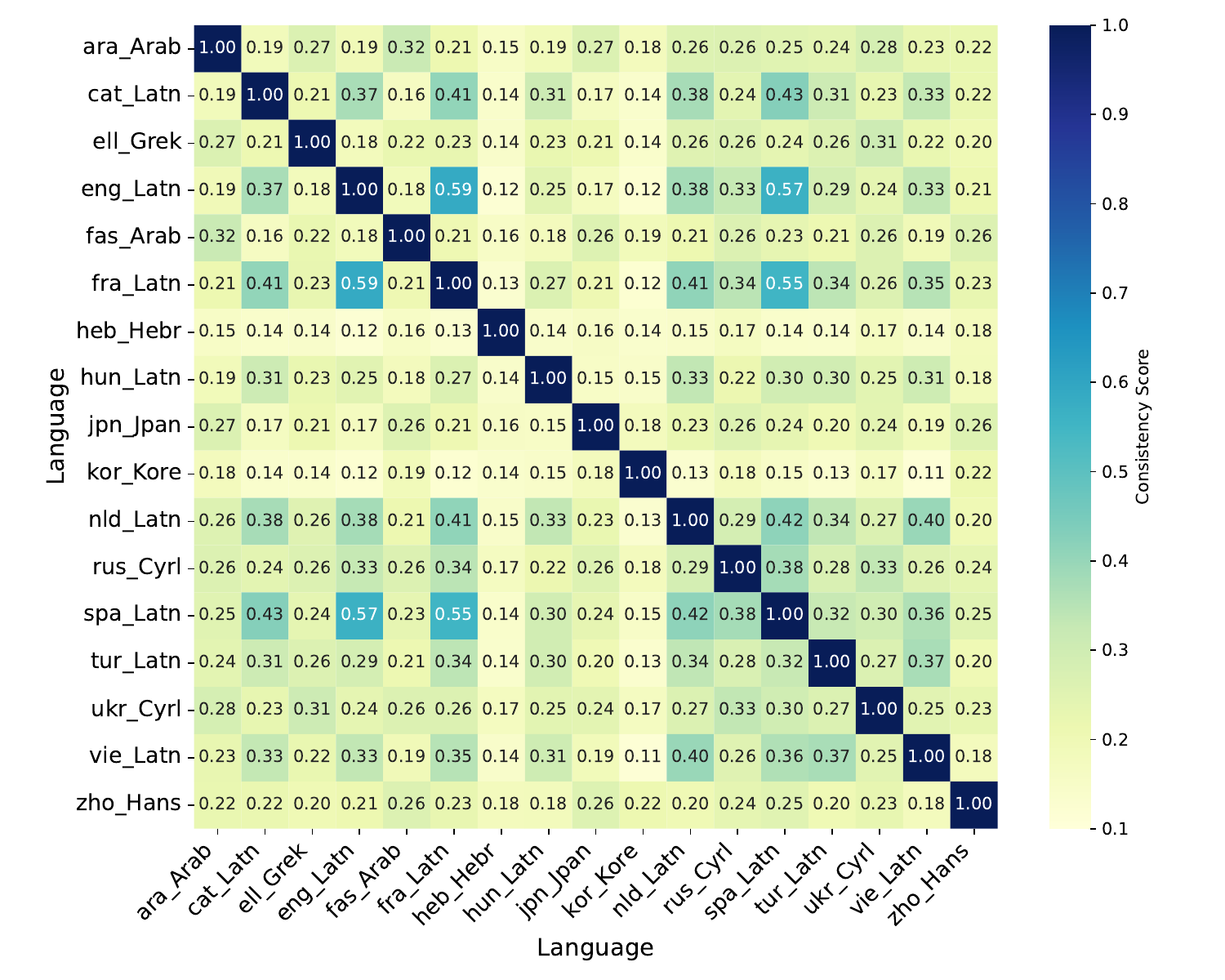}
    }
    \subcaptionbox{\scriptsize OLMo (1B) \textsc{SubInj}}{%
        \includegraphics[width=0.22\textwidth]{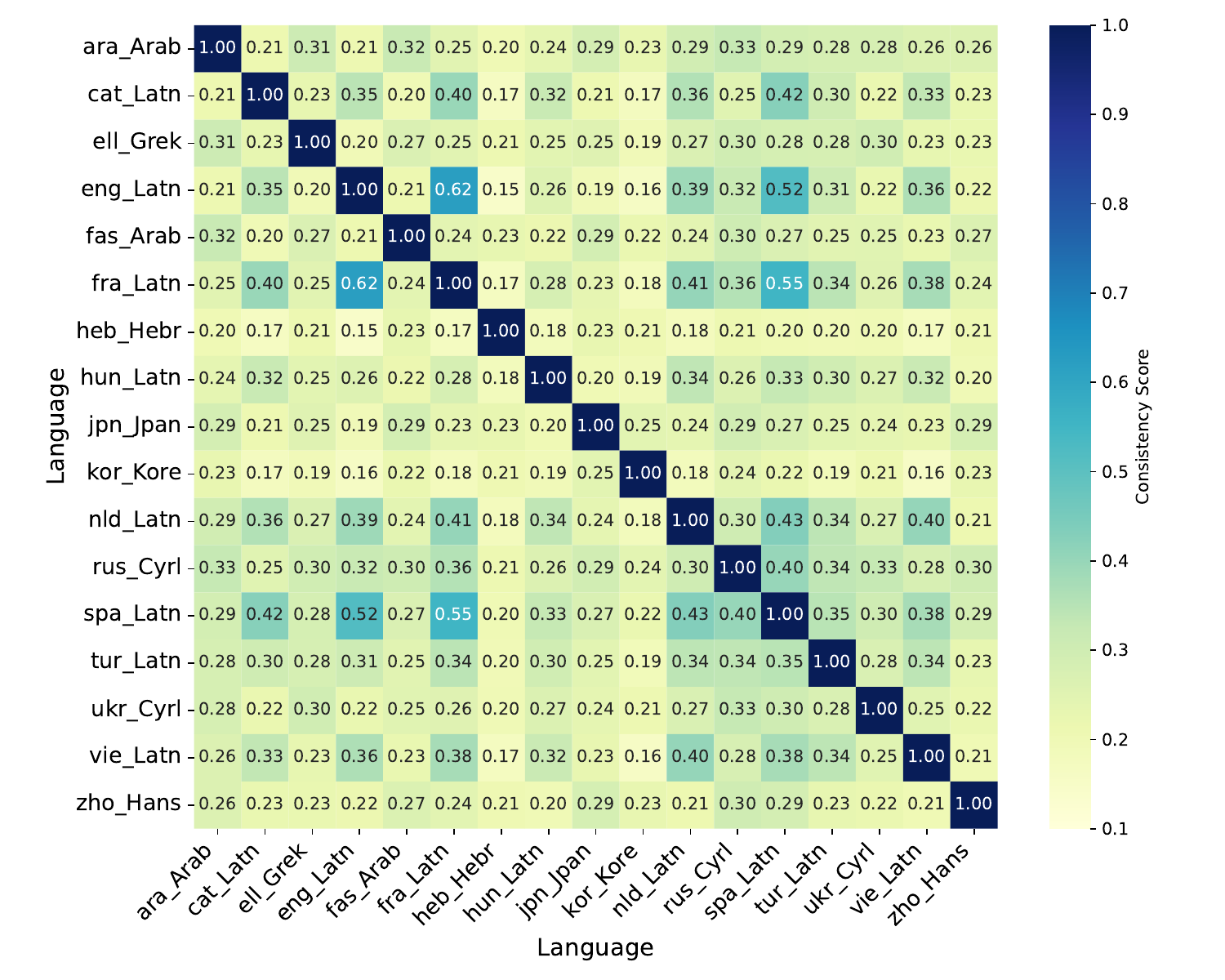}
    }
    \subcaptionbox{\scriptsize OLMo (13B) \textsc{SubSub}}{%
        \includegraphics[width=0.22\textwidth]{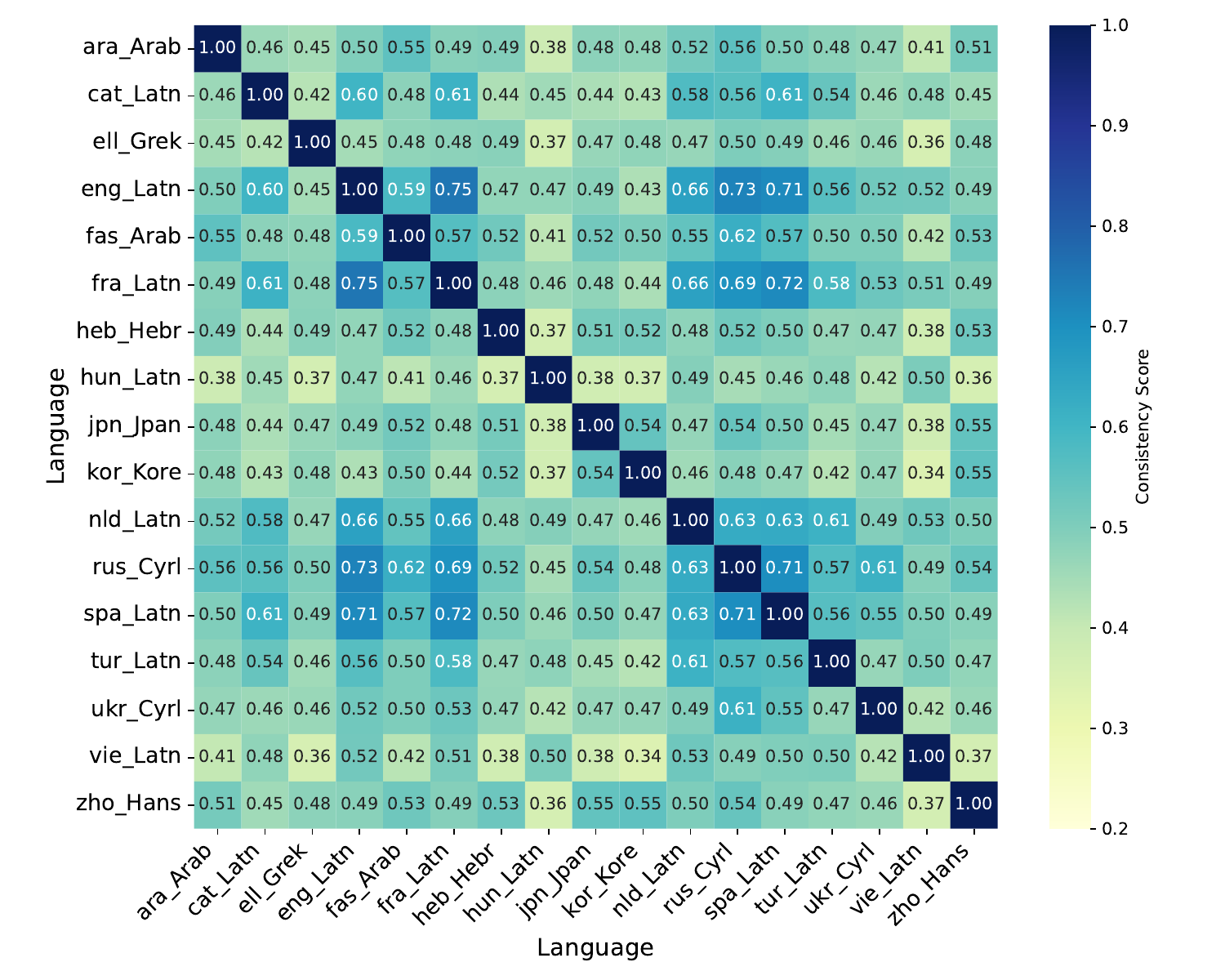}
    }
    \subcaptionbox{\scriptsize OLMo (13B) \textsc{SubInj}}{%
        \includegraphics[width=0.22\textwidth]{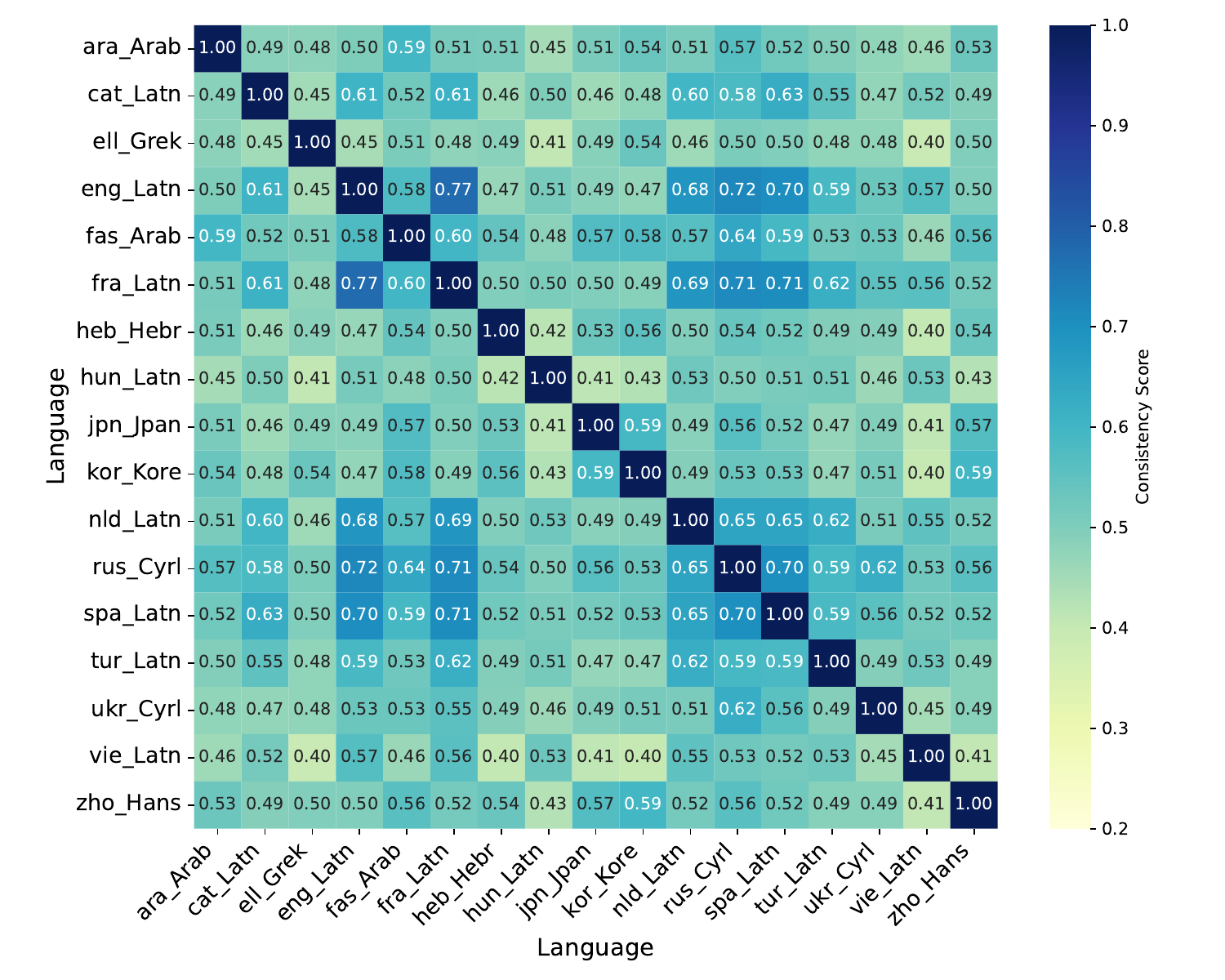}
    }
    \caption{Factual recall consistency across language pairs using \textbf{Spanish} translations of subjects in \textsc{SubSub} and \textsc{SubInj}.}
    \label{fig:perf_es}
\end{figure*}

\begin{figure*}[h]
    \centering
    \subcaptionbox{\scriptsize LLaMA (1B) \textsc{SubSub}}{%
        \includegraphics[width=0.22\textwidth]{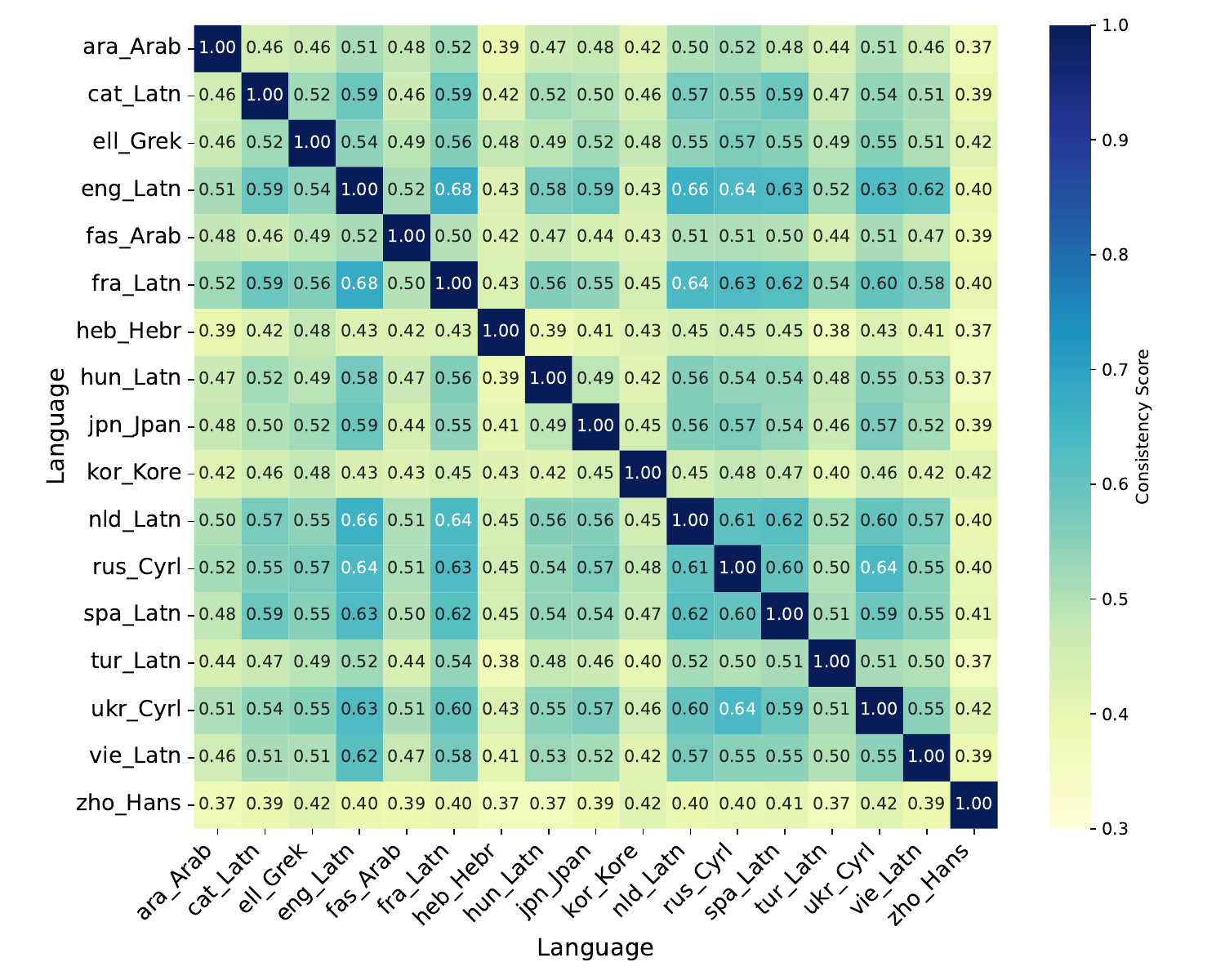}
    }
    \subcaptionbox{\scriptsize LLaMA (1B) \textsc{SubInj}}{%
        \includegraphics[width=0.22\textwidth]{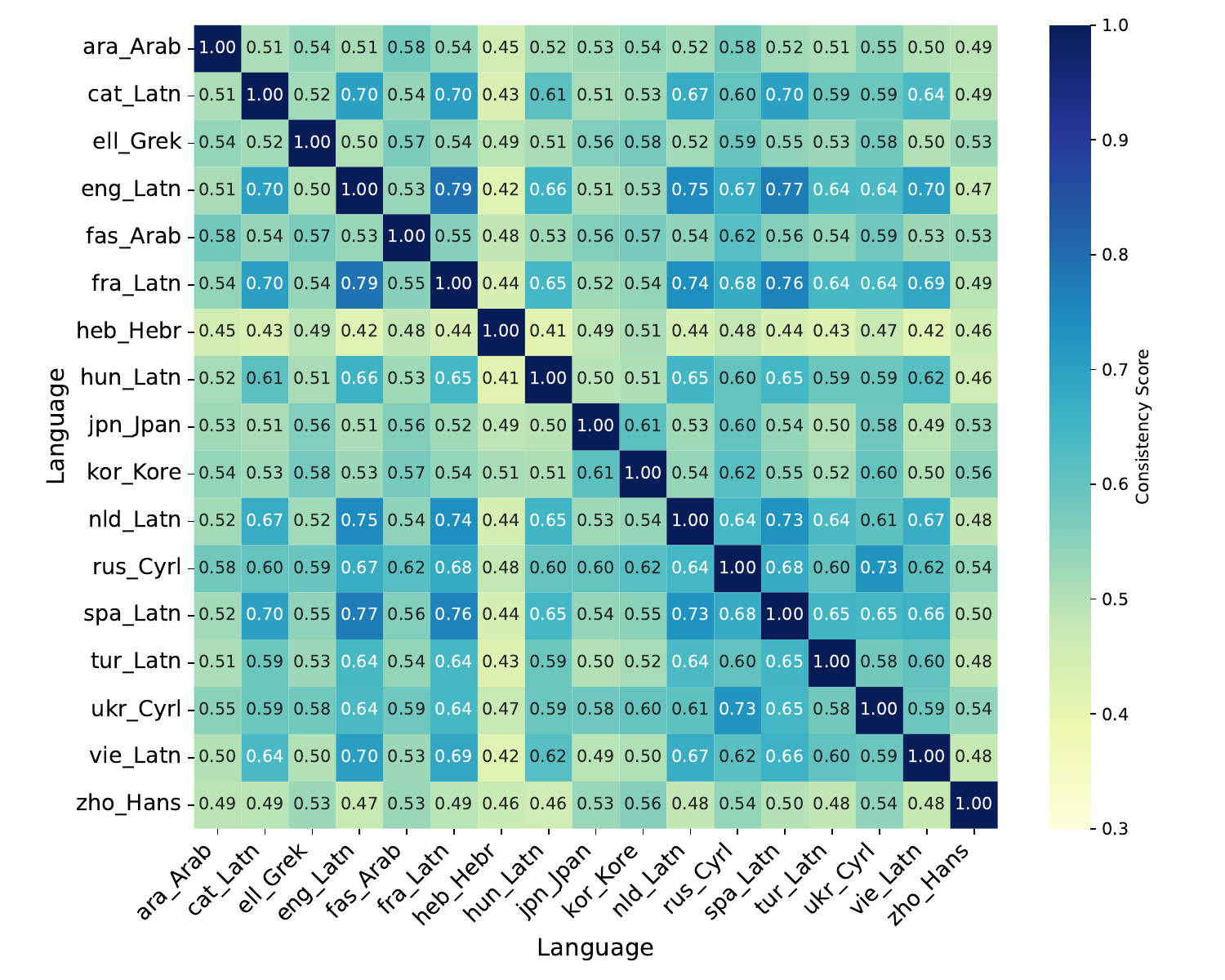}
    }
    \subcaptionbox{\scriptsize LLaMA (8B) \textsc{SubSub}}{%
        \includegraphics[width=0.22\textwidth]{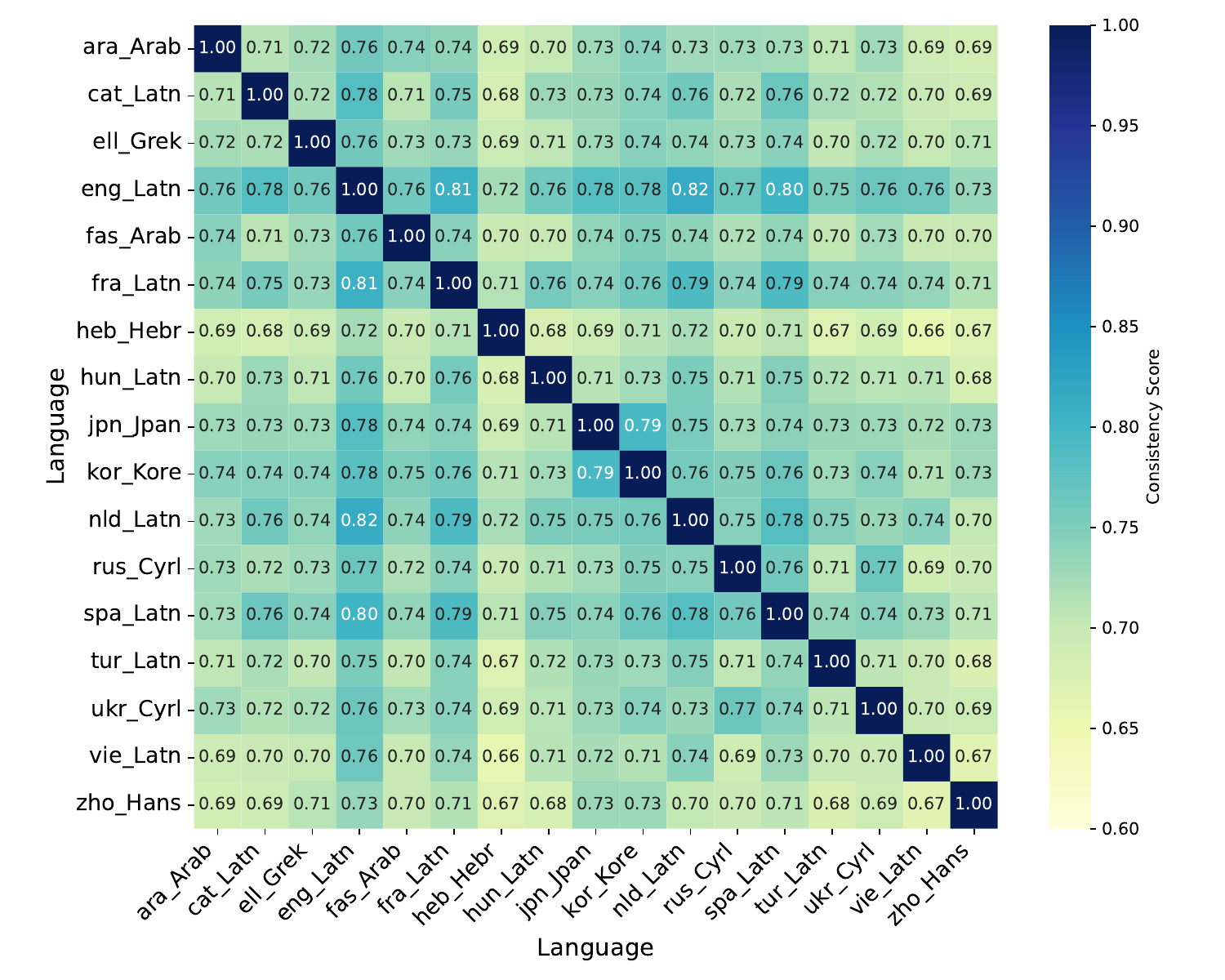}
    }
    \subcaptionbox{\scriptsize LLaMA (8B) \textsc{SubInj}}{%
        \includegraphics[width=0.22\textwidth]{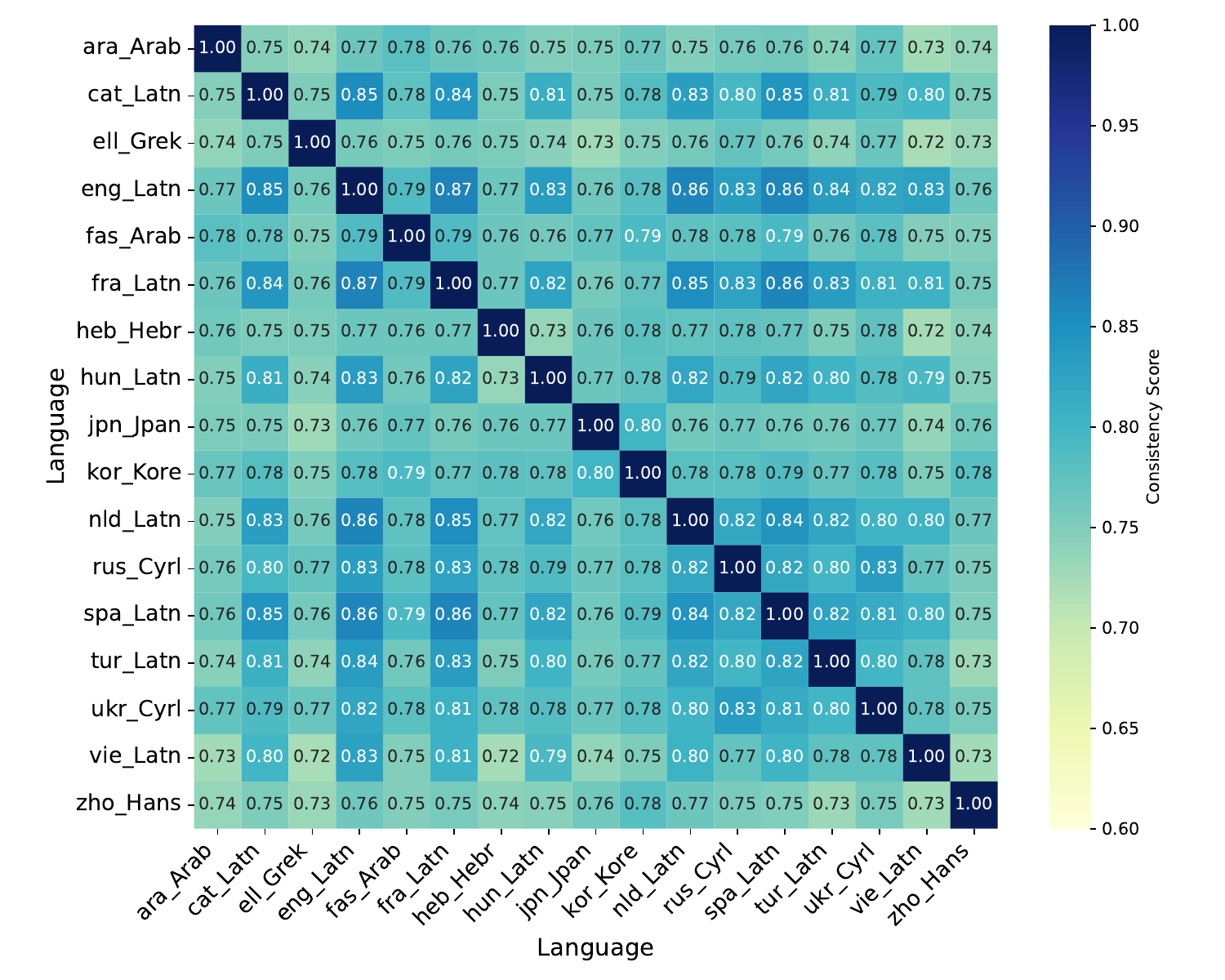}
    }

    \subcaptionbox{\scriptsize OLMo (1B) \textsc{SubSub}}{%
        \includegraphics[width=0.22\textwidth]{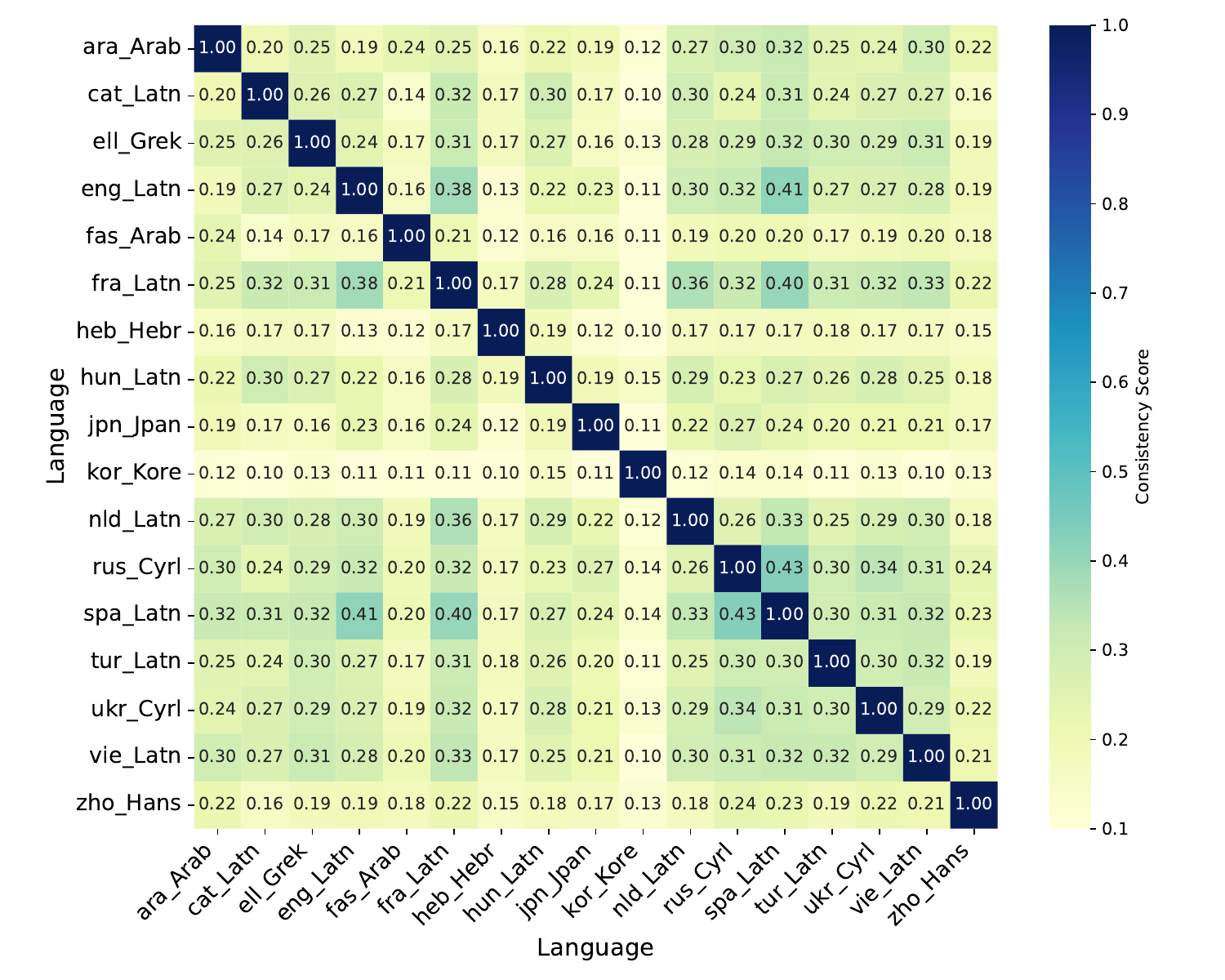}
    }
    \subcaptionbox{\scriptsize OLMo (1B) \textsc{SubInj}}{%
        \includegraphics[width=0.22\textwidth]{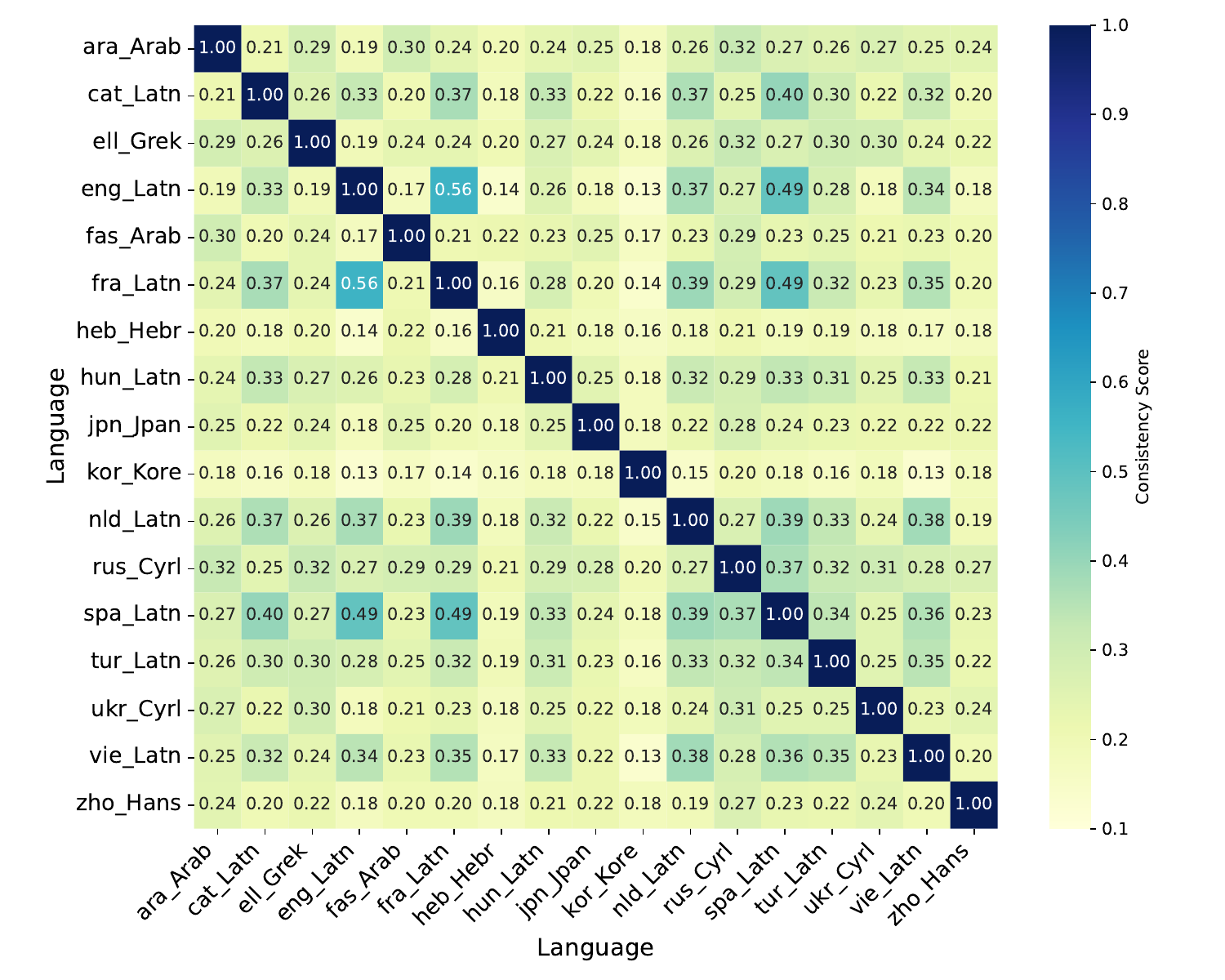}
    }
    \subcaptionbox{\scriptsize OLMo (13B) \textsc{SubSub}}{%
        \includegraphics[width=0.22\textwidth]{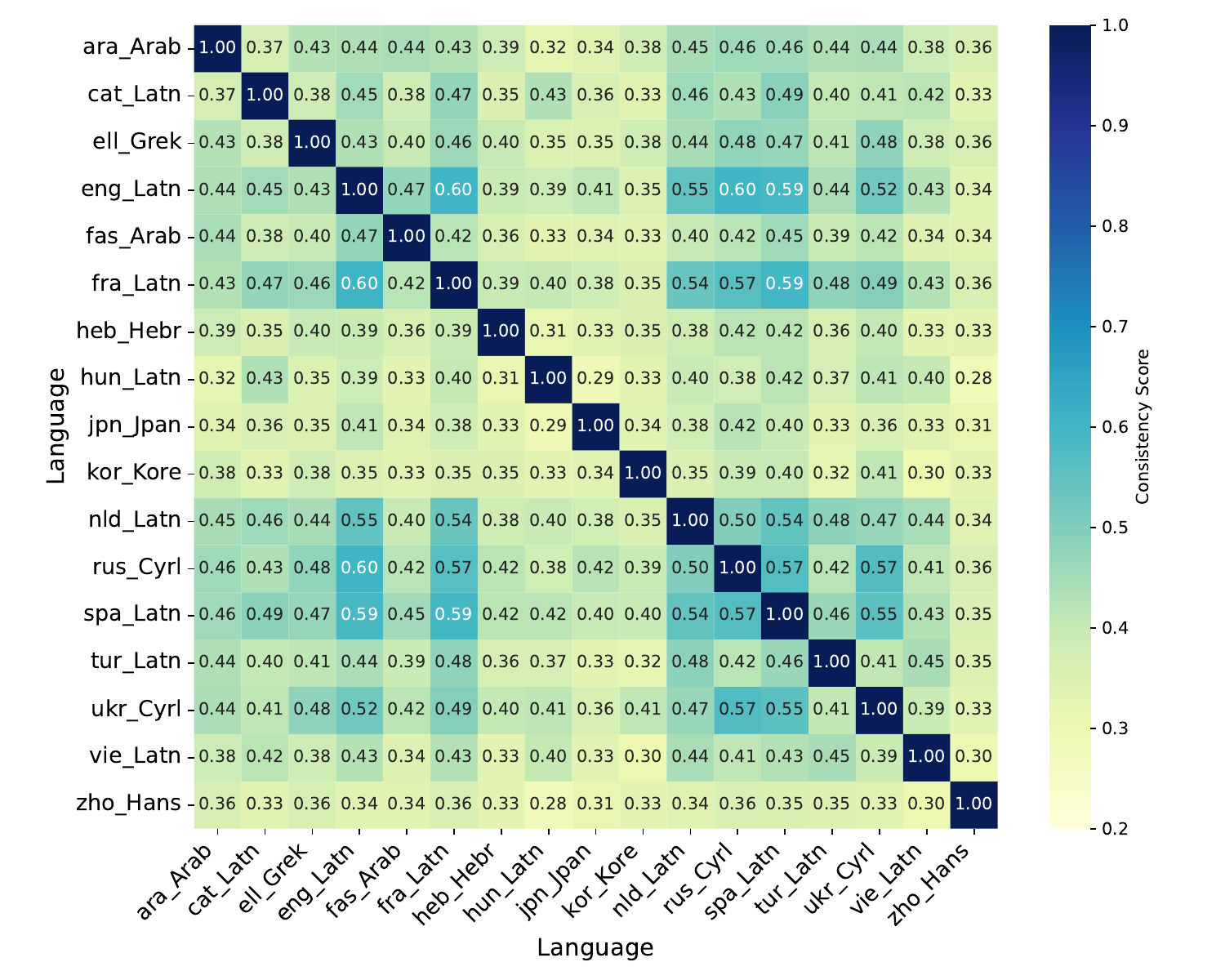}
    }
    \subcaptionbox{\scriptsize OLMo (13B) \textsc{SubInj}}{%
        \includegraphics[width=0.22\textwidth]{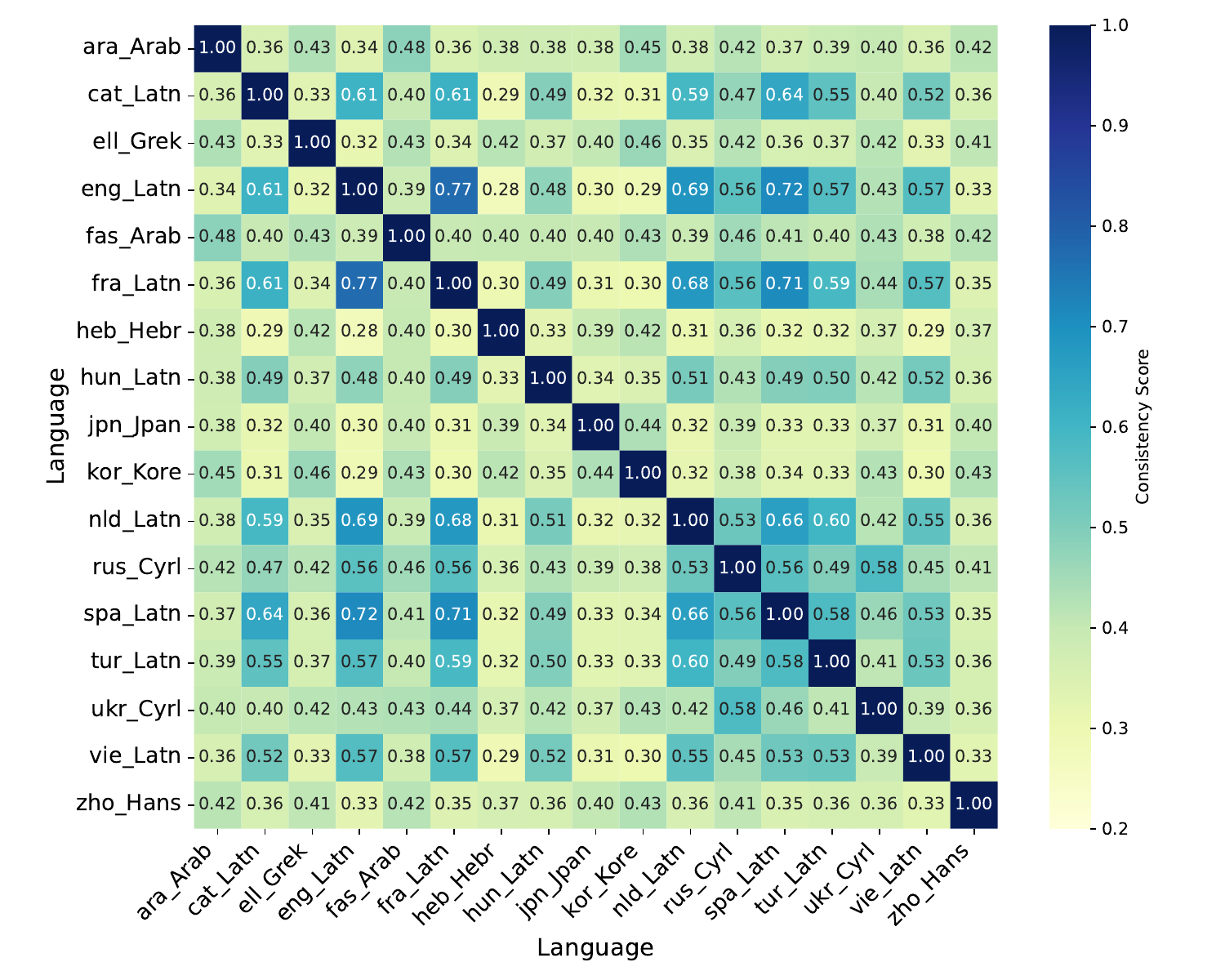}
    }
    \caption{Factual recall consistency across language pairs using \textbf{Japanese} translations of subjects in \textsc{SubSub} and \textsc{SubInj}.}
    \label{fig:perf_ja}
\end{figure*}

In \secref{remedies}, we propose two simple but effective strategies -- \textsc{SubSub} and \textsc{SubInj} -- which inject or substitute the subject entity in prompts with its English translation. 
These interventions yield substantial improvements in both factual recall accuracy and consistency, particularly for smaller multilingual models and English-centric models.
To better understand the role of English (the pivot language) in these gains, we conduct a complementary study using \textbf{Spanish} and \textbf{Japanese} translations of subject entities instead using four models: LLaMA (1B), LLaMA (8B), OLMo (1B), and OLMo (13B).
Table~\ref{tab:subinj_perf_en_ja_es} presents the aggregated performance; Figures~\ref{fig:perf_en}, \ref{fig:perf_es}, \ref{fig:perf_ja} report the consistency across language pairs using English, Spanish, and Japanese subject translations, respectively.

\textbf{Japanese underperforms, especially in English-centric models.}
Using Japanese as the pivot language results in limited or even negative gains. 
For example, in OLMo (1B), \textsc{SubSub} with Japanese degrades performance compared to the base prompt. 
These findings suggest that when the pivot language differs substantially from the model's conceptual embedding space, entity alignment becomes harder, leading to poorer factual recall accuracy and consistency. 
This highlights the critical role of conceptual space alignment in driving factual recall.

\textbf{Conceptual space bias explains differences across pivot languages.}
These observations support our broader hypothesis: the model's ability to project entities into a shared, language-agnostic conceptual space facilitates consistent factual recall. 
In practice, however, this space is not neutral -- it is biased toward high-resource or pretraining-dominant languages like English.
As a result, using English-translated subjects helps the model activate this space more effectively than using other languages like Japanese.
Spanish, which shares the script and extensive lexical overlap, therefore offers comparable improvements with English.


\begin{table*}[t]
\centering
\footnotesize
\setlength{\tabcolsep}{2.5pt}
\begin{tabular}{llrrrrr|rrrrr}
\toprule
\multirow{2}{*}{Model} & \multirow{2}{*}{Lang} & \multicolumn{5}{c}{Recall (ACC)} & \multicolumn{5}{c}{Consistency (CO)} \\
 & & Base & \textsc{SubSub} & {\textsc{SubInj}} & $\uparrow_{\textsc{SubSub}}$ (\%) & $\uparrow_{\textsc{SubInj}}$ (\%) & Base & {\textsc{SubSub}} & {\textsc{SubInj}} & $\uparrow_{\textsc{SubSub}}$ (\%) & $\uparrow_{\textsc{SubInj}}$ (\%) \\
\midrule
LLaMA (1B) & eng & 0.53 & \underline{0.60} & \textbf{0.61} & 11.7 & 14.9 & 0.54 & \underline{0.61} & \textbf{0.63} & 13.2 & 17.8 \\
 & jpn & \underline{0.53} & {0.41} & \textbf{0.55} & -23.3 & 2.8 & \underline{0.54} & {0.50} & \textbf{0.57} & -7.2 & 5.2 \\
 & spa & 0.53 & \underline{0.56} & \textbf{0.59} & 4.3 & 10.6 & 0.54 & \underline{0.59} & \textbf{0.61} & 9.2 & 14.4 \\
 \hline
LLaMA (8B) & eng & 0.71 & \underline{0.75} & \textbf{0.76} & 5.6 & 7.2 & 0.74 & \underline{0.80} & \textbf{0.82} & 7.7 & 10.0 \\
 & jpn & \underline{0.71} & {0.66} & \textbf{0.73} & -6.9 & 3.4 & \underline{0.74} & {0.73} & \textbf{0.78} & -1.9 & 5.0 \\
 & spa & 0.71 & \underline{0.72} & \textbf{0.75} & 1.5 & 5.7 & 0.74 & \underline{0.77} & \textbf{0.80} & 4.3 & 8.2 \\
 \hline
OLMo (1B) & eng & 0.27 & \underline{0.28} & \textbf{0.31} & 2.4 & 11.2 & 0.27 & \underline{0.24} & \textbf{0.28} & -8.7 & 4.8 \\
 & jpn & \textbf{0.27} & {0.15} & \underline{0.27} & -46.2 & -2.7 & \textbf{0.27} & {0.23} & \underline{0.25} & -13.4 & -5.1 \\
 & spa & 0.27 & \underline{0.25} & \textbf{0.29} & -8.6 & 5.0 & 0.27 & \underline{0.24} & \textbf{0.27} & -7.8 & 1.7 \\
 \hline
OLMo (13B) & eng & 0.43 & \underline{0.54} & \textbf{0.56} & 25.9 & 31.6 & 0.39 & \underline{0.52} & \textbf{0.55} & 35.3 & 43.9 \\
 & jpn & \underline{0.43} & {0.27} & \textbf{0.45} & -36.2 & 4.7 & 0.39 & \underline{0.41} & \textbf{0.43} & 5.8 & 10.5 \\
 & spa & 0.43 & \underline{0.49} & \textbf{0.53} & 15.1 & 24.1 & 0.39 & \underline{0.50} & \textbf{0.53} & 30.2 & 36.9 \\
\bottomrule
\end{tabular}
\caption{
Performance comparison of the proposed remedies \textsc{SubSub} and \textsc{SubInj} for improving entity-level alignment, using English, Japanese, and Spanish translations of subjects. 
Both methods improve recall and consistency when the intervention leverages translations from \textbf{high-resource languages} such as English and Spanish, indicating that the alignment benefits from the conceptual space biased towards these languages.
}
\label{tab:subinj_perf_en_ja_es}
\end{table*}

\begin{figure*}[h]
    \centering
    \subcaptionbox{\scriptsize LLaMA (1B)}{%
        \includegraphics[width=0.22\textwidth]{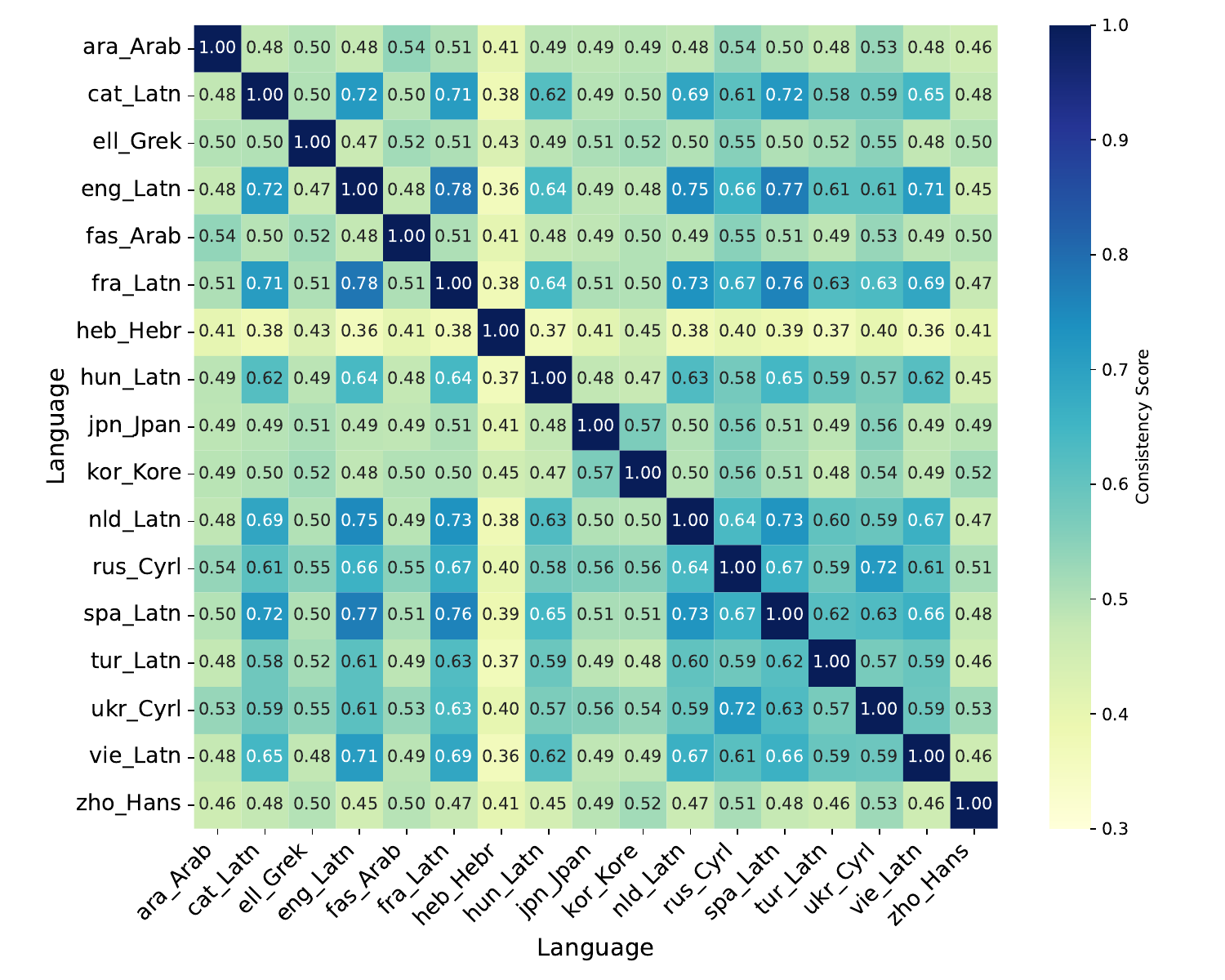}
    }
    \subcaptionbox{\scriptsize LLaMA (8B)} {%
        \includegraphics[width=0.22\textwidth]{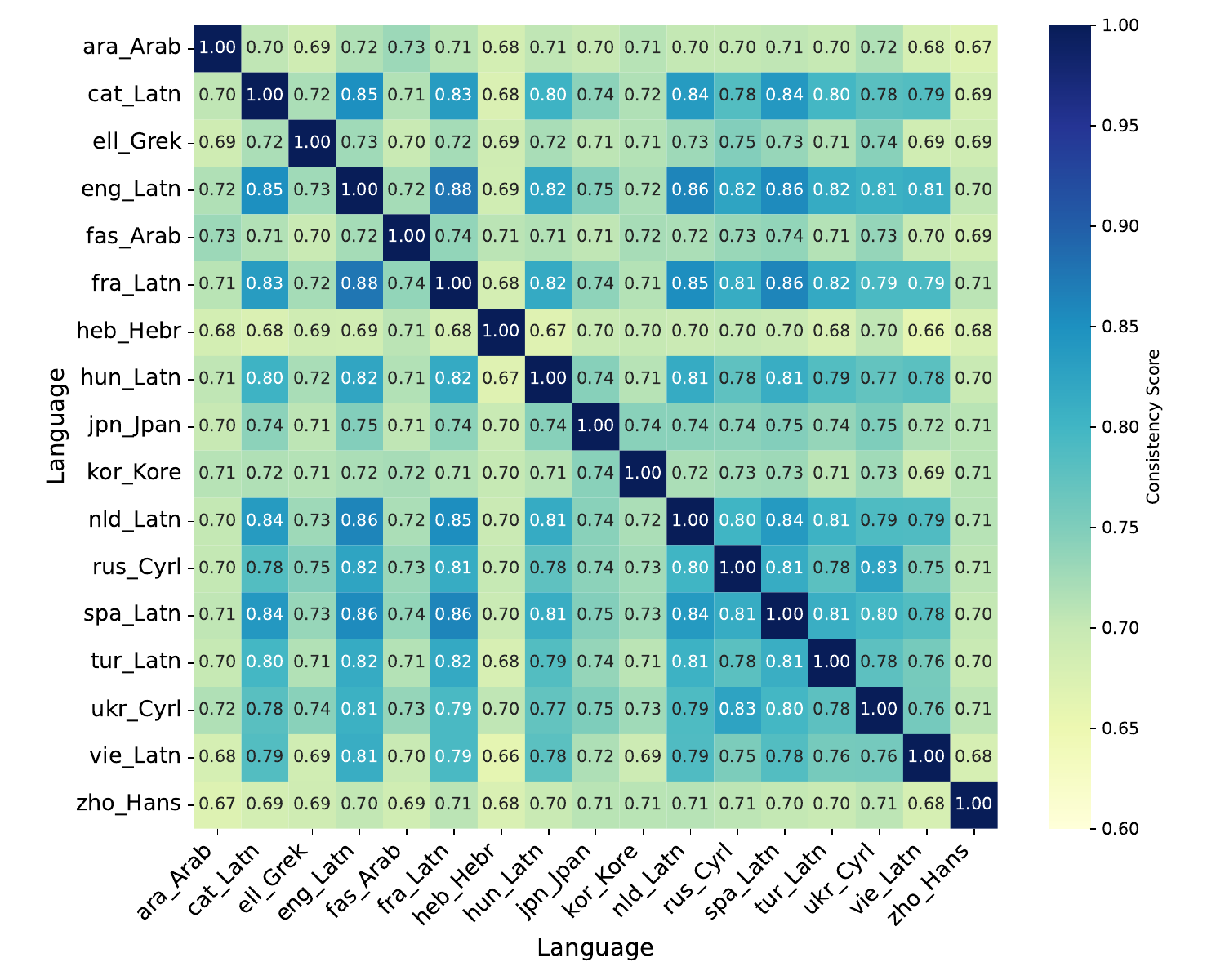}
    }
    \subcaptionbox{\scriptsize OLMo (1B)}{%
        \includegraphics[width=0.22\textwidth]{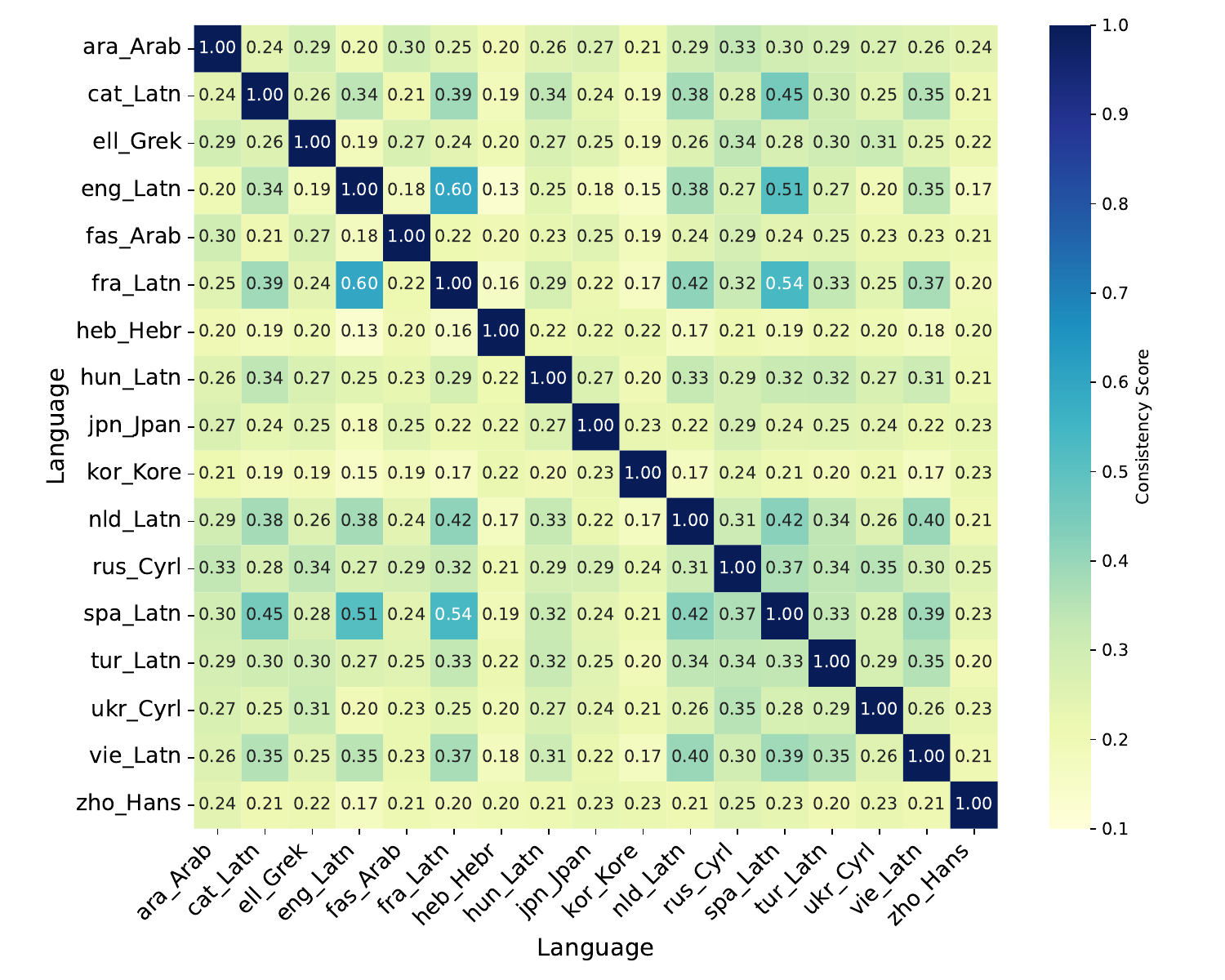}
    }
    \subcaptionbox{\scriptsize OLMo (13B)}{%
        \includegraphics[width=0.22\textwidth]{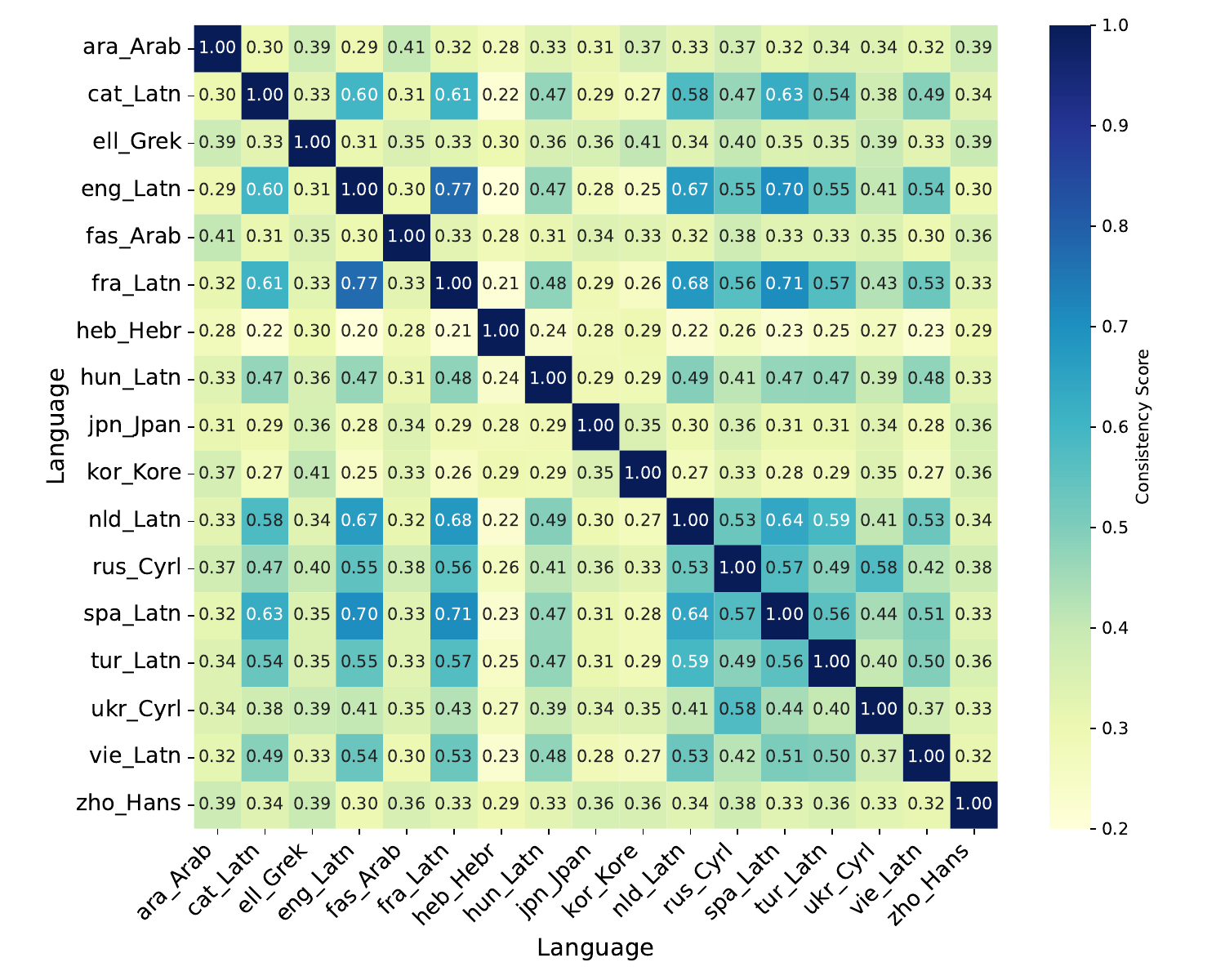}
    }
    \caption{Factual recall consistency across language pairs under \textbf{baseline prompting}, with language-specific factual queries issued in their original languages.}

    \label{fig:perf_orig}
\end{figure*}

\section{Mechanistic Interpretability: Additional Results}\seclabel{mech-interp-additional}

As mentioned in \secref{mechanistic}, we use Logit Lens analysis to examine how \textsc{SubSub} and \textsc{SubInj} influence internal model representations. Here, we present the results on LLaMA (1B) and OLMo (1B).

\begin{figure*}[h!]
    \centering

    \begin{subfigure}{\linewidth}
        \centering
        \includegraphics[width=0.47\linewidth]{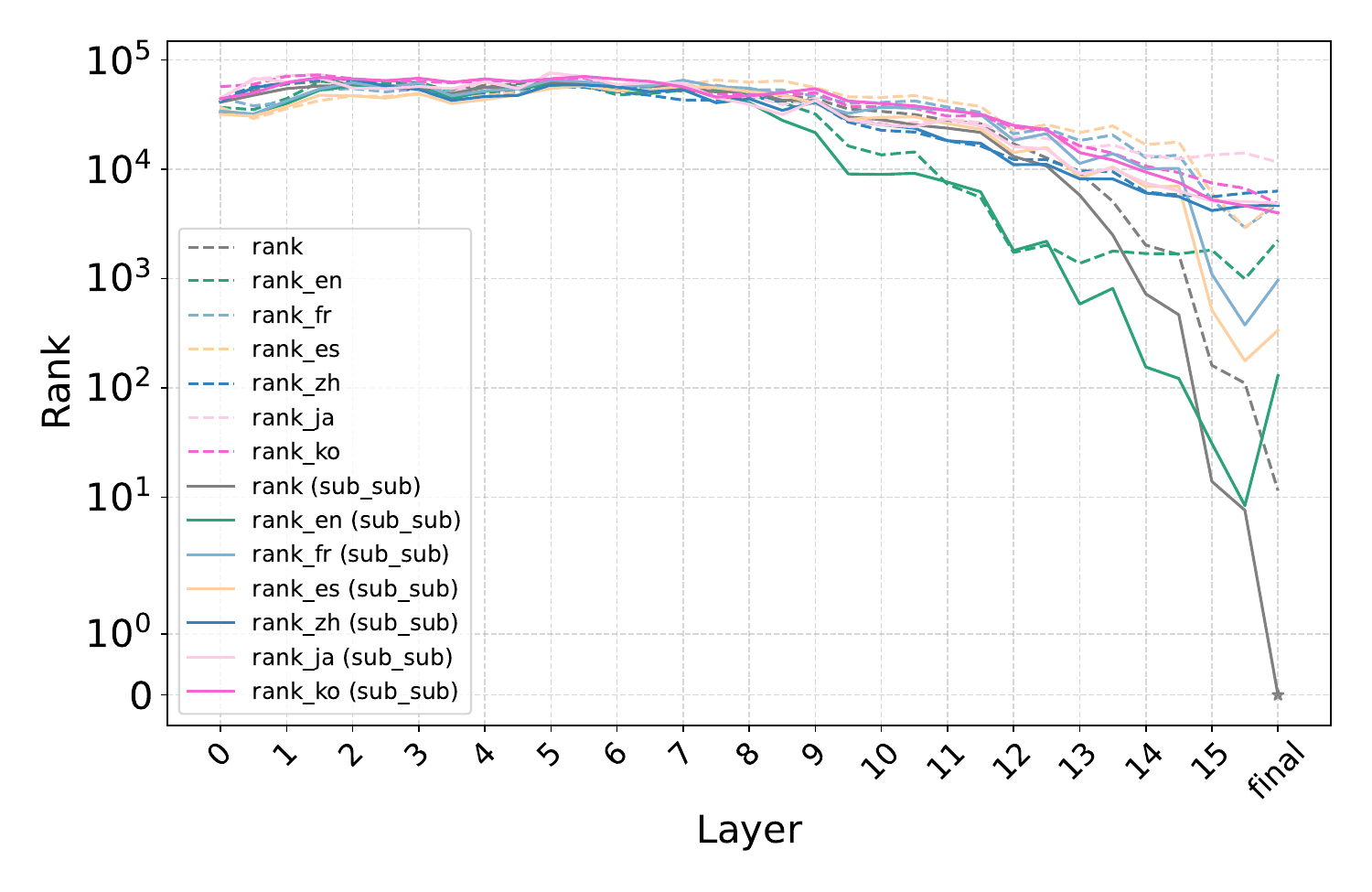}
        \includegraphics[width=0.47\linewidth]{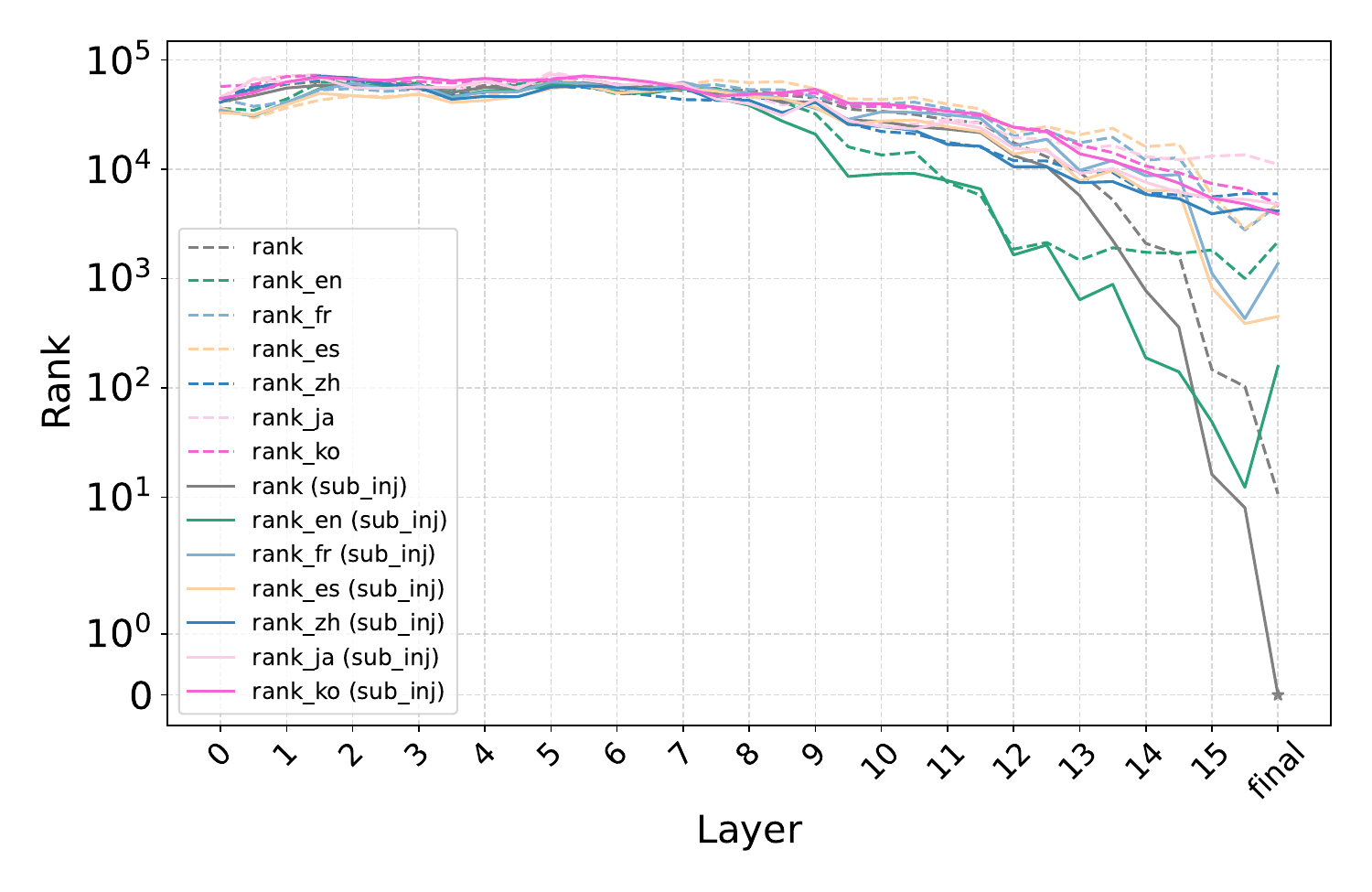}
        \caption{\textbf{LLaMA (1B)}: Left -- with \textsc{SubSub}; Right -- with \textsc{SubInj}.}
        \label{fig:logit-lens-rank-llama3-1b}
    \end{subfigure}

    \begin{subfigure}{\linewidth}
        \centering
        \includegraphics[width=0.47\linewidth]{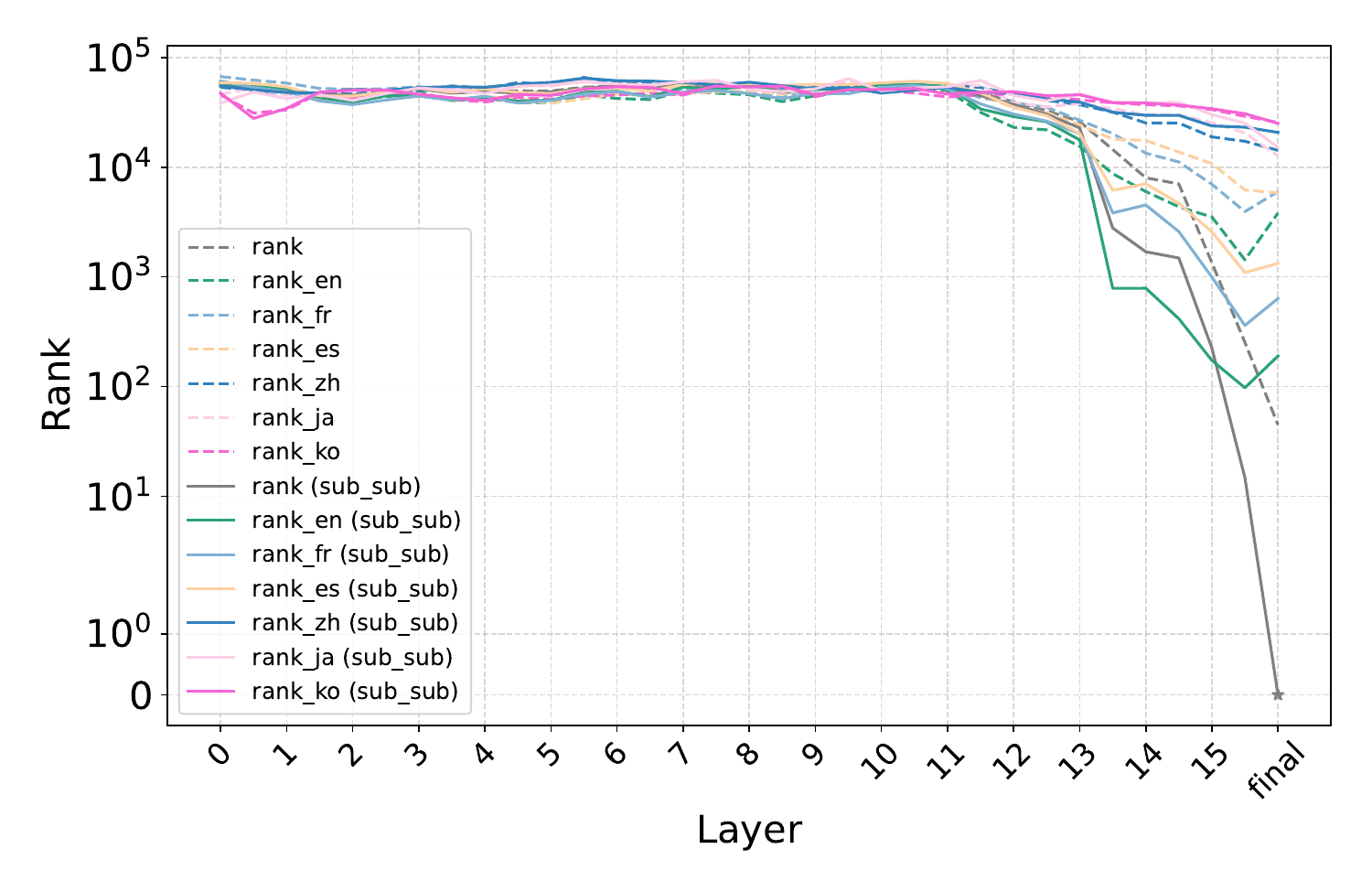}
        \includegraphics[width=0.47\linewidth]{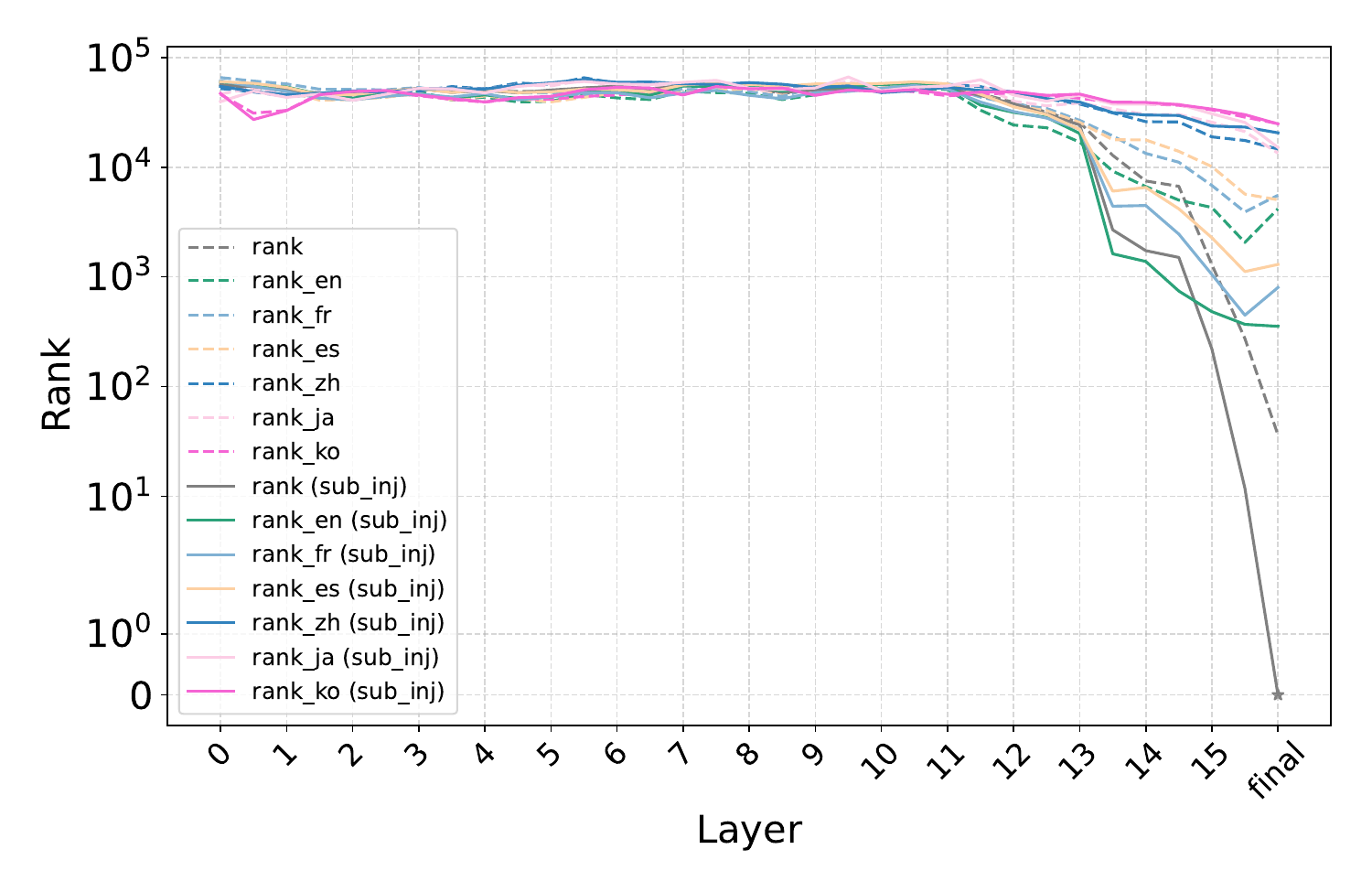}
        \caption{\textbf{OLMo (1B)}: Left -- with \textsc{SubSub}; Right -- with \textsc{SubInj}.}
        \label{fig:logit-lens-rank-olmo-1b}
    \end{subfigure}

    \caption{\textbf{Logit Lens analysis of object token ranks across model layers.} We plot the rank of the target object token (lower is better) in its input language, along with its equivalents in six other languages, using lines of different colors. Dashed lines represent the original prompts, while solid lines correspond to prompts modified with \textsc{SubSub} and \textsc{SubInj}. 
    Both interventions consistently result in lower ranks across languages compared to the original prompts, indicating entity representations are more aligned in the common conceptual space, and consequently, more consistent crosslingual factual prediction.
    }
    \label{fig:logit-lens-rank-appendix}
\end{figure*}

Figure~\ref{fig:logit-lens-rank-appendix} shows the rank of the target object token and its equivalents across six languages under original prompts, \textsc{SubSub}, and \textsc{SubInj}. We observe consistent patterns with the larger models: both interventions reduce object ranks across layers, this confirms that our findings generalize across different model scales.

\begin{figure*}[h!]
    \centering

    \begin{subfigure}{\linewidth}
        \centering
        \includegraphics[width=0.47\linewidth]{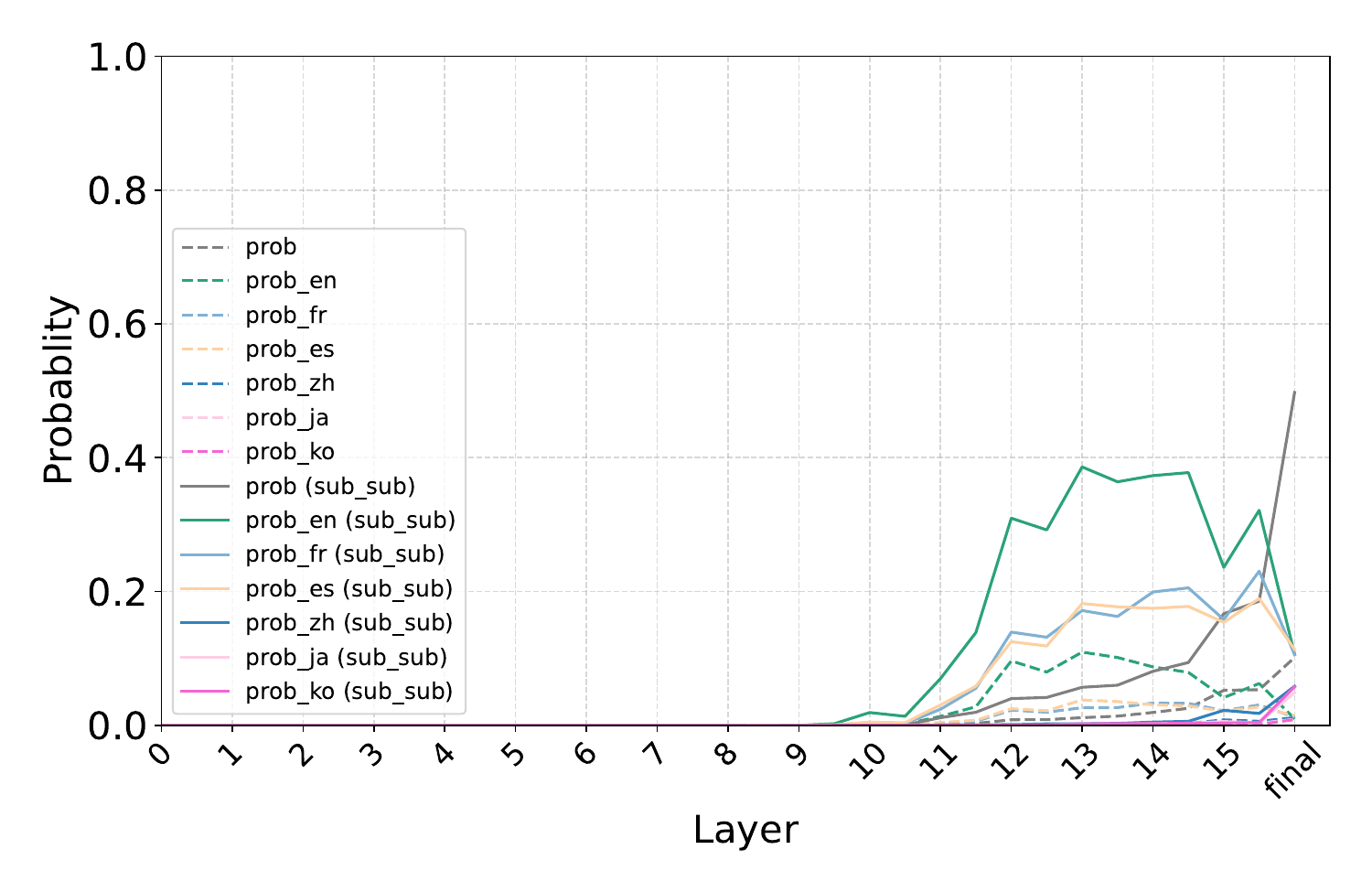}
        \includegraphics[width=0.47\linewidth]{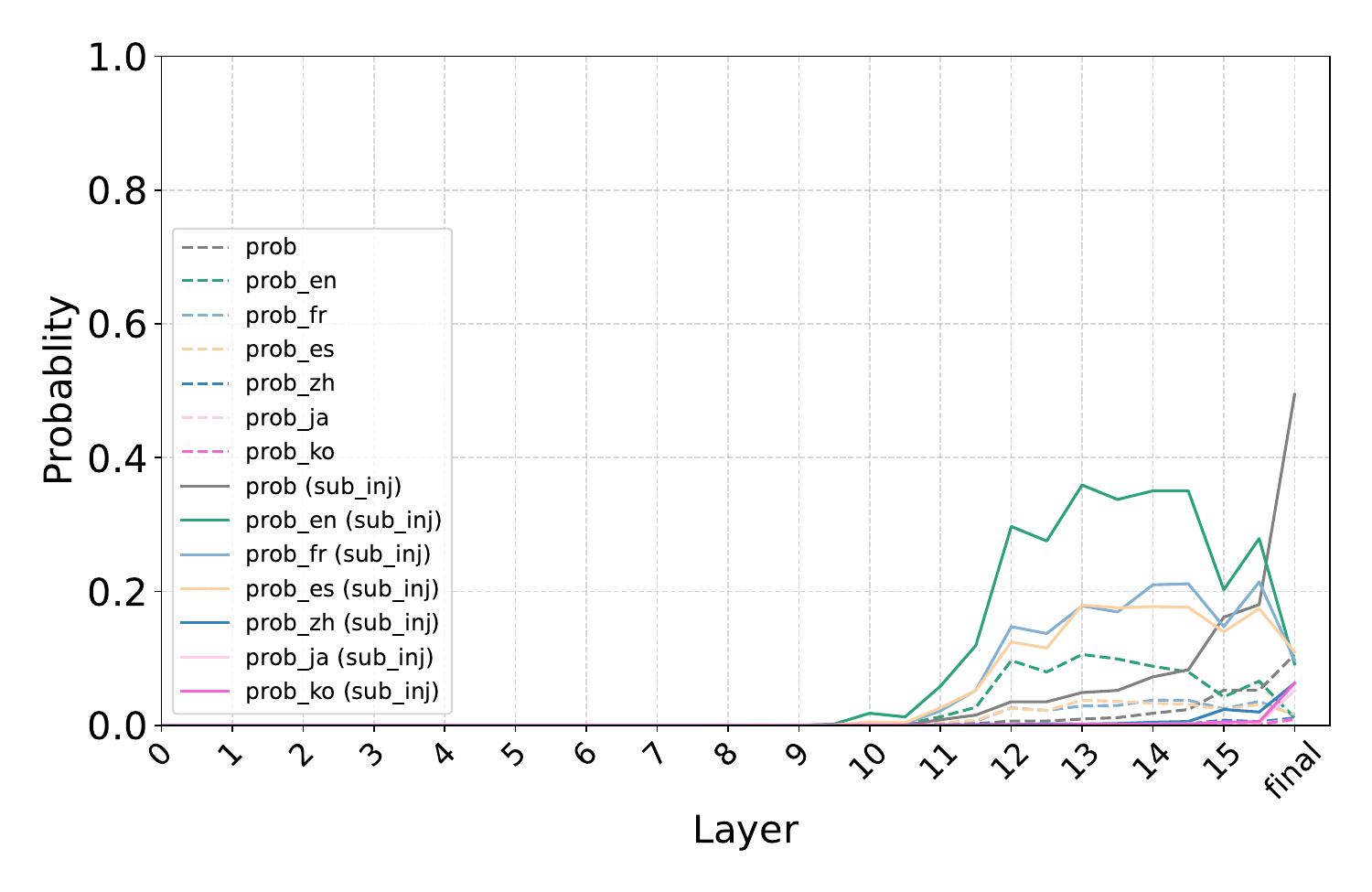}
        \caption{\textbf{LLaMA (1B)}: Left -- with \textsc{SubSub}; Right -- with \textsc{SubInj}.}
        \label{fig:logit-lens-rank-llama3-1b-prob}
    \end{subfigure}

    \begin{subfigure}{\linewidth}
        \centering
        \includegraphics[width=0.47\linewidth]{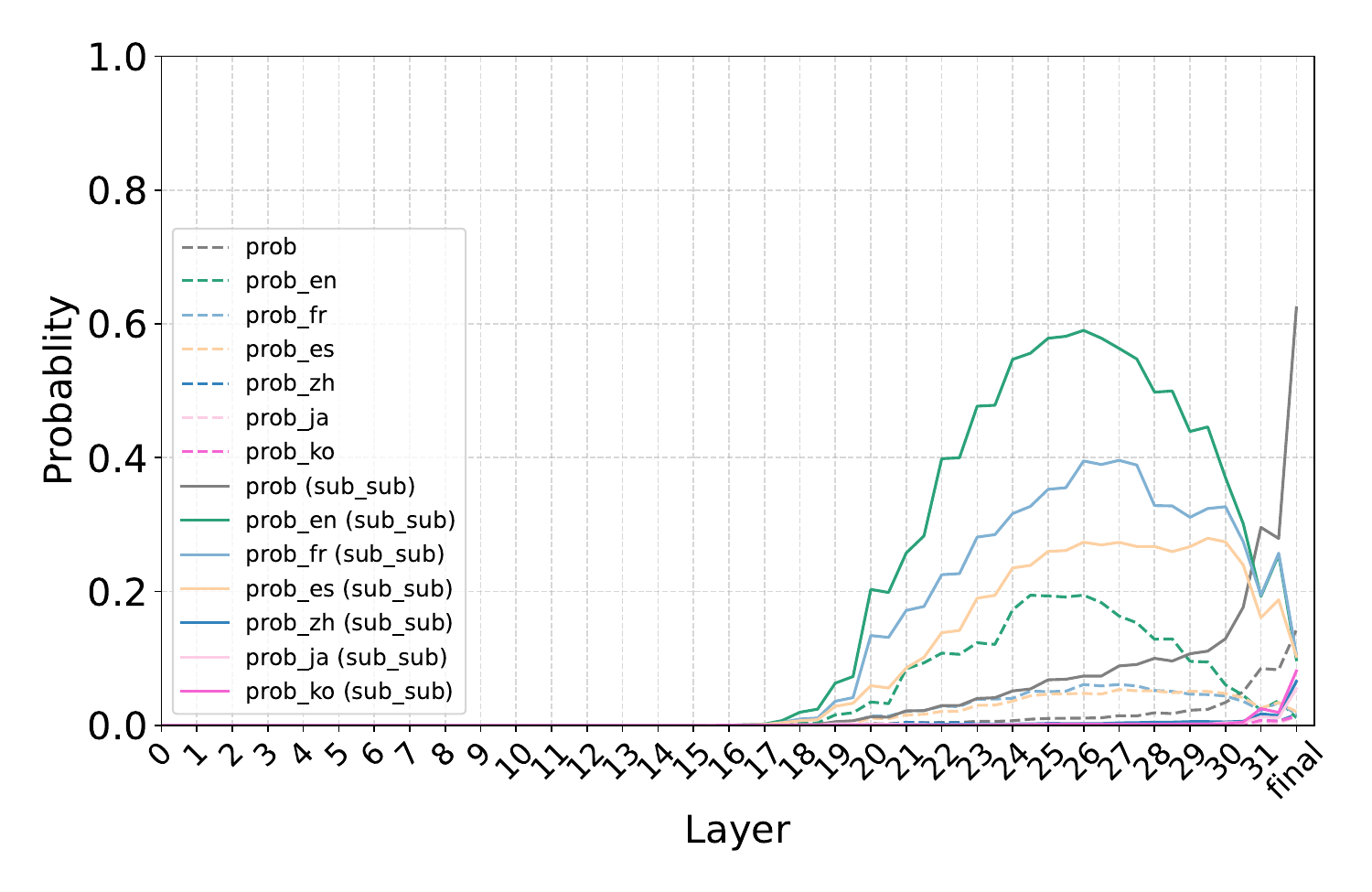}
        \includegraphics[width=0.47\linewidth]{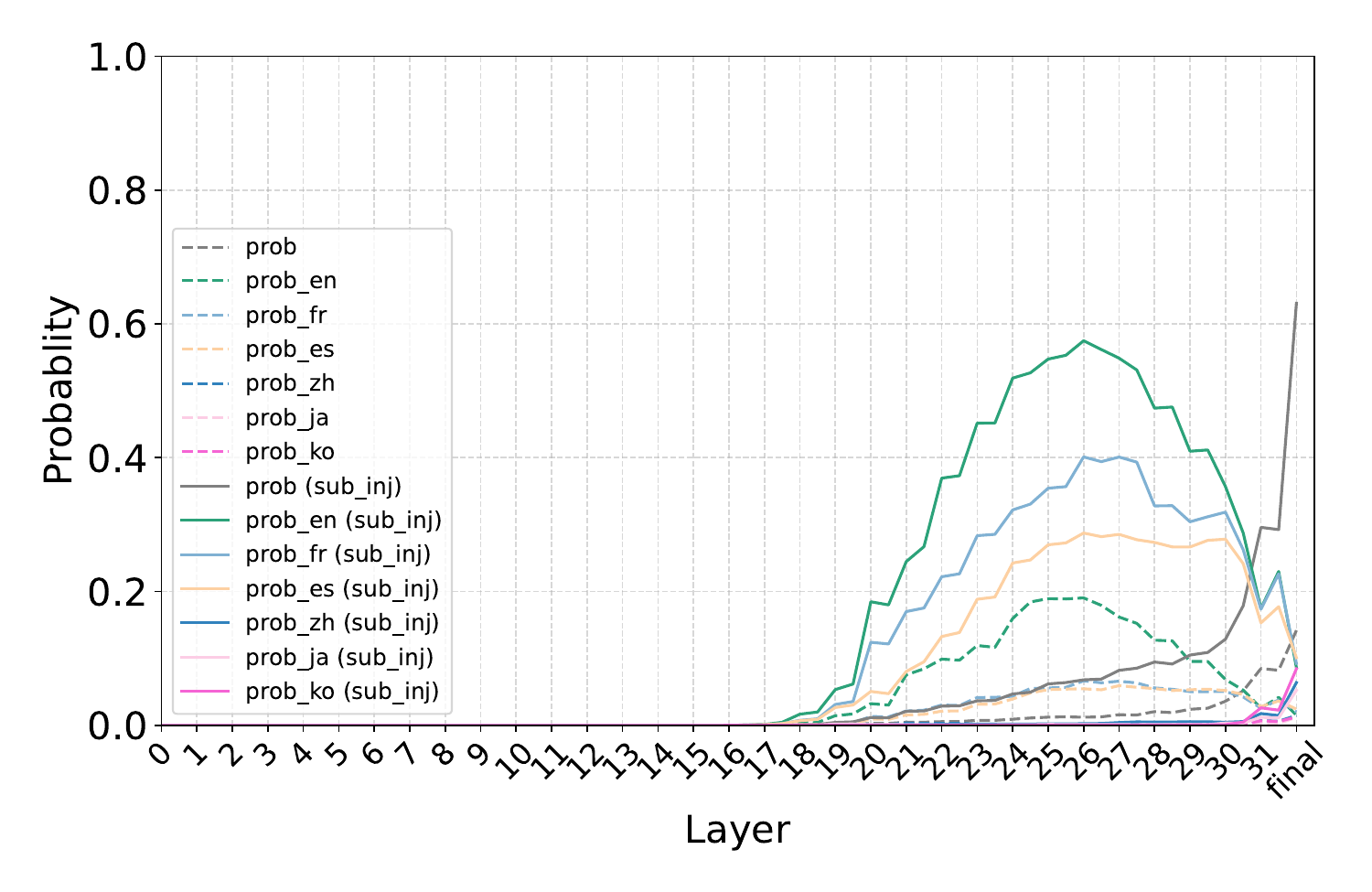}
        \caption{\textbf{LLaMA (8B)}: Left -- with \textsc{SubSub}; Right -- with \textsc{SubInj}.}
        \label{fig:logit-lens-rank-llama3-8b-prob}
    \end{subfigure}

    \begin{subfigure}{\linewidth}
        \centering
        \includegraphics[width=0.47\linewidth]{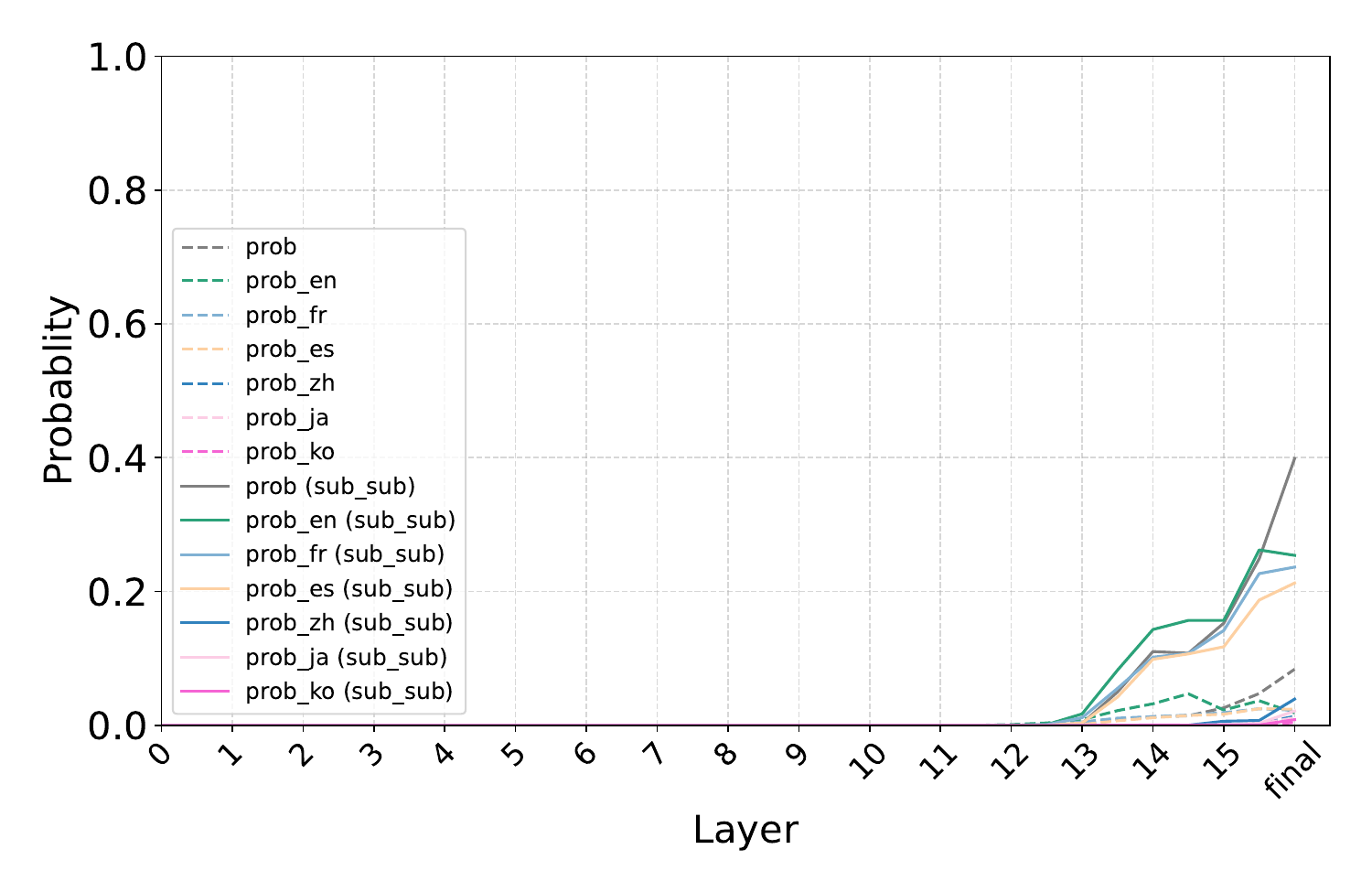}
        \includegraphics[width=0.47\linewidth]{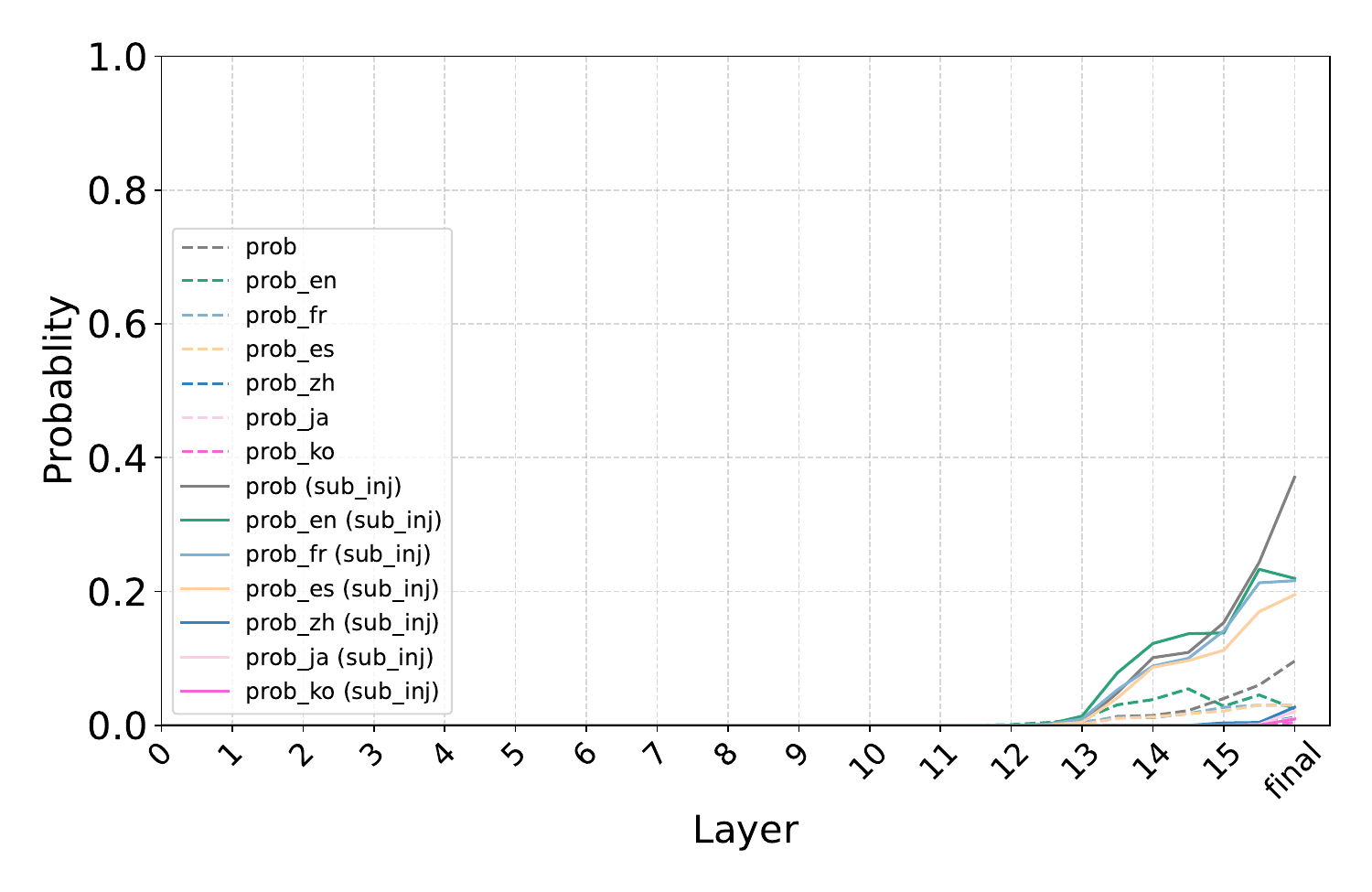}
        \caption{\textbf{OLMo (1B)}: Left -- with \textsc{SubSub}; Right -- with \textsc{SubInj}.}
        \label{fig:logit-lens-rank-olmo-1b-prob}
    \end{subfigure}
    
    \begin{subfigure}{\linewidth}
        \centering
        \includegraphics[width=0.47\linewidth]{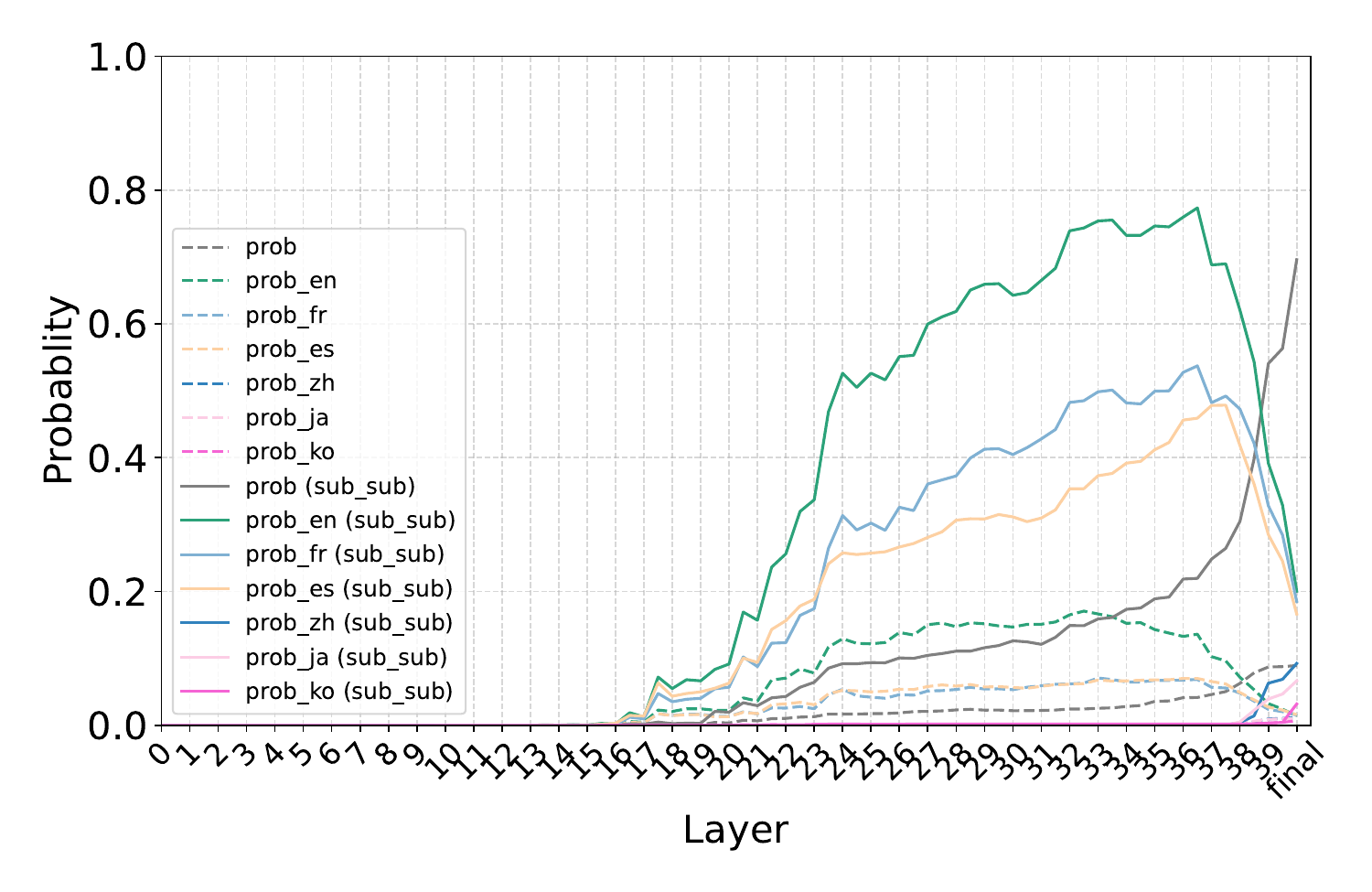}
        \includegraphics[width=0.47\linewidth]{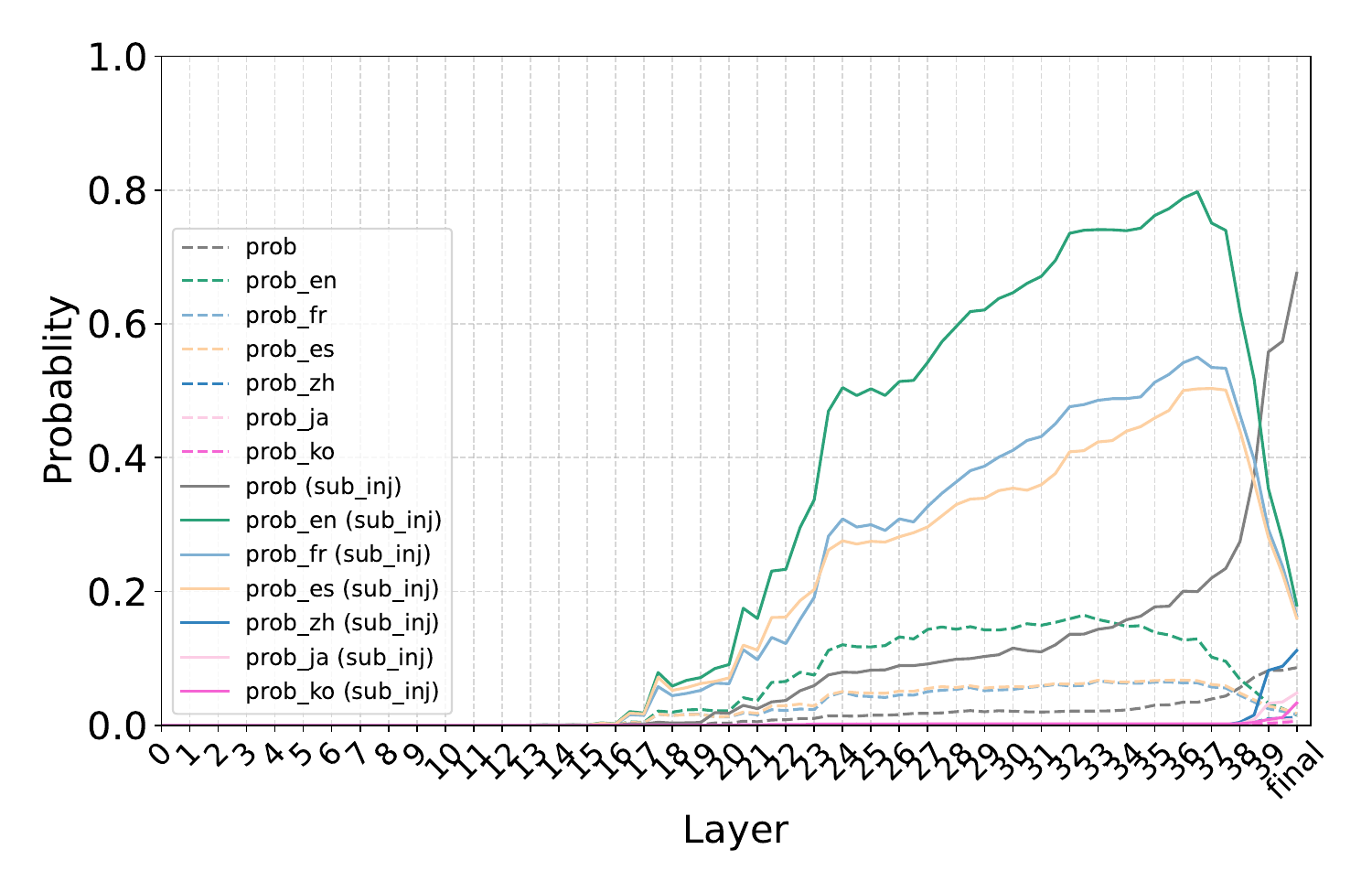}
        \caption{\textbf{OLMo (13B)}: Left -- with \textsc{SubSub}; Right -- with \textsc{SubInj}.}
        \label{fig:logit-lens-rank-olmo-13b-prob}
    \end{subfigure}

    \caption{\textbf{Logit Lens analysis of object token probabilities across model layers.} We plot the probability of the target object token in its input language, along with its equivalents in six other languages, using lines of different colors. Dashed lines represent the original prompts, while solid lines correspond to prompts modified with \textsc{SubSub} and \textsc{SubInj}. Both interventions consistently result in higher probabilities across languages compared to the original prompts, indicating entity representations are more aligned in the conceptual space, and consequently, more consistent crosslingual factual prediction.}
    \label{fig:logit-lens-prob-appendix}
\end{figure*}

In addition to rank analysis, we also examine probability curves for the target object token in each language under the three prompting conditions (original prompt, \textsc{SubSub}, and \textsc{SubInj}). As shown in Figure~\ref{fig:logit-lens-prob-appendix}, the interventions consistently increase the predicted probability of the correct object across layers. This complements the rank analysis and further demonstrates that providing English subject cues enhances crosslingual entity alignment, making the correct object more accessible during decoding.

\end{document}